\documentclass[10pt,twocolumn,letterpaper]{article}

\usepackage[pagenumbers]{iccv} 

\definecolor{iccvblue}{rgb}{0.21,0.49,0.74}
\usepackage[pagebackref,breaklinks,colorlinks,allcolors=iccvblue]{hyperref}
\usepackage{multirow}
\usepackage{makecell}
\def\paperID{1609} 
\def\confName{ICCV}
\def\confYear{2025}

\title{SeaS: Few-shot Industrial Anomaly Image Generation with\\Separation and Sharing Fine-tuning}

\author{Zhewei~Dai$^{1,}$\thanks{ Contributed Equally, $^{\dag}$ Corresponding Authors.} , Shilei~Zeng$^{1, *}$, Haotian~Liu$^{1}$, Xurui~Li$^{1}$, Feng~Xue$^{4}$, Yu~Zhou$^{1, 2, 3,\dag}$ 
\\
  $^{1}$ School of Electronic Information and Communications, Huazhong University of Science and Technology\\
  $^{2}$ Hubei Key Laboratory of Smart Internet Technology, Huazhong University of Science and Technology\\
  $^{3}$ Artificial Intelligence Research Institute, Wuhan JingCe Electronic Group Co.,LTD \\
   $^{4}$ Department of Information Engineering and Computer Science, University of Trento \\
  \tt\small{\{zwdai,shlzeng,htliu\_master,xrli\_plus,yuzhou\}@hust.edu.cn}, \tt\small{xuefengbupt@gmail.com}
}

\begin{document}
\maketitle
\begin{abstract}
\noindent We introduce SeaS,
a unified industrial generative model for automatically creating diverse anomalies, authentic normal products, and precise anomaly masks.
While extensive research exists, most efforts either focus on specific tasks,
i.e., anomalies or normal products only, or require separate models for each anomaly type.
Consequently, prior methods either offer limited generative capability or depend on a vast array of anomaly-specific models.
We demonstrate that U-Net's differentiated learning ability captures the distinct visual traits of slightly-varied normal products and diverse anomalies,
enabling us to construct a unified model for all tasks.
Specifically, we first introduce an Unbalanced Abnormal (UA) Text Prompt,
comprising one normal token and multiple anomaly tokens.
More importantly, our Decoupled Anomaly Alignment (DA) loss decouples anomaly attributes and binds them to distinct anomaly tokens of UA,
enabling SeaS to create unseen anomalies by recombining these attributes.
Furthermore,
our Normal-image Alignment (NA) loss aligns the normal token to normal patterns,
making generated normal products globally consistent and locally varied.
Finally, SeaS produces accurate anomaly masks by fusing discriminative U-Net features with high-resolution VAE features.
SeaS sets a new benchmark for industrial generation,
significantly enhancing downstream applications, 
with average improvements of $+8.66\%$ pixel-level AP for synthesis-based AD approaches, $+1.10\%$ image-level AP for unsupervised AD methods, and $+12.79\%$ IoU for supervised segmentation models. Code is available at \href{https://github.com/HUST-SLOW/SeaS}{https://github.com/HUST-SLOW/SeaS}.
\end{abstract}
\vspace{-0.6cm}
\section{Introduction}
\label{sec:intro}
In the industrial scenario,
generative models are used to synthesise various visual elements,
which meet the requirements of different anomaly detection (AD) methods and supervised segmentation models as below.
\begin{itemize}
    \item Generating pseudo-anomalies for synthesis-based AD approaches \citep{zavrtanik2021draem,chen2024unifiedglass}.
    \item Generating pseudo-normal images for unsupervised AD methods \citep{lu2023hvqtrans,roth2022patchcore,he2025mambaad}.
    \item Generating complete anomaly images and corresponding accurate masks for training supervised segmentation models.
\end{itemize}

\begin{figure}[t]
\begin{center}
\includegraphics[width=1\linewidth]{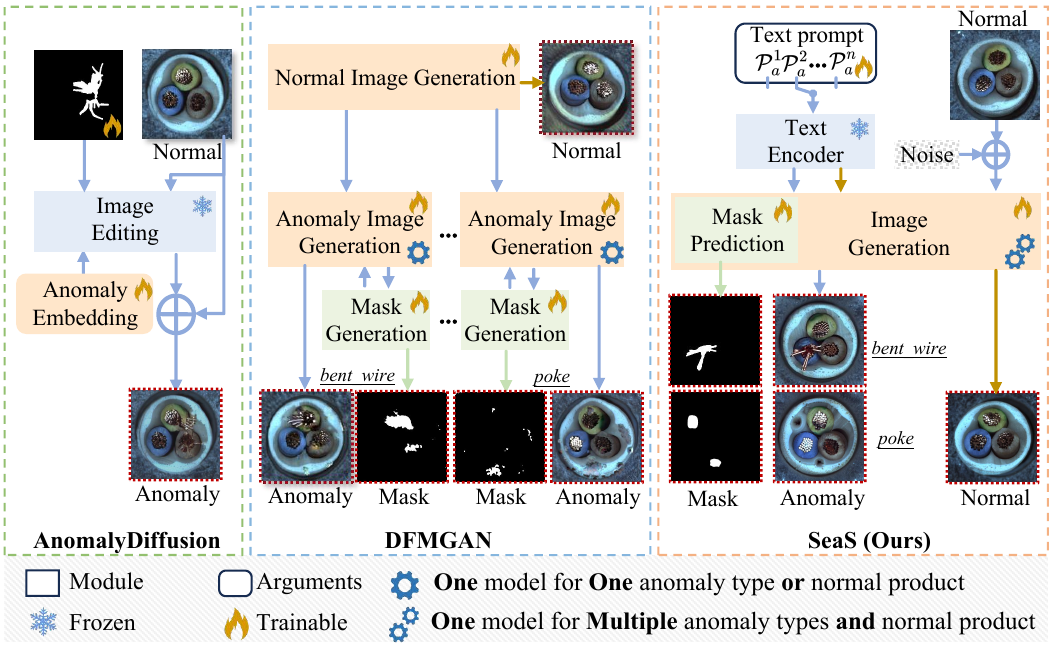}
\vspace{-0.7cm}
\caption{
\textbf{(a) AnomalyDiffusion} only generates anomalies and edits them onto the input normal images guided by the predefined masks.
\textbf{(b) DFMGAN} trains multiple dedicated generators per anomaly type or normal product, and it cannot produce accurate masks.
\textbf{(c) SeaS} trains a unified model capable of generating anomaly images and masks for multiple anomaly types, as well as normal images.
}
\label{introduction}
\end{center}
\vspace{-1.1cm}
\end{figure}

These requirements above have been covered by previous algorithms.
CutPaste \citep{li2021cutpaste} creates anomalies by pasting cropped normal regions onto normal product images (short for normal images).
AnomalyDiffusion \citep{hu2023anomalydiffusion} generates anomalies using diffusion models and edits anomalies onto normal images guided by pre-defined masks (Fig. \ref{introduction}(a)),
but it cannot create pseudo-normal images and might suffer from misaligned masks.
DFMGAN \citep{Duan2023DFMGAN} trains separate models for each normal product and anomaly type (Fig. \ref{introduction}(b)) but cannot produce accurate masks,
limiting its effectiveness in training supervised segmentation models.
In summary, existing methods either focus only on the generation of normal products or anomalies,
or require multiple isolated models to complete all tasks separately.
They cannot flexibly use a unified model to tackle them all,
i.e., achieving diverse anomalies, authentic normal products, and pixel-accurate masks.
In this paper,
we study the unified industrial anomaly generation solution,
meeting the needs of various downstream tasks.

The novelty of this work stems from a key observation on a single industrial production line:
\textbf{\textit{normal products exhibit a globally consistent surface with minor local variations, whereas anomalies exhibit high variability}}.
These characteristics can be effectively captured by U-Net due to its differential learning capability in a diffusion process.
Building on this insight,
we propose a \textbf{Se}p\textbf{a}ration and \textbf{S}haring Fine-tuning method (SeaS), using a shared U-Net to model distinct variations. 
Firstly, to explicitly model the variations of normal products and anomalies, 
we propose Unbalanced Abnormal (UA) Text Prompt. 
Its unbalanced design includes one normal token and multiple anomaly tokens, %
thus decoupling the presentations of the slightly varied normal product surface and diverse anomaly semantics.
Secondly, to learn highly-diverse anomalies,
we propose a Decoupled Anomaly Alignment (DA) loss to bind the attributes of anomalies to different anomaly tokens of UA.
Recombining the decoupled attributes may produce anomalies that have never been seen in the training dataset,
therefore increasing the diversity of generated anomalies.
Thirdly, for slightly varied normal products, we propose the Normal-image Alignment (NA) loss.
It enables the network to learn the key features of the normal product from normal images,
so that the normal token of UA expresses the products' global consistency.
The two training processes above are separated but conducted on a shared U-Net. 
SeaS enables U-Net to simultaneously model the different variations in both normal products and anomalies, representing the discriminative features for mask prediction. However, the low-resolution features of U-Net lead to a coarse mask when predicted directly. 
Thus, we propose a Refined Mask Prediction (RMP) branch.
It combines U-Net features with high-resolution VAE features to generate accurate and crisp masks progressively. 
The generated anomaly images achieve IS scores by $1.88$ (MVTec AD), $1.27$ (VisA), and $1.95$ (MVTec 3D AD), with IC-LPIPS of $0.34$, $0.26$, and $0.30$. 
On multiple datasets, SeaS-generated images boost synthesis-based AD approaches by an average $+8.66\%$ pixel-level AP, improve unsupervised AD methods by an average $+1.10\%$ image-level AP, and enhance supervised segmentation models by an average $+12.79\%$ IoU.

In summary, the key contribution of our approach lies in:
\begin{itemize}
\item We propose a unified generative model for industrial visual elements. It achieves diverse anomalies, globally consistent normal products, and pixel-level accurate masks using only one model, which sets a new standard for this field.

\item 
The newly designed separated and shared fine-tuning models different variations of normal products and anomalies, enabling precise control over their generation, and obtaining discriminative features for mask prediction.

\item 
SeaS greatly improves the performance of various synthesis-based and unsupervised AD methods, and empowers supervised segmentation models with decent performance.
\end{itemize}

\section{Related Work}
\label{sec:related_work}
\begin{figure*}[!t]
\centering
\includegraphics[width=1\linewidth]{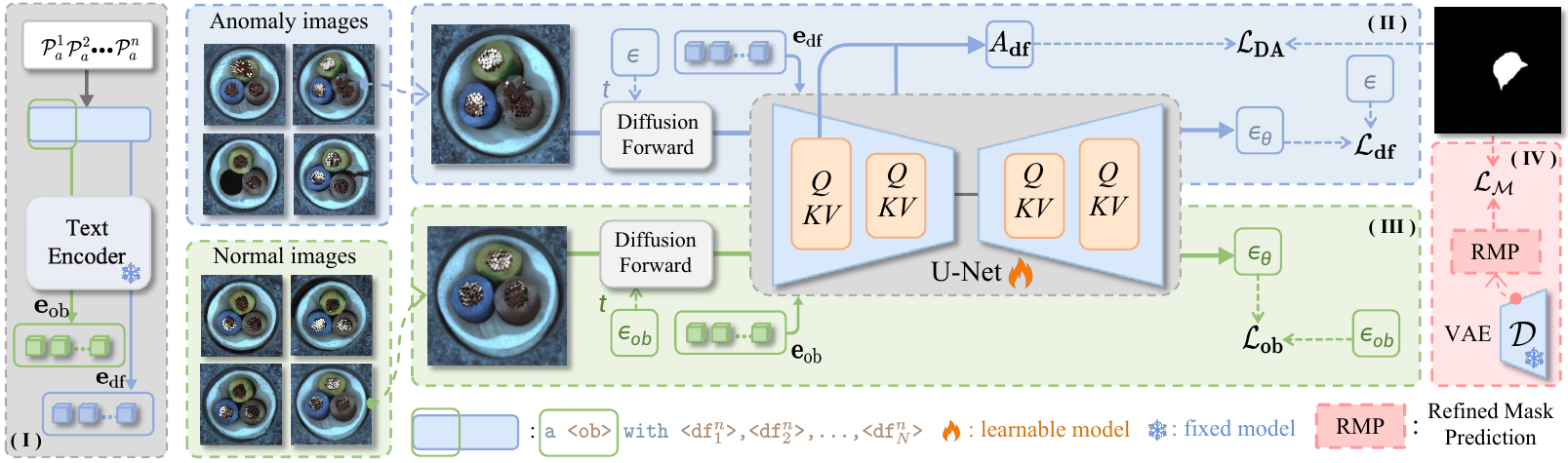}
\vspace{-0.55cm}
\caption{\textbf{Overall framework of SeaS.} It consists of four parts: (\uppercase\expandafter{\romannumeral1}) the Unbalanced Abnormal Text Prompt, (\uppercase\expandafter{\romannumeral2}) the Decoupled Anomaly Alignment for aligning the anomaly tokens \texttt{<}$\texttt{df}_{n}$\texttt{>} to the anomaly area of abnormal images, (\uppercase\expandafter{\romannumeral3}) the Normal-image Alignment for maintaining authenticity through normal images, and (\uppercase\expandafter{\romannumeral4}) the Refined Mask Prediction branch for generating accurate masks.}
\label{pipeline}
\vspace{-0.6cm}
\end{figure*}

\textbf{Anomaly Image Generation.}
Early non-generative methods \citep{devries2017cutout, li2021cutpaste, zavrtanik2021draem} rely on data augmentation to synthesize pseudo-anomalies, but suffer from low fidelity due to inconsistent anomaly patterns.
Some generative methods \citep{schluter2022nsa,hu2023anomalydiffusion,2024anogen} only generate anomalies and merge them into the real normal images. NSA \citep{schluter2022nsa} uses Poisson Image Editing \citep{perez2003poisson} to fuse the cropped normal region. 
However, these methods cannot create pseudo-normal images and require anomaly masks as inputs, with unreasonable mask positions compromising fidelity and consistency.
GAN-based methods \citep{Duan2023DFMGAN, zhang2021defectgan, niu2020sdgan} generate the entire anomaly images.
DFMGAN\citep{Duan2023DFMGAN} trains multiple isolated models to generate normal images and anomaly images for each anomaly type, and the produced masks often do not align accurately with anomalies, limiting their utility in training supervised segmentation models. 
Different from these existing approaches, we propose a unified generative model based on Stable Diffusion to generate diverse anomalies, globally consistent normal products and pixel-level accurate masks.

\textbf{Fine-tuning Diffusion Models.}
Fine-tuning is a potent strategy for enhancing specific capabilities of pre-trained diffusion models \citep{gal2022TI, zhang2023adding, brooks2023instructpix2pix}. 
Personalized methods \citep{ruiz2023dreambooth, gal2022TI, chen2023disenbooth} utilize a small set of images to fine-tune the diffusion model, thereby generating images of the same object. 
Several methods for multi-concept image fine-tuning \citep{kumari2023multi,  xiao2023fastcomposer,avrahami2023bas,han2023svdiff,jin2023image} use cross-attention maps to align embeddings with individual concepts in the image.
Nevertheless, they do not consider the different variations in different image regions, which is important for industrial anomaly image generation.
Thus, we propose a separation and sharing fine-tuning strategy to model the different degrees of variations of anomalies and normal products, which independently learns products and anomalies on a shared U-Net.

\textbf{Mask Prediction with Generation Method.}
Previous methods on mask prediction for generated images are mainly based on features in GANs \citep{zhang2021datasetgan, li2022bigdatasetgan}. However, these approaches do not guarantee the generation of accurate masks for exceedingly small datasets.
Based on Stable Diffusion \citep{rombach2022high}, some recent methods, i.e., DiffuMask \citep{wu2023diffumask}, DatasetDM \citep{wu2023datasetdm} and DatasetDiffusion \citep{nguyen2024dataset}, produce masks by exploiting the potential of the cross-attention maps.
However, due to the low resolution of the cross-attention maps, they are directly interpolated to a higher resolution to match the image size without any auxiliary information, which leads to significant boundary uncertainty.
We incorporate the high-resolution features from the VAE decoder as auxiliary information for resolution retrieving, fusing them with U-Net decoder features, which are discriminative due to the modelling of the different variations in normal products and anomalies, to generate accurate high-resolution masks.
\section{Method}
\vspace{-5pt}
The training phase of the proposed Separation and Sharing (SeaS) Fine-tuning strategy is shown in Fig. \ref{pipeline}. 
In Sec. \ref{Sec:preliminaries}, we introduce the preliminaries of our approach.
In Sec. \ref{training}, we first design an Unbalanced Abnormal Text Prompt, 
which contains a set of tokens that characterize normal products and anomalies separately.
Subsequently, we propose the Decoupled Anomaly Alignment (DA) loss to bind anomaly image regions to anomaly tokens, 
and leverage Normal-image Alignment (NA) loss to empower normal token to express the globally-consistent normal product surface.
The two training processes are implemented separately for abnormal and normal images but on a shared U-Net architecture.
Then, based on the well-trained U-Net,
we design a Refined Mask Prediction branch to generate accurate masks corresponding to the generated anomaly images in Sec. \ref{mask_seg}.
Finally, we detail the generation of abnormal image-mask pairs and normal images in Sec. \ref{sec:inference}.

\vspace{-2pt}
\subsection{Preliminaries}
\label{Sec:preliminaries}
\vspace{-2pt}

\textbf{Stable Diffusion.}
Given an input image $x_0$, Stable Diffusion \citep{rombach2022high} firstly transforms $x_0$ into a latent space as $z = \varepsilon(x_0)$, 
and then adds a randomly sampled noise $\epsilon \sim N(0,\textbf{I})$ into $z$ as $\hat{z}_{t} = \alpha_{t} z + \beta_{t} \epsilon$, 
where $t$ is the randomly sampled timestep.
Then, the U-Net is employed to predict the noise $\epsilon$.
Let $c_\theta(\mathcal{P})$ be the CLIP text encoder that maps conditioning text prompt $\mathcal{P}$ into a conditioning vector $\textbf{e}$.
The training loss of Stable Diffusion can be stated as follows:
\begin{equation}
  \mathcal{L}_{\text{SD}}=\mathbb{E}_{z=\varepsilon(x_0),\mathcal{P},\epsilon \sim N(0,\textbf{I}), t} \Big[||\epsilon - \epsilon_{\theta}(\hat{z}_{t},t, \textbf{e})||_2^2\Big]
  \label{eq:Stable_Diff}
\end{equation}
where $\epsilon_{\theta}$ is the predicted noise.

\textbf{Cross-Attention Map in U-Net.}
Aiming to control the generation process,
the conditioning mechanism is implemented by calculating cross-attention between the conditioning vector $\textbf{e}\in \mathbb{R}^{Z \times C_1}$
and image features $\textbf{v} \in \mathbb{R}^{r \times r \times C_2}$ of the U-Net inner layers \citep{hertz2022prompt,chefer2023attend,xie2023boxdiff}.
The cross-attention map $A^{m,l} \in \mathbb{R}^{r \times r \times Z}$ can be calculated as:
\begin{equation}
  A^{m,l} = \text{softmax}(\frac{QK^{\top}}{\sqrt{d}}),  Q=\phi_q(\textbf{v}),  K=\phi_k(\textbf{e})
  \label{eq:crossattention}
\end{equation}
where $Q \in \mathbb{R}^{r \times r \times C} $ denotes a query projected by a linear layer $\phi_q$ from $\textbf{v}$, 
$r$ is the resolution of the feature map in U-Net, 
and $l$ is the index of the U-Net inner layer.
$K \in \mathbb{R}^{Z \times C} $ denotes a key through another linear layer $\phi_k$ from $\textbf{e}$, 
and $Z$ is the number of text embeddings after padding.

\vspace{-2pt}
\subsection{Separation and Sharing Fine-tuning}
\label{training}
\vspace{-2pt}

\begin{figure}[!t]
\centering
\includegraphics[width=1\linewidth]{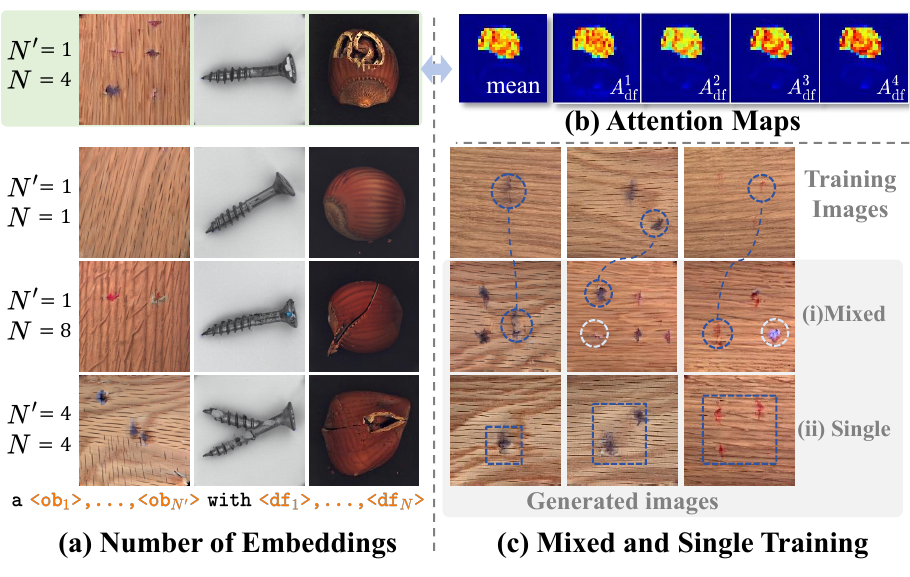}
\vspace{-0.8cm}
\caption{  \ (a) Generated images with the different number of tokens.
(b) Cross-attention maps.
(c) Examples of diverse generated images. 
} 
\vspace{-0.6cm}
\label{Fig:Method_fig}
\end{figure}

\textbf{Unbalanced Abnormal Text Prompt.} 
Through the experimental observation, 
we found that the typical text prompt, like \texttt{a photo of a bottle with defect} \citep{CVPR2023winclip}, or \texttt{damaged bottle} \citep{iclr2024anomalyclip}, is suboptimal for industrial anomaly generation.
The balanced semantic words for normal products and anomalies may fail to capture their differential variation degrees.
Therefore, we design the Unbalanced Abnormal (UA) Text Prompt for each anomaly type of each product, i.e.,

\centerline{$\mathcal{P}$ = \texttt{a <ob> with} \texttt{<}$\texttt{df}_{1}$\texttt{>,}\texttt{<}$\texttt{df}_{2}$\texttt{>} \texttt{,...,} \texttt{<}$\texttt{df}_{N}$\texttt{>}}
\noindent
where \texttt{<ob>} and \texttt{<}$\texttt{df}_{n}$\texttt{>} ($n \in \{1,2,..., N\}$) are the tokens of the industrial normal products (short for Normal Token) and the anomalies (short for Anomaly Token) respectively.
We use a set of $N$ Anomaly Tokens for each anomaly type, with different sets corresponding to different anomaly types.
As shown in Fig. \ref{Fig:Method_fig},
in SeaS, we separately employ normal images to train the embedding corresponding to \texttt{<ob>}, and abnormal images to train the embeddings corresponding to \texttt{<}$\texttt{df}_{n}$\texttt{>}.
Experimental observations indicate that one \texttt{<ob>} is sufficient to express normal product,
while multiple \texttt{<}$\texttt{df}_{n}$\texttt{>} are necessary for controlling the generation of anomalies.
As shown in Fig. \ref{Fig:Method_fig}(a),
when we use the UA prompt $\mathcal{P}$ (the dotted green box in (a)),
the cross-attention maps in (b) show that different tokens have different responses in the abnormal regions, 
which indicates that they focus on different attributes of the anomalies, 
and performing the average operation on the cross-attention maps produces never-seen anomalies.
When we use only one \texttt{<}$\texttt{df}$\texttt{>}, 
it is difficult to align it to several different anomalies that belong to the same category. 
Therefore, during inference, if the denoised anomaly feature has a larger distance to \texttt{<}$\texttt{df}$\texttt{>}, it will be assigned a smaller response by the U-Net, 
which leads to the ``anomaly missing" phenomenon,
e.g., the generated images in the case of ($N'=1, N=1$).
In addition, if we utilize a large number of \texttt{<}$\texttt{df}_{n}$\texttt{>}, 
we find that each \texttt{<}$\texttt{df}_{n}$\texttt{>} may focus on some local properties of an anomaly, such a case increases the diversity but may reduce the authenticity of the anomalies, as shown in the case $N'=1, N=8$. 
Similarly, if we use multiple learnable \texttt{<ob>}, e.g., $N'=4, N=4$, each \texttt{<ob>} pays attention to the local character of the normal product, which may reduce the 
global consistency of the normal product.

\begin{figure*}[t]
\centering
\includegraphics[width=1\linewidth]{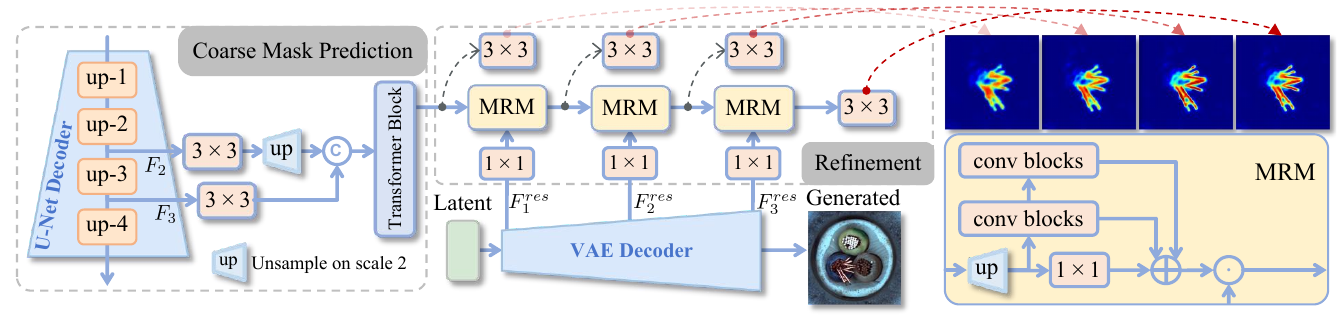}
\vspace{-0.7cm}
\caption{The \textbf{Refined Mask Prediction (RMP)} branch during inference. The Coarse Feature Extraction utilizes features from the up-2 and up-3 layers of the U-Net Decoder to extract coarse features. The cascaded Mask Refinement Module (MRM) further obtains the mask accurately aligned with the anomaly with the assistance of high-resolution features of the VAE Decoder.}
\vspace{-0.6cm}
\label{Fig:Mask}
\end{figure*}

\textbf{Decoupled Anomaly Alignment.}
Given a few abnormal images $x_{\text{df}}$ and their corresponding masks, 
we aim to align the anomaly tokens \texttt{<}$\texttt{df}_{n}$\texttt{>} to the anomaly area of $x_{\text{df}}$ by tuning the U-Net and the learnable embedding corresponding to \texttt{<}$\texttt{df}_{n}$\texttt{>}.
Therefore, we propose the Decoupled Anomaly Alignment (DA) loss, i.e.,
\vspace{-0.2cm}
\begin{equation}
      \mathcal{L}_{\textbf{DA}}=\sum_{l=1}^L(||\frac{1}{N}\sum_{n=1}^NA_{\text{df}}^{n,l}-M^l||^2 + ||A_{\text{ob}}^{l}\odot M^l||^2)
      \label{eq:decoupleloss}
\end{equation}
where $A_{\text{df}}^{n,l} \in \mathbb{R}^{r \times r \times 1}$ is the cross-attention map corresponding to the $n$-th anomaly token \texttt{<}$\texttt{df}_{n}$\texttt{>},
$N$ is the number of anomaly token in $\mathcal{P}$.
$L$ is the total number of U-Net layers used in alignment.
$M^{l}$ is the binary mask with $r \times r$ resolution, 
where the abnormal area is 1 and the background is 0.
$A^{l}_{\text{ob} }\in \mathbb{R}^{r \times r \times 1}$ is the cross-attention map corresponding to the normal token \texttt{<ob>}, 
$\odot$ is the element-wise product. 
DA loss performs the mandatory decoupling of the anomaly and the normal product.
The first term of DA loss is to align the abnormal area to \texttt{<}$\texttt{df}_{n}$\texttt{>} according to the mask $M^{l}$.
The second term of DA loss reduces the response value of $A^{l}_{\text{ob}}$ in the abnormal area, which prevents \texttt{<ob>} from aligning to the abnormal area of $x_{\text{df}}$. 
Further analysis of how the DA loss ensures the diversity of anomalies is provided in Appendix \ref{appendix:DAloss}. 
 Therefore, the total loss for the anomaly image $x_{\text{df}}$ is:

\vspace{-0.3cm}
\begin{equation}
\mathcal{L}_{\textbf{df}} = \mathcal{L}_{\textbf{DA}} +||\epsilon_{\text{df}} - \epsilon_{\theta}(\hat{z}_{\text{df}},t_{\text{df}}, \textbf{e}_{\text{df}})||_2^2  \label{eq:l_de} 
\end{equation} 
In second term of Eq. \ref{eq:l_de}, 
we use random noises $\epsilon_{\text{df}}$ and timesteps $t_{\text{df}}$ to perform forward diffusion on abnormal images $x_{\text{df}}$, then obtain the noisy latent $\hat{z}_{\text{df}}$.
The conditioning vector $\textbf{e}_{\text{df}} \in \mathbb{R}^{Z \times C_1} $ is used to guide the U-Net in predicting noise, and then calculate the loss with the noise $\epsilon_{\text{df}}$.

\textbf{Normal-image Alignment.} 
As we discussed, 
increasing the number of the normal token \texttt{<ob>} leads to a higher diversity, 
while it may reduce the authenticity of the generated normal image and destruct global consistency.
However, aligning only one \texttt{<ob>} to a few of the training images may suffer from the issue of overfitting.
Therefore, we add a Normal-image Alignment (NA) loss to overcome such a dilemma, which is stated as follows,
\begin{equation}
  \mathcal{L}_{\textbf{ob}} = ||\epsilon_{\text{ob}} - \epsilon_{\theta}(\hat{z}_{\text{ob}},t_{\text{ob}},\textbf{e}_{\text{ob}})||_2^2
  \label{eq:important}
\end{equation}

Instead of aligning the normal region of $x_{\text{df}}$ to \texttt{<ob>}, 
in calculating the NA loss, we use random noises $\epsilon_{\text{ob}}$ and timesteps $t_{\text{ob}}$ to perform forward diffusion on the normal images $x_{\text{ob}}$. 
Then the noisy latent $\hat{z}_{ob}$ and the embedding $\textbf{e}_{\text{ob}}$ corresponding to the normal tokens of $\mathcal{P}$, i.e., ``a \texttt{<ob>}", are input into the U-Net in predicting noise, and then calculate the NA loss with $\epsilon_{\text{ob}}$.

\textbf{Mixed Training.} 
Based on the separated DA loss for abnormal images and NA loss for the normal ones, 
the objective of Separation and Sharing Fine-tuning is formed as:
\begin{equation}
  \mathcal{L}=\mathcal{L}_{\textbf{df}} + \mathcal{L}_{\text{\textbf{ob}}}
  \label{eq:important}
\end{equation}
In the training process, 
instead of training a single U-Net model for each anomaly type, 
we train a unified U-Net model for each product. 
Specifically, 
given a product image set,
which contains $G$ anomaly categories of masked abnormal images and some normal images.
We group all the abnormal images of a product into a unified set $X_\text{df} = \{x^{1}_\text{df},x^{2}_\text{df}, .., x^{H}_\text{df}\}$.
For each anomaly type, we use $\mathcal{P}$ with different sets of anomaly tokens.
In addition, we sample a fixed number of normal images to consist of the normal training set $X_\text{ob} = \{x^{1}_\text{ob},x^{2}_\text{ob}, .., x^{P}_\text{ob}\}$.
During each step of our fine-tuning process,
we sample same number of images from both $X_\text{df}$ and $X_\text{ob}$, 
and mixed them into a batch.
We found that such a mixed training strategy not only alleviates the overfitting caused by the limited number of each anomaly type, but also increases the diversity of the anomaly image, while still maintaining reasonable authenticity, as shown in Fig. \ref{Fig:Method_fig}(c), 
(i) indicates that the model with mixed training may generate new anomalies, e.g., the anomalies inside the dotted white line. In contrast, the anomalies in (ii) overfit the training images. More ablation studies on the mixed training strategy are shown in Tab. \ref{tab:appendix_mix} in appendix \ref{appendix:ablation}.

\vspace{-2pt}
\subsection{Refined Mask Prediction}
\label{mask_seg}
\vspace{-2pt}
The design of the separated and shared approach enables U-Net to simultaneously model the different degrees of variations in both normal products and anomalies, representing the discriminative features for mask prediction.
To further obtain pixel-accurate masks, we design a cascaded Refined Mask Prediction (RMP) branch, which is grafted onto the U-Net trained within SeaS (mentioned in Sec. \ref{training}).
As shown in Fig. \ref{Fig:Mask},
RMP consists of two steps,
firstly capturing discriminative features from U-Net and secondly combining them with high-resolution features of VAE decoder to generate anomaly-matched masks.

\textbf{Coarse Feature Extraction.} 
The first step aims to extract a coarse but highly-discriminative feature for anomalies from the U-Net decoder.
Specifically,
let $F_1 \in \mathbb{R}^{32 \times 32 \times 1280  }$ and $F_2 \in \mathbb{R}^{ 64 \times 64 \times 640 }$ denote the output feature of ``\texttt{up-2}'' and ``\texttt{up-3}'' layers of the decoder in U-Net, respectively.
We first leverage a $1 \times 1$ convolution block to compress the channel of $F_1$ and $F_2$ to $\overline{F}_1 \in \mathbb{R}^{32 \times 32 \times 128  }$ and $\overline{F}_2 \in \mathbb{R}^{ 64 \times 64 \times 64}$, respectively.
Then, we upsample $\overline{F}_1$ to $64\times64$ resolution and concatenate it with $\overline{F}_2$.
Finally, four transformer layers are employed to fuse the concatenated features and obtain a unified coarse feature $\hat{F} \in \mathbb{R}^{ 64 \times 64 \times 192 }$.

\textbf{Mask Refinement Module.}
Directly upsampling the coarse feature $\hat{F}$ to high resolution will result in a loss of anomaly details.
Therefore, we design the Mask Refinement Module (MRM) to refine the coarse feature $\hat{F}$ in a progressive manner.
As shown in Fig. \ref{Fig:Mask},
each MRM takes in two features,
i.e., the high-resolution features from VAE and the discriminative feature to be refined.
Firstly, the discriminative feature is upsampled to align with the high-resolution ones of VAE.
To preserve the discriminative ability, the upsampled feature is processed through two chained convolution blocks for capturing multi-scale anomaly features and a $1\times1$ convolution for capturing local features.
These features are then summed
and multiplied with the VAE features element-wisely to enhance the anomalies' boundary.
Finally, MRM employs a $3\times3$ convolution to fuse the added features and outputs a refined feature.

To refine $\hat{F}$, we employ three MRMs positioned in sequence.
Each MRM takes the previous MRM’s output as the discriminant feature to be refined,
while the first MRM takes $\hat{F}$ as the discriminative input.
For another input of each MRM,
we use the outputs from the 1-st, 2-nd, and 3-rd ``\texttt{up-blocks}'' of the VAE decoder respectively.
In this way,
the features obtained by the last MRM have the advantages of both high resolution and high discriminability.
Finally, we use a $3 \times 3$ convolution and a softmax to generate the refined anomaly mask $\hat{M}_\text{df}^\prime \in \mathbb{R}^{ 512 \times 512 \times 2 }$ using the output of the last MRM.

\textbf{Loss Functions.}
During training, we use $x_{\text{df}}$ and $x_{\text{ob}}$ as inputs. For $x_{\text{df}}$, we obtain the coarse mask $\hat{M}_\text{df} \in \mathbb{R}^{ 64 \times 64 \times 2 }$ from the Coarse Feature Extraction and $\hat{M}^\prime_\text{df}$ after the MRMs. Similarly, for $x_{\text{ob}}$, we obtain the $\hat{M}_\text{ob} \in \mathbb{R}^{64 \times 64 \times 2 }$ from Coarse Feature Extraction and directly upsample it to the original resolution, denoted as $\hat{M}^\prime_\text{ob} \in \mathbb{R}^{512 \times 512 \times 2}$.
Then we conduct the supervision on both low-resolution and high-resolution predictions as, 

\begin{equation}
\begin{aligned}
\mathcal{L}_\mathcal{M} =& \mathcal{F} (\hat{M}_\text{df} , \mathbf{M}_\text{df}) +
                        \mathcal{F} (\hat{M}_\text{ob} , \mathbf{M}_\text{ob}) +  \\
                        &\mathcal{F} (\hat{M}_\text{df}^\prime , \mathbf{M}_\text{df}^\prime) + 
                        \mathcal{F} (\hat{M}_\text{ob}^\prime , \mathbf{M}_\text{ob}^\prime)
\end{aligned}
\label{eq:maskloss2}
\end{equation}
where $\mathcal{F}$ indicates the Focal Loss \citep{lin2017focal}. 
$\mathbf{M}_\text{ob} \in \mathbb{R}^{ 64 \times 64 \times 1 }$ and
$\mathbf{M}_\text{ob}^\prime \in \mathbb{R}^{ 512 \times 512 \times 1 }$ are used to suppress noise in normal images, with each pixel value set to 0.
$\mathbf{M}_\text{df} \in \mathbb{R}^{ 64 \times 64 \times 1 }$ and
$\mathbf{M}_\text{df}^\prime \in \mathbb{R}^{ 512 \times 512 \times 1 }$ are the ground truth masks of abnormal images. 
More ablation studies on the effect of normal images in training RMP branch are shown in Tab. \ref{tab:appendix_normal} and Fig. \ref{fig:appendix_normal} in appendix \ref{appendix:ablation}.

\vspace{-2pt}
\subsection{Inference} 
\label{sec:inference}
\vspace{-2pt}
During the generation,
aiming further to ensure the global consistency of the normal products,
we random select a normal image \(x_{\text{ob}}\) from $X_{\text{ob}}$ as input, 
and add random noise to \(x_{\text{ob}}\), which resulting in an initial noisy latent \(\hat{z_0}\).  
Next, for the generation of abnormal images, \(\hat{z_0}\) is input into the U-Net for noise prediction, with the process guided by the conditioning vector \(\textbf{e}_{\text{df}}\) (mentioned in Eq. \ref{eq:l_de}), which is corresponding to the whole UA Text Prompt $\mathcal{P}$. For generating normal images to further enhance unsupervised AD methods, we use the conditioning vector \(\textbf{e}_{\text{ob}}\) corresponding to the normal tokens of $\mathcal{P}$.
Regarding the masks corresponding to anomalies, in the final three denoising steps, 
the RMP branch (Sec. \ref{mask_seg}) leverages the features from the U-Net decoder and VAE decoder to generate the final anomaly mask. Specifically, we average the refined anomaly mask from these steps to obtain the refined mask $\hat{M}_\text{df}^\prime \in \mathbb{R}^{ 512 \times 512 \times 2 }$.
Then we take the threshold $\tau$ for the second channel of $\hat{M}_\text{df}^\prime$ to segment the final anomaly mask $M_\text{df} \in \mathbb{R}^{ 512 \times 512 \times 1 }$. The effect of $\tau$ on the downstream supervised segmentation models is shown in Tab. \ref{tab:appendix_threshold} in appendix \ref{appendix:ablation}.
In the last denoising step, the output of the generation model is used as the generated abnormal image.
\section{Experiments}
\label{sec：exp}

\begin{table*}[]
\caption{Comparison on IS and IC-LPIPS on MVTec AD, VisA, and MVTec AD 3D.
Bold indicates the best performance.}
\vspace{-0.55cm}
\label{table:generation}
\begin{center}
\setlength{\tabcolsep}{3pt}
\resizebox{0.83\linewidth}{!}{
\begin{tabular}{c|cccc|cccc|cccc}
\hline
\multirow{2}{*}{\textbf{Methods}} & \multicolumn{4}{c|}{\textbf{MVTec AD}}                               & \multicolumn{4}{c|}{\textbf{VisA}}                                   & \multicolumn{4}{c}{\textbf{MVTec 3D AD}}   \\ \cline{2-13}
& \textbf{IS$\uparrow$}   & \textbf{IC-LPIPS$\uparrow$} & \textbf{KID$\downarrow$} & \textbf{IC-LPIPS(a)$\uparrow$} & \textbf{IS$\uparrow$}   & \textbf{IC-LPIPS$\uparrow$} & \textbf{KID$\downarrow$} & \textbf{IC-LPIPS(a)$\uparrow$} & \textbf{IS$\uparrow$}   & \textbf{IC-LPIPS$\uparrow$} & \textbf{KID$\downarrow$} & \textbf{IC-LPIPS(a)$\uparrow$} \\ \hline \hline
Crop\&Paste\cite{ICME2021crop-and-paste} & 1.51          & 0.14          & -            & -                     & -             & -             & -            & -                     & -             & -             & -            & -                     \\
SDGAN\cite{niu2020sdgan} & 1.71          & 0.13          & -            & -                     & -             & -             & -            & -                     & -             & -             & -            & -                     \\
Defect-GAN\cite{zhang2021defectgan} & 1.69          & 0.15          & -            & -                     & -             & -             & -            & -                     & -             & -             & -            & -                     \\
DFMGAN\cite{Duan2023DFMGAN} & 1.72          & 0.20          & 0.12 &     0.14 & 1.25          & 0.25          & 0.24  & 0.05        & 1.80          & 0.29          & 0.19 &   0.08  \\
AnomalyDiffusion\cite{hu2023anomalydiffusion}  & 1.80          & 0.32          & -      & 0.12  & 1.26          & 0.25          &    -  & 0.04 & 1.61          & 0.22          &  -  &  0.07  \\ \hline
\textbf{SeaS}                              & \textbf{1.88} & \textbf{0.34} &\textbf{0.04}       &   \textbf{0.18} & \textbf{1.27} & \textbf{0.26} & \textbf{0.02}  & \textbf{0.06} & \textbf{1.95} & \textbf{0.30} &\textbf{0.06} &  \textbf{0.09} \\ \hline
\end{tabular}}
\vspace{-0.55cm}
\end{center}
\end{table*}

\vspace{-2pt}
\subsection{Experimental Settings}
\label{sec:expsettings}
\vspace{-2pt}

\noindent\textbf{Implementation Details.}
We train SeaS by fine-tuning the pre-trained Stable Diffusion v1-4 \citep{rombach2022high}.
In anomaly image generation experiments,
we use $60$ normal images and $\frac{1}{3}$ masked anomaly images for each anomaly type in training. We train one generative model per product, covering all anomaly types.
During inference, we generate $1,000$ anomaly image-mask pairs for a single anomaly type.
More details are given in appendix \ref{appendix:implementation}.

\begin{figure}[t]
\centering
\includegraphics[width=1\linewidth]{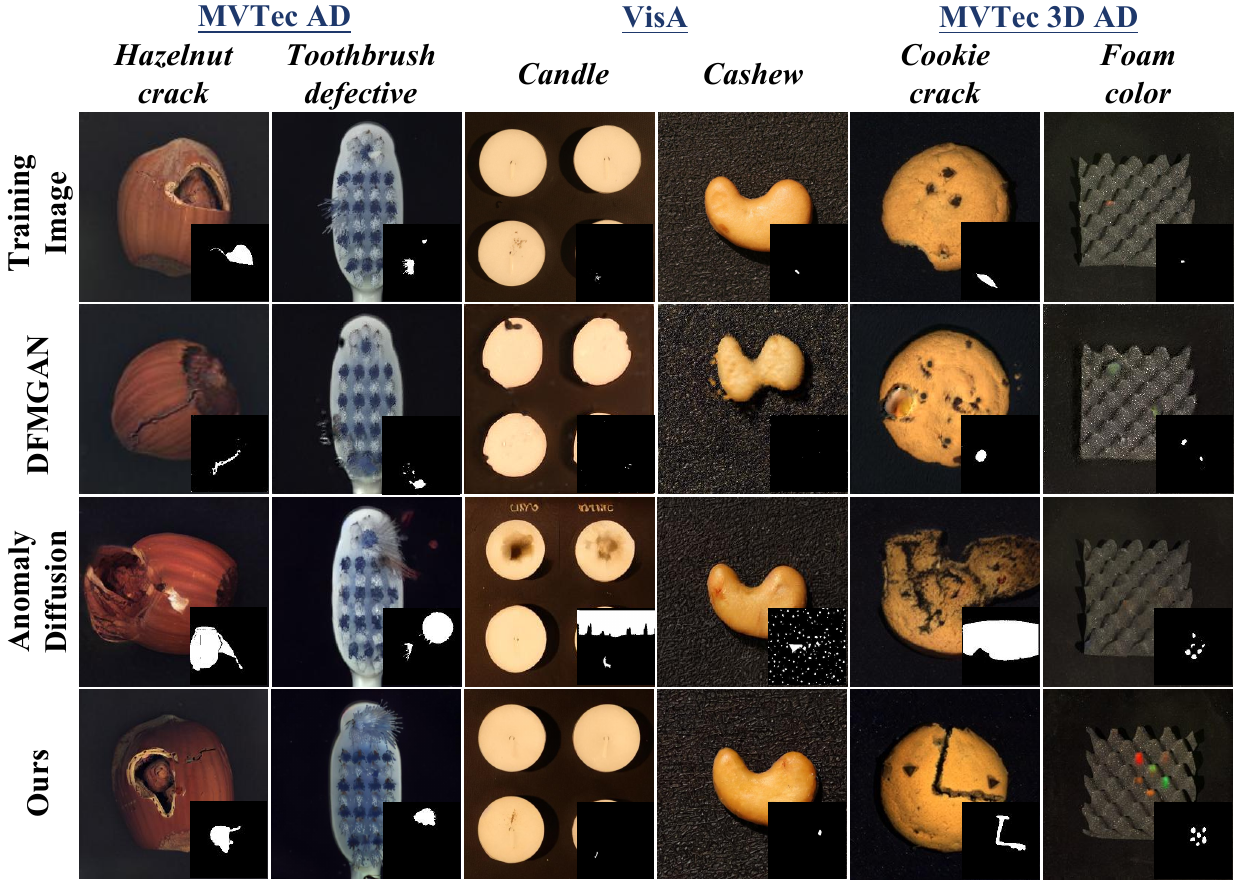}
\vspace{-18pt}
\caption{Visualization of the generation results on MVTec AD, VisA and MVTec 3D AD. The sub-image in the lower right corner is the generated mask, none means that the method cannot generate masks.}
\vspace{-13pt}
\label{fig:generation}
\end{figure}

\begin{figure}[t]
\centering
\includegraphics[width=1\linewidth]{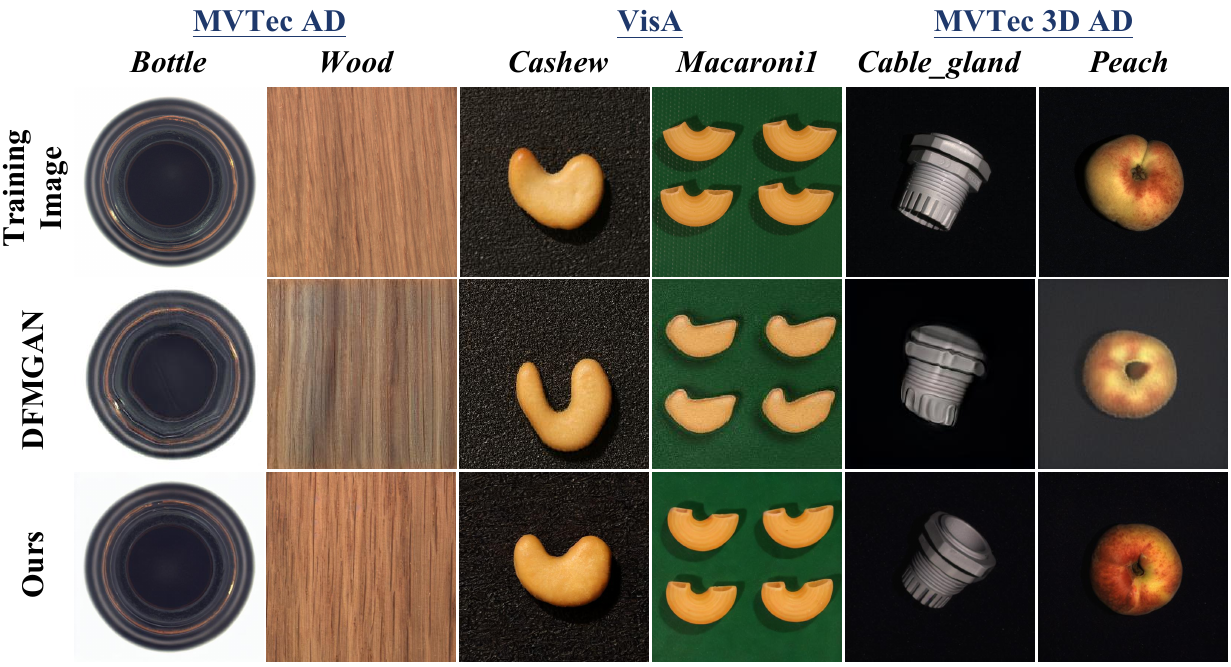}
\vspace{-18pt}
\caption{Visualization of the generated normal images on MVTec AD, VisA and MVTec 3D AD.}
\vspace{-19pt}
\label{fig:generation2}
\end{figure}

\noindent\textbf{Datasets.} 
We conduct experiments on MVTec AD dataset \citep{bergmann2019mvtec},
VisA dataset\cite{zou2022visa},
and MVTec 3D AD dataset (only RGB images) \citep{bergmannmvtec3d}.
MVTec AD dataset contains $15$ product categories,
each with up to $8$ different anomalies.
VisA dataset covers $12$ objects in $3$ domains. 
MVTec 3D AD dataset includes $10$ product categories,
each with up to $4$ different anomalies.
It contains more challenges,
i.e., lighting condition variations, and product pose variations.

\begin{table*}[!htb]
\caption{Comparison on combining generated anomalies with synthesis-based anomaly detection methods across multiple datasets.}
\vspace{-0.55cm}
\label{table:combined_replace}
\renewcommand{\arraystretch}{1.05}
\begin{center}
\setlength{\tabcolsep}{2pt}
\resizebox{1\linewidth}{!}{
\begin{tabular}{c|ccc|cccc|ccc|cccc|ccc|cccc}  
    \toprule
    \multirow{3}{*}{\makecell{\textbf{Segmentation}\\\textbf{Models}}} & \multicolumn{7}{c|}{\textbf{MVTec AD}} & \multicolumn{7}{c|}{\textbf{VisA}} & \multicolumn{7}{c}{\textbf{MVTec 3D AD}} \\
    \cmidrule{2-22}  
    & \multicolumn{3}{c|}{\textbf{Image-level}} & \multicolumn{4}{c|}{\textbf{Pixel-level}} & \multicolumn{3}{c|}{\textbf{Image-level}} & \multicolumn{4}{c|}{\textbf{Pixel-level}} & \multicolumn{3}{c|}{\textbf{Image-level}} & \multicolumn{4}{c}{\textbf{Pixel-level}} \\
    & \textbf{AUROC} & \textbf{AP} & \textbf{$F_1$-max} & \textbf{AUROC} & \textbf{AP} & \textbf{$F_1$-max} & \textbf{IoU} & \textbf{AUROC} & \textbf{AP} & \textbf{$F_1$-max} & \textbf{AUROC} & \textbf{AP} & \textbf{$F_1$-max} & \textbf{IoU} & \textbf{AUROC} & \textbf{AP} & \textbf{$F_1$-max} & \textbf{AUROC} & \textbf{AP} & \textbf{$F_1$-max} & \textbf{IoU} \\
    \midrule
    \midrule
    DRAEM \citep{zavrtanik2021draem} & 98.00 & 98.45 & 96.34 & 97.90 & 67.89 & 66.04 & \textbf{60.30} & 86.28&85.30&  81.66  &  92.92   &17.15&22.95& 13.57 &79.16&90.90 & 89.78 & 86.73&  14.02& 17.00& 12.42  \\
    DRAEM + SeaS & \textbf{98.64} & \textbf{99.40} & \textbf{97.89} & \textbf{98.11} & \textbf{76.55} & \textbf{72.70} &  58.87 &  \textbf{88.12}   &  \textbf{87.04}   &   \textbf{83.04}   &  \textbf{98.45}   & \textbf{49.05}   & \textbf{48.62}   &  \textbf{35.00} &  \textbf{85.45}   &  \textbf{93.58}  &  \textbf{90.85}  &   \textbf{95.43}  &  \textbf{20.09}   &  \textbf{26.10}  &   \textbf{17.07}   \\
    \midrule
    GLASS \citep{chen2024unifiedglass} & 99.92 & 99.98 & 99.60 & 99.27 & 74.09 & 70.42 & 57.14 &   97.68   &  96.89  &   93.03   &   \textbf{98.47}   &  45.58   &   48.39   &   39.92  &  92.34  & 96.85  & \textbf{93.37}  & 98.46 & 48.46 & 49.13 & 45.03 \\
    GLASS + SeaS & \textbf{99.97} & \textbf{99.99} & \textbf{99.81} & \textbf{99.29} & \textbf{76.82} & \textbf{72.38} & \textbf{57.45} &  \textbf{97.88}  &  \textbf{97.39}  & \textbf{93.21}  & 98.43 & \textbf{48.06} & \textbf{49.32} & \textbf{40.00} & \textbf{92.95} & \textbf{97.38} & 93.21 & \textbf{98.73} & \textbf{48.55} & \textbf{49.28} & \textbf{46.02} \\
    \midrule
    Average \citep{zavrtanik2021draem,chen2024unifiedglass}& 98.96 & 99.22 & 97.97 & 98.59 & 70.99 & 68.23 & \textbf{58.72} &   91.98   &  91.10  &   87.35   &   95.70  &  31.37   &   35.67   &  26.75 &  85.75  & 93.88  & 91.58  & 92.60 & 31.24 & 33.07 & 28.73 \\
    \textbf{Average(+ SeaS)} & \textbf{99.31} & \textbf{99.70} & \textbf{98.85} & \textbf{98.70} & \textbf{76.69} & \textbf{72.54} & 58.16 &  \textbf{93.00}  &  \textbf{92.22}  & \textbf{88.13}  & \textbf{98.44} & \textbf{48.56} & \textbf{48.97} & \textbf{37.50} & \textbf{89.20} & \textbf{95.48} & \textbf{92.03} & \textbf{97.08} & \textbf{34.32} & \textbf{37.69} & \textbf{31.55} \\
    
    \bottomrule
\end{tabular}
}
\vspace{-0.55cm}
\end{center}
\end{table*}

\begin{table*}[!htb]
\caption{Comparison on combining generated normal images with unsupervised anomaly detection methods across multiple datasets.}
\vspace{-0.55cm}
\label{table:combined_unad}
\renewcommand{\arraystretch}{1.05}
\begin{center}
\setlength{\tabcolsep}{2pt}
\resizebox{1\linewidth}{!}{
\begin{tabular}{c|ccc|cccc|ccc|cccc|ccc|cccc}  
    \toprule
    \multirow{3}{*}{\makecell{\textbf{Segmentation}\\\textbf{Models}}} & \multicolumn{7}{c|}{\textbf{MVTec AD}} & \multicolumn{7}{c|}{\textbf{VisA}} & \multicolumn{7}{c}{\textbf{MVTec 3D AD}} \\
    \cmidrule{2-22}  
    & \multicolumn{3}{c|}{\textbf{Image-level}} & \multicolumn{4}{c|}{\textbf{Pixel-level}} & \multicolumn{3}{c|}{\textbf{Image-level}} & \multicolumn{4}{c|}{\textbf{Pixel-level}} & \multicolumn{3}{c|}{\textbf{Image-level}} & \multicolumn{4}{c}{\textbf{Pixel-level}} \\
    & \textbf{AUROC} & \textbf{AP} & \textbf{$F_1$-max} & \textbf{AUROC} & \textbf{AP} & \textbf{$F_1$-max} & \textbf{IoU} & \textbf{AUROC} & \textbf{AP} & \textbf{$F_1$-max} & \textbf{AUROC} & \textbf{AP} & \textbf{$F_1$-max} & \textbf{IoU} & \textbf{AUROC} & \textbf{AP} & \textbf{$F_1$-max} & \textbf{AUROC} & \textbf{AP} & \textbf{$F_1$-max} & \textbf{IoU} \\
    \midrule
    \midrule
    HVQ-Trans \citep{lu2023hvqtrans} & 96.38 & 98.09 & 95.30 & \textbf{97.60} & 47.95 & 53.32 & \textbf{45.03} & 90.11 & 88.18 & 84.08  & 98.10  & 28.67 & 35.05 & \textbf{24.03} &68.15 &84.38  & 85.20 & 96.40 & 17.23 & 24.59 & \textbf{20.51}  \\
     HVQ-Trans + SeaS & \textbf{97.25} & \textbf{98.48} & \textbf{95.78} & 97.58 & \textbf{48.53} & \textbf{53.84} & 44.61 &   \textbf{92.12}   &   \textbf{90.35}  &  \textbf{86.23}   &   \textbf{98.15}  &  \textbf{29.52} & \textbf{36.00} & 23.60 & \textbf{71.26} & \textbf{90.35} & \textbf{89.23} & \textbf{96.56} & \textbf{19.34} & \textbf{26.40} & 20.47 \\
    \midrule
    PatchCore \citep{roth2022patchcore}  & 98.63 & 99.47 & 98.18 & \textbf{98.37} & 56.13 & 58.83 & 49.45 &   94.84   &  95.98   & 91.69   & 98.38  &  48.58   &  49.69   & 42.44 & 83.44 & 94.89 & 92.24   & 98.55 &   34.52 &  39.09   & 39.29 \\
    PatchCore + SeaS & \textbf{98.64} & \textbf{99.48} & \textbf{98.22} & \textbf{98.37} & \textbf{63.98} & \textbf{64.07} & \textbf{55.43} &   \textbf{94.97}  &   \textbf{96.06}   & \textbf{91.81} &  \textbf{98.41} & \textbf{48.60} &  \textbf{49.72}  & \textbf{42.46}  &  \textbf{83.88}   & \textbf{94.97}   &  \textbf{92.32}   & \textbf{98.56} &  \textbf{34.65}  &  \textbf{39.41}  &   \textbf{39.43}   \\
    \midrule
    MambaAD \citep{he2025mambaad}  & 98.54 &99.52 & 97.77 & \textbf{97.67} & 56.23 & 59.34 & 51.31&  94.19 &  94.44  & 89.55 &   98.49  &  39.27&  \textbf{44.18}   &  \textbf{37.68} &  85.92   &   95.69  & 92.51 & 98.57 & \textbf{37.30} & \textbf{41.08} & 39.44  \\
    MambaAD + SeaS & \textbf{98.80} & \textbf{99.64} & \textbf{98.40} & 97.66 & \textbf{56.86} & \textbf{59.70} & \textbf{51.51} &   \textbf{94.23}   &   \textbf{94.65}   &   \textbf{89.93}   &   \textbf{98.70}   &  \textbf{39.33}   &   43.99   &  36.62  &  \textbf{88.67}  &  \textbf{96.60}  &   \textbf{93.41}   &   \textbf{98.74}  &  35.46  &  39.59  & \textbf{39.51}   \\
        \midrule
    Average \citep{lu2023hvqtrans,roth2022patchcore,he2025mambaad} & 97.85 &99.03 & 97.08 & \textbf{97.88} & 53.44 & 57.16 & 48.60&  93.05 &  92.87  & 88.44 &   98.32  &  38.84&  42.97   &  \textbf{34.72} &  79.17   &  91.65& 89.98 & 97.84 & 29.68 & 34.92 & 33.08  \\
    \textbf{Average(+ SeaS)} & \textbf{98.23} & \textbf{99.20} & \textbf{97.47} & 97.87 & \textbf{56.46} & \textbf{59.20} & \textbf{50.52} &   \textbf{93.77}   &   \textbf{93.69}   &   \textbf{89.32}   &   \textbf{98.42}   &  \textbf{39.15}   &  \textbf{43.24}   &  34.23  &  \textbf{81.27}  &  \textbf{93.97}  &   \textbf{91.65}   &   \textbf{97.95}  &  \textbf{29.82}  &  \textbf{35.13}  & \textbf{33.14}   \\
    \bottomrule
\end{tabular}
}
\vspace{-0.55cm}
\end{center}
\end{table*}

\begin{table*}[!htb]
\caption{Comparison on trained supervised segmentation models for anomaly detection and segmentation across multiple datasets.}
\vspace{-0.55cm}
\label{table:seg_combined}
\renewcommand{\arraystretch}{1.05}
\begin{center}
\setlength{\tabcolsep}{2pt}
\resizebox{1\linewidth}{!}{
\begin{tabular}{cc|ccc|cccc|ccc|cccc|ccc|cccc}
    \toprule

    \multirow{3}{*}{\makecell{\textbf{Segmentation}\\\textbf{Models}}}&\multirow{3}{*}{\makecell{\textbf{Generative}\\\textbf{Models}}} & \multicolumn{7}{c|}{\textbf{MVTec AD}} & \multicolumn{7}{c|}{\textbf{VisA}} & \multicolumn{7}{c}{\textbf{MVTec 3D AD}} \\
    \cmidrule{3-23}
    & & \multicolumn{3}{c|}{\textbf{Image-level}} & \multicolumn{4}{c|}{\textbf{Pixel-level}} & \multicolumn{3}{c|}{\textbf{Image-level}} & \multicolumn{4}{c|}{\textbf{Pixel-level}} & \multicolumn{3}{c|}{\textbf{Image-level}} & \multicolumn{4}{c}{\textbf{Pixel-level}} \\
    & & \textbf{AUROC} & \textbf{AP} & \textbf{$F_1$-max} & \textbf{AUROC} & \textbf{AP} & \textbf{$F_1$-max} & \textbf{IoU} & \textbf{AUROC} & \textbf{AP} & \textbf{$F_1$-max} & \textbf{AUROC} & \textbf{AP} & \textbf{$F_1$-max} & \textbf{IoU} & \textbf{AUROC} & \textbf{AP} & \textbf{$F_1$-max} & \textbf{AUROC} & \textbf{AP} & \textbf{$F_1$-max} & \textbf{IoU} \\
    \midrule
    \midrule
   \multirow{4}{*}{\makecell{BiSeNet V2\\\citep{yu2021bisenet}}} 
   & DFMGAN & 90.90 & 94.43 & 90.33 & 94.57 & 60.42 & 60.54 & 45.83 & 63.07 & 62.63 & 66.48 & 75.91 & 9.17 & 15.00 & 9.66 & 61.88 & 81.80 & 84.44 & 75.89 & 15.02 & 21.73 & 15.68 \\
    & AnomalyDiffusion & 90.08 & 94.84 & 91.84 & 96.27 & 64.50 & 62.27 & 42.89 & 76.11 & 77.74 & 73.13 & 89.29 & 34.16 & 37.93 & 15.93 & 61.49 & 81.35 & 85.36 & \textbf{92.39} & 15.15 & 20.09 & 14.70 \\
    & \textbf{SeaS} & \textbf{96.00} & \textbf{98.14} & \textbf{95.43} & \textbf{97.21} & \textbf{69.21} & \textbf{66.37} & \textbf{55.28} & \textbf{85.61} & \textbf{86.64} & \textbf{80.49} & \textbf{96.03} & \textbf{42.80} & \textbf{45.41} & \textbf{25.93} & \textbf{73.60} & \textbf{87.75} & \textbf{85.82} & 90.41 & \textbf{26.04} & \textbf{32.61} & \textbf{28.55} \\
    \midrule
   \multirow{4}{*}{\makecell{UPerNet\\\citep{xiao2018unifiedupernet}}} 
    & DFMGAN & 90.74 & 94.43 & 90.37 & 92.33 & 57.01 & 56.91 & 46.64 & 71.69 & 71.64 & 70.70 & 75.09 & 12.42 & 18.52 & 15.47 & 67.56 & 84.53 & 84.99 & 75.12 & 19.54 & 26.04 & 18.78 \\
    & AnomalyDiffusion & 96.62 & 98.61 & 96.21 & 96.87 & 69.92 & 66.95 & 50.80 & 83.18 & 84.08 & 78.88 & 95.00 & 39.92 & 45.37 & 20.53 & 76.56 & 90.42 & 87.35 & 88.48 & 28.95 & 35.81 & 25.04 \\
    & \textbf{SeaS} & \textbf{98.29} & \textbf{99.20} & \textbf{97.34} & \textbf{97.87} & \textbf{74.42} & \textbf{70.70} & \textbf{61.24} & \textbf{90.34} & \textbf{90.73} & \textbf{84.33} & \textbf{97.01} & \textbf{55.46} & \textbf{55.99} & \textbf{35.91} & \textbf{82.57} & \textbf{92.59} & \textbf{88.72} & \textbf{91.93} & \textbf{38.51} & \textbf{43.53} & \textbf{38.56} \\
    \midrule
   \multirow{4}{*}{\makecell{LFD\\\citep{zhou2024lfdroadseg}}} 
    & DFMGAN & 91.08 & 95.40 & 90.58 & 94.91 & 67.06 & 65.09 & 45.49 & 65.38 & 62.25 & 66.59 & 81.21 & 15.14 & 18.70 & 6.44 & 62.23 & 82.17 & 85.38 & 72.15 & 9.54 & 14.29 & 14.81 \\
    & AnomalyDiffusion & 95.15 & 97.78 & 94.66 & 96.30 & 69.77 & 66.99 & 45.77 & 81.97 & 82.36 & 77.35 & 88.00 & 30.86 & 38.56 & 16.61 & 77.06 & 89.44 & 87.20 & \textbf{92.68} & 24.29 & 32.74 & 19.90 \\
    & \textbf{SeaS} & \textbf{95.88} & \textbf{97.89} & \textbf{95.15} & \textbf{98.09} & \textbf{77.15} & \textbf{72.52} & \textbf{56.47} & \textbf{83.07} & \textbf{82.88} & \textbf{77.24} & \textbf{92.91} & \textbf{43.87} & \textbf{46.46} & \textbf{26.37} & \textbf{78.96} & \textbf{91.22} & \textbf{87.28} & 91.61 & \textbf{40.25} & \textbf{43.47} & \textbf{39.00} \\
    \midrule
   \multirow{4}{*}{\makecell{Average}}
    & DFMGAN & 90.91 & 94.75 & 90.43 & 93.94 & 61.50 & 60.85 & 45.99 & 66.71 & 65.51 & 67.92 & 77.40 & 12.24 & 17.41 & 10.52 & 63.89 & 83.83 & 84.94 & 74.39 & 14.70 & 20.69 & 16.42 \\
    & AnomalyDiffusion & 93.95 & 97.08 & 94.24 & 96.48 & 68.06 & 65.40 & 46.49 & 80.42 & 81.39 & 76.45 & 90.76 & 34.98 & 40.62 & 17.69 & 71.70 & 87.07 & 86.64& 91.18 & 22.80 & 29.55 & 19.88 \\
    & \textbf{SeaS} & \textbf{96.72} & \textbf{98.41} & \textbf{95.97} & \textbf{97.72} & \textbf{73.59} & \textbf{69.86} & \textbf{57.66} & \textbf{86.34} & \textbf{86.75} & \textbf{80.69} & \textbf{95.32} & \textbf{47.38} & \textbf{49.29} & \textbf{29.40} & \textbf{78.38} & \textbf{90.52} & \textbf{87.27} & \textbf{91.32} & \textbf{34.93} & \textbf{39.87} & \textbf{35.37} \\
    \bottomrule 
\end{tabular}
}
\vspace{-0.9cm}
\end{center}
\end{table*}

\noindent\textbf{Evaluation Metrics.}
For image generation, unlike existing methods \citep{Duan2023DFMGAN, hu2023anomalydiffusion} that only assess the whole anomaly images, our evaluation contains three levels: anomaly images, normal images, and anomalies, using 4 metrics:
(1)Inception Score (IS) and Intra-cluster pairwise LPIPS distance (IC-LPIPS) \citep{Ojha_CDC} for authenticity and diversity of anomaly images.
(2) KID \citep{binkowski2018demystifyingkid} for authenticity of normal images.
(3)IC-LPIPS calculated only in anomaly regions (short for IC-LPIPS(a)) for the diversity of anomalies.
For pixel-level anomaly segmentation and image-level anomaly detection, we use 3 metrics: Area Under Receiver Operator Characteristic curve (AUROC), Average Precision (AP) and $F_1$-score at optimal threshold ($F_1$-max). 
We also report Intersection over Union (IoU) for segmentation.

\vspace{-6pt}
\subsection{Comparison in Anomaly Image Generation}
\vspace{-2pt}

\noindent\textbf{Comparison Methods.}
For image generation, we compare SeaS with current anomaly image generation methods, like Crop\&Paste \citep{ICME2021crop-and-paste}, SDGAN \citep{niu2020sdgan}, Defect-GAN \citep{zhang2021defectgan}(all without open-source code), DFMGAN \citep{Duan2023DFMGAN}, and AnomalyDiffusion \citep{hu2023anomalydiffusion} in terms of fidelity and diversity.
For diverse generated anomalies, we combine SeaS-generated anomalies with synthesis-based AD approaches like DRAEM \citep{zavrtanik2021draem} and GLASS \citep{chen2024unifiedglass}.
For authentic generated normal images, we use SeaS-generated normal images to augment the training sets of unsupervised AD methods like HVQ-Trans \citep{lu2023hvqtrans}, PatchCore \citep{roth2022patchcore}, and MambaAD \citep{he2025mambaad}.
For anomaly image-mask pairs, we generate them with DFMGAN, AnomalyDiffusion, and SeaS, to train segmentation models like BiSeNet V2 \citep{yu2021bisenet}, UPerNet \citep{xiao2018unifiedupernet}, and LFD \citep{zhou2024lfdroadseg} respectively.
Different from AnomalyDiffusion,
which trains one segmentation model per product, we train a unified supervised segmentation model for all products, which is more challenging.

\noindent\textbf{Anomaly image generation quality.}
In Tab. \ref{table:generation}, we compare SeaS with some state-of-the-art anomaly image generation methods on fidelity (IS and KID) and diversity (IC-LPIPS and IC-LPIPS(a)).
SeaS outperforms other methods in IS and IC-LPIPS, showing superior fidelity and diversity. 
It also excels in generating authentic normal images and diverse anomalies. Compared to AnomalyDiffusion, which cannot generate normal images, SeaS leads in IC-LPIPS(a). SeaS also surpasses DFMGAN in both KID and IC-LPIPS(a).
We exhibit the generated anomaly images in Fig. \ref{fig:generation}, SeaS-generated anomaly images have higher fidelity (e.g., \emph{hazelnut\_crack}).
Compared with other methods, SeaS can generate images with different types, colors, and shapes of anomalies rather than overfitting to the training images (e.g., \emph{foam\_color}).
SeaS-generated masks are also precisely aligned with the anomaly regions (e.g., \emph{toothbrush\_defective}).
We also present the authentic generated normal images in Fig. \ref{fig:generation2}. More qualitative and quantitative anomaly image generation results are in appendix \ref{appendix:generation_results}.

\noindent\textbf{Combining generated anomalies with synthesis-based AD methods.}
We replace the synthesized pseudo-anomalies in DRAEM \citep{zavrtanik2021draem} and GLASS \citep{chen2024unifiedglass} with SeaS-generated anomalies.
As shown in Tab. \ref{table:combined_replace}, SeaS-generated anomalies, which offer sufficient diversity, consistently improve synthesis-based AD methods by suppressing false negatives, leading to better performance across multiple datasets. More training details are given in appendix \ref{appendix:implementation}.

\noindent\textbf{Combining generated normal images with AD methods.}
We use SeaS-generated normal images to supplement the training sets of existing state-of-the-art unsupervised AD methods, the results are given in Tab. \ref{table:combined_unad}.
Using SeaS-generated normal images with minor local variations and global consistency, unsupervised AD methods reduce false positives and perform well across multiple metrics, improving industrial anomaly detection on various datasets. More training details are given in appendix \ref{appendix:implementation}
.

\begin{figure}[!t]
\vspace{-2pt}
\centering
\includegraphics[width=1\linewidth]
{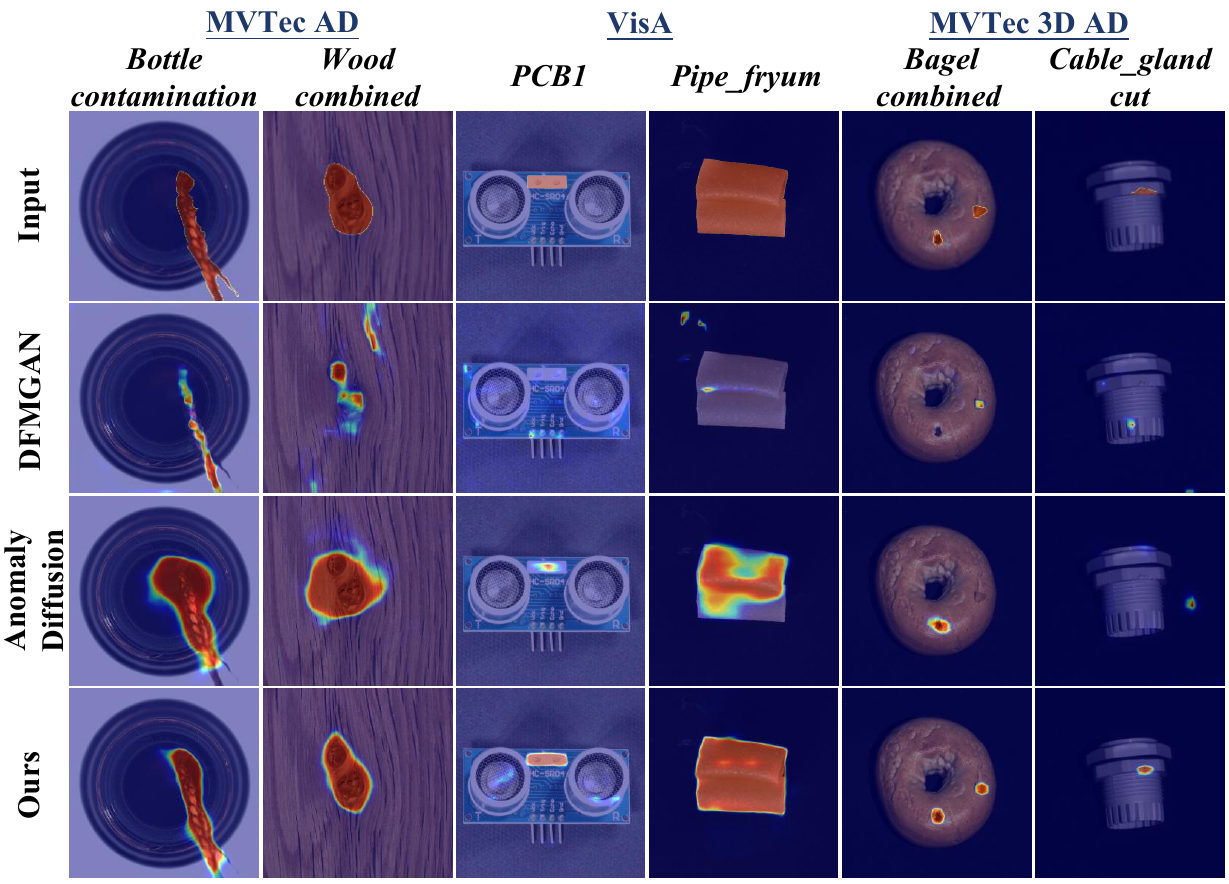}
\vspace{-0.7cm}
\caption{
Qualitative supervised anomaly segmentation results with BiSeNet V2 on MVTec AD.
}
\label{Fig:result_compare}
\vspace{-0.75cm}
\end{figure}

\noindent\textbf{Training supervised segmentation models for anomaly segmentation and detection.}
We generate 1,000 image-mask pairs for each anomaly type and use them, 
along with all normal images in the original training sets, to train a unified supervised segmentation model.
The models are tested on the remaining images not included in the training set. 
All methods are trained using the same number of images and the training settings, detailed in appendix \ref{appendix_seg}.
As shown in Tab. \ref{table:seg_combined}, the segmentation results consistently demonstrate that our method outperforms others across all the segmentation models, with average IoU improvements of 11.17\% (MVTec AD), 11.71\% (VisA), and 15.49\% (MVTec 3D AD). 
Segmentation anomaly maps are shown in Fig. \ref{Fig:result_compare}. 
Using our generated image-mask pairs to train BiSeNet V2, there are fewer false positives in \emph{wood\_combined} and fewer false negatives in \emph{bottle\_contamination}.
We also use the maximum value of the segmentation anomaly map as the image-level anomaly score for anomaly detection, achieving gains of 2.77\% (MVTec AD), 5.92\% (VisA), and 6.68\% (MVTec 3D AD) in image-AUROC.
More qualitative comparison results are in appendix \ref{appendix:seg_result1} and appendix \ref{appendix:seg_result2}.

\vspace{-2pt}
\subsection{Ablation Study}
\vspace{-2pt}

We train additional models to assess the effect of each component: 
\textbf{(a)} the model with predefined typical text prompt with fixed generic semantic words (short for with TP in Tab. \ref{tab:ablation_ag});
\textbf{(b)} the model without mixing different types of anomaly images in the same product;
\textbf{(c)} the model without NA loss;
\textbf{(d)} the model without the second term of DA loss in Eq. \ref{eq:decoupleloss} (short for ST in Tab. \ref{tab:ablation_ag});
\textbf{(e)} our complete model.
We use these models to generate 1,000 anomaly image-mask pairs per anomaly type and train BiSeNet V2 for supervised anomaly segmentation.
In Tab. \ref{tab:ablation_ag}, the results show that omitting any component leads to a decrease in fidelity and diversity of the generated images, as well as in the segmentation results.
These validate the effectiveness of the components we proposed. 
More ablation studies on SeaS are shown in appendix \ref{appendix:ablation}.

\begin{table}[t]
    \centering
    \caption{Ablation on the generation model.}
    \vspace{-0.25cm}
    \label{tab:ablation_ag}
    \setlength{\tabcolsep}{2pt}{
        \resizebox{0.87\linewidth}{!}{
            \begin{tabular}{m{2.5cm}|cccccc}
                \specialrule{0.6pt}{0.1pt}{0pt}
                \multirow{2}{*}{\hspace*{1em} Method}& \multicolumn{6}{c}{Metrics} \\
                &IS&IC-L&AUROC& AP &$F_1$-max&IoU \\
                \hline
                \hline
                (a) with TP &1.72&0.33&94.72&57.16&55.67&50.46 \\ 
                (b) w/o Mixed &1.79 &0.32&95.82&66.07&64.50&53.11\\ 
                (c) w/o NA &1.67&0.31 &96.20&66.03&64.09&53.97\\ 
                (d) w/o ST &1.86&0.33&96.44&67.73&65.23&54.99\\
                \textbf{(e) All (Ours)}&\textbf{1.88}&\textbf{0.34}&\textbf{97.21}&\textbf{69.21}&\textbf{66.37}&\textbf{55.28}\\
                \specialrule{0.8pt}{0.1pt}{0pt}            
            \end{tabular}
        }   
    }
\vspace{-0.5cm}
\end{table}

\noindent\textbf{Refined Mask Prediction branch.}
To verify the validity of the components in the RMP branch, we conduct ablation studies on MRM, the progressive manner to refine coarse feature (short for PM in Tab. \ref{tab:ablation_mask}) and coarse mask supervision (short for CMS in Tab. \ref{tab:ablation_mask}).
\textbf{1)} the model only with CMS, which means we do not use MRM to fuse the high-resolution features in RMP, but directly obtain the mask from the coarse features $ \hat{F} \in \mathbb{R}^{ 64 \times 64 \times 192 } $ through convolution and bilinear interpolation upsampling;
\textbf{2)} the model with MRM;
\textbf{3)} the model utilizing three MRMs in a progressive manner to refine coarse features;
\textbf{4)} our complete model.
We report the BiSeNet V2 results in Tab. \ref{tab:ablation_mask}, which demonstrates that each component in the RMP is indispensable for downstream supervised anomaly segmentation.
More ablation studies about RMP are in appendix \ref{appendix:ablation}.

\begin{table}[h!]
\vspace{-10pt}
    \centering
    \caption{Ablation on the RMP branch.}
    \vspace{-0.3cm}
    \label{tab:ablation_mask}
    \setlength{\tabcolsep}{2pt}{
        \resizebox{0.81\linewidth}{!}{
            \begin{tabular}{>{\centering\arraybackslash}m{1cm}>{\centering\arraybackslash}m{1cm}>{\centering\arraybackslash}m{1cm}|cccc}
                \specialrule{0.6pt}{0.1pt}{0pt}
                \multicolumn{3}{c|}{Method}& \multicolumn{4}{c}{Metrics} \\
                MRM & PM & CMS & AUROC&  AP  & $F_1$-max & IoU \\
                \hline
                \hline
                 & &$\checkmark$&97.00&65.28&62.56&53.93\\
                $\checkmark$& & &94.54&60.52&59.06&49.42\\
                $\checkmark$&$\checkmark$& &94.04&62.04&59.82&50.44\\    
                $\checkmark$&$\checkmark$&$\checkmark$&\textbf{97.21}&\textbf{69.21}&\textbf{66.37}&\textbf{55.28}\\
                \specialrule{0.8pt}{0.1pt}{0pt}
            \end{tabular}
        }
    }
\vspace{-0.3cm}
\end{table}

\section{Conclusion}
In this paper, we propose a unified generation method named SeaS. We explore an implicit characteristic that anomalies exhibit high variability, 
while normal products maintain global consistency. 
We design a Separation and Sharing Fine-tuning strategy to model different variations of normal products and anomalies, 
enabling the Refined Mask Prediction branch to predict accurate masks with discriminative features.
Our method greatly improves synthesis-based and supervised AD methods, and empowers supervised segmentation models.

\clearpage
\section{Acknowledgement}
This work was supported by the National Natural Science Foundation of China under Grant No.62176098. The computation is completed in the HPC Platform of Huazhong University of Science and Technology.

{
    \small
    \bibliographystyle{ieeenat_fullname}
    \bibliography{main}

\begin{thebibliography}{47}
\providecommand{\natexlab}[1]{#1}
\providecommand{\url}[1]{\texttt{#1}}
\expandafter\ifx\csname urlstyle\endcsname\relax
  \providecommand{\doi}[1]{doi: #1}\else
  \providecommand{\doi}{doi: \begingroup \urlstyle{rm}\Url}\fi

\bibitem[Avrahami et~al.(2023)Avrahami, Aberman, Fried, Cohen-Or, and Lischinski]{avrahami2023bas}
Omri Avrahami, Kfir Aberman, Ohad Fried, Daniel Cohen-Or, and Dani Lischinski.
\newblock Break-a-scene: Extracting multiple concepts from a single image.
\newblock In \emph{SIGGRAPH Asia 2023 Conference Papers}, pages 1--12, 2023.

\bibitem[Barratt and Sharma(2018)]{barratt2018note}
Shane Barratt and Rishi Sharma.
\newblock A note on the inception score.
\newblock \emph{arXiv preprint arXiv:1801.01973}, 2018.

\bibitem[Bergmann et~al.(2019)Bergmann, Fauser, Sattlegger, and Steger]{bergmann2019mvtec}
Paul Bergmann, Michael Fauser, David Sattlegger, and Carsten Steger.
\newblock Mvtec ad--a comprehensive real-world dataset for unsupervised anomaly detection.
\newblock In \emph{Proceedings of the IEEE/CVF Conference on Computer Vision and Pattern Recognition}, pages 9592--9600, 2019.

\bibitem[Bergmann. et~al.(2022)Bergmann., Jin., Sattlegger., and Steger.]{bergmannmvtec3d}
Paul Bergmann., Xin Jin., David Sattlegger., and Carsten Steger.
\newblock The mvtec 3d-ad dataset for unsupervised 3d anomaly detection and localization.
\newblock In \emph{Proceedings of the 17th International Joint Conference on Computer Vision, Imaging and Computer Graphics Theory and Applications}, pages 202--213, 2022.

\bibitem[Bi{\'n}kowski et~al.(2018)Bi{\'n}kowski, Sutherland, Arbel, and Gretton]{binkowski2018demystifyingkid}
Miko{\l}aj Bi{\'n}kowski, Danica~J Sutherland, Michael Arbel, and Arthur Gretton.
\newblock Demystifying mmd gans.
\newblock In \emph{International Conference on Learning Representations}, 2018.

\bibitem[Brooks et~al.(2023)Brooks, Holynski, and Efros]{brooks2023instructpix2pix}
Tim Brooks, Aleksander Holynski, and Alexei~A Efros.
\newblock Instructpix2pix: Learning to follow image editing instructions.
\newblock In \emph{Proceedings of the IEEE/CVF Conference on Computer Vision and Pattern Recognition}, pages 18392--18402, 2023.

\bibitem[Chefer et~al.(2023)Chefer, Alaluf, Vinker, Wolf, and Cohen-Or]{chefer2023attend}
Hila Chefer, Yuval Alaluf, Yael Vinker, Lior Wolf, and Daniel Cohen-Or.
\newblock Attend-and-excite: Attention-based semantic guidance for text-to-image diffusion models.
\newblock \emph{ACM Transactions on Graphics}, 42\penalty0 (4):\penalty0 1--10, 2023.

\bibitem[Chen et~al.(2024{\natexlab{a}})Chen, Zhang, Wang, Duan, Zhou, and Zhu]{chen2023disenbooth}
Hong Chen, Yipeng Zhang, Xin Wang, Xuguang Duan, Yuwei Zhou, and Wenwu Zhu.
\newblock Disenbooth: Disentangled parameter-efficient tuning for subject-driven text-to-image generation.
\newblock In \emph{International Conference on Learning Representations}, 2024{\natexlab{a}}.

\bibitem[Chen et~al.(2024{\natexlab{b}})Chen, Luo, Lv, and Zhang]{chen2024unifiedglass}
Qiyu Chen, Huiyuan Luo, Chengkan Lv, and Zhengtao Zhang.
\newblock A unified anomaly synthesis strategy with gradient ascent for industrial anomaly detection and localization.
\newblock In \emph{European Conference on Computer Vision}, pages 37--54. Springer, 2024{\natexlab{b}}.

\bibitem[DeVries and Taylor(2017)]{devries2017cutout}
Terrance DeVries and Graham~W Taylor.
\newblock Improved regularization of convolutional neural networks with cutout.
\newblock \emph{arXiv preprint arXiv:1708.04552}, 2017.

\bibitem[Duan et~al.(2023)Duan, Hong, Niu, and Zhang]{Duan2023DFMGAN}
Yuxuan Duan, Yan Hong, Li Niu, and Liqing Zhang.
\newblock Few-shot defect image generation via defect-aware feature manipulation.
\newblock In \emph{Proceedings of the AAAI Conference on Artificial Intelligence}, pages 571--578, 2023.

\bibitem[Gal et~al.(2022)Gal, Alaluf, Atzmon, Patashnik, Bermano, Chechik, and Cohen-or]{gal2022TI}
Rinon Gal, Yuval Alaluf, Yuval Atzmon, Or Patashnik, Amit~Haim Bermano, Gal Chechik, and Daniel Cohen-or.
\newblock An image is worth one word: Personalizing text-to-image generation using textual inversion.
\newblock In \emph{International Conference on Learning Representations}, 2022.

\bibitem[Gui et~al.(2024)Gui, Gao, Liu, Wang, and Wu]{2024anogen}
Guan Gui, Bin-Bin Gao, Jun Liu, Chengjie Wang, and Yunsheng Wu.
\newblock Few-shot anomaly-driven generation for anomaly classification and segmentation.
\newblock In \emph{European Conference on Computer Vision}, pages 210--226, 2024.

\bibitem[Han et~al.(2023)Han, Li, Zhang, Milanfar, Metaxas, and Yang]{han2023svdiff}
Ligong Han, Yinxiao Li, Han Zhang, Peyman Milanfar, Dimitris Metaxas, and Feng Yang.
\newblock Svdiff: Compact parameter space for diffusion fine-tuning.
\newblock In \emph{Proceedings of the IEEE/CVF International Conference on Computer Vision}, pages 7323--7334, 2023.

\bibitem[He et~al.(2025)He, Bai, Zhang, He, Chen, Gan, Wang, Li, Tian, and Xie]{he2025mambaad}
Haoyang He, Yuhu Bai, Jiangning Zhang, Qingdong He, Hongxu Chen, Zhenye Gan, Chengjie Wang, Xiangtai Li, Guanzhong Tian, and Lei Xie.
\newblock Mambaad: Exploring state space models for multi-class unsupervised anomaly detection.
\newblock \emph{Advances in Neural Information Processing Systems}, 37:\penalty0 71162--71187, 2025.

\bibitem[Hertz et~al.(2022)Hertz, Mokady, Tenenbaum, Aberman, Pritch, and Cohen-or]{hertz2022prompt}
Amir Hertz, Ron Mokady, Jay Tenenbaum, Kfir Aberman, Yael Pritch, and Daniel Cohen-or.
\newblock Prompt-to-prompt image editing with cross-attention control.
\newblock In \emph{International Conference on Learning Representations}, 2022.

\bibitem[Hu et~al.(2024)Hu, Zhang, Yi, Du, Chen, Liu, Wang, and Wang]{hu2023anomalydiffusion}
Teng Hu, Jiangning Zhang, Ran Yi, Yuzhen Du, Xu Chen, Liang Liu, Yabiao Wang, and Chengjie Wang.
\newblock Anomalydiffusion: Few-shot anomaly image generation with diffusion model.
\newblock In \emph{Proceedings of the AAAI Conference on Artificial Intelligence}, 2024.

\bibitem[Jeong et~al.(2023)Jeong, Zou, Kim, Zhang, Ravichandran, and Dabeer]{CVPR2023winclip}
Jongheon Jeong, Yang Zou, Taewan Kim, Dongqing Zhang, Avinash Ravichandran, and Onkar Dabeer.
\newblock Winclip: Zero-/few-shot anomaly classification and segmentation.
\newblock In \emph{Proceedings of the IEEE/CVF Conference on Computer Vision and Pattern Recognition}, pages 19606--19616, 2023.

\bibitem[Jin et~al.(2024)Jin, Tanno, Saseendran, Diethe, and Teare]{jin2023image}
Chen Jin, Ryutaro Tanno, Amrutha Saseendran, Tom Diethe, and Philip Teare.
\newblock An image is worth multiple words: Learning object level concepts using multi-concept prompt learning.
\newblock In \emph{International Conference on Machine Learning}, 2024.

\bibitem[Kumari et~al.(2023)Kumari, Zhang, Zhang, Shechtman, and Zhu]{kumari2023multi}
Nupur Kumari, Bingliang Zhang, Richard Zhang, Eli Shechtman, and Jun-Yan Zhu.
\newblock Multi-concept customization of text-to-image diffusion.
\newblock In \emph{Proceedings of the IEEE/CVF Conference on Computer Vision and Pattern Recognition}, pages 1931--1941, 2023.

\bibitem[Li et~al.(2021)Li, Sohn, Yoon, and Pfister]{li2021cutpaste}
Chun-Liang Li, Kihyuk Sohn, Jinsung Yoon, and Tomas Pfister.
\newblock Cutpaste: Self-supervised learning for anomaly detection and localization.
\newblock In \emph{Proceedings of the IEEE/CVF conference on Computer Vision and Pattern Recognition}, pages 9664--9674, 2021.

\bibitem[Li et~al.(2022)Li, Ling, Kim, Kreis, Fidler, and Torralba]{li2022bigdatasetgan}
Daiqing Li, Huan Ling, Seung~Wook Kim, Karsten Kreis, Sanja Fidler, and Antonio Torralba.
\newblock Bigdatasetgan: Synthesizing imagenet with pixel-wise annotations.
\newblock In \emph{Proceedings of the IEEE/CVF Conference on Computer Vision and Pattern Recognition}, pages 21330--21340, 2022.

\bibitem[Lin et~al.(2021)Lin, Cao, Zhu, and Li]{ICME2021crop-and-paste}
Dongyun Lin, Yanpeng Cao, Wenbin Zhu, and Yiqun Li.
\newblock Few-shot defect segmentation leveraging abundant defect-free training samples through normal background regularization and crop-and-paste operation.
\newblock In \emph{2021 IEEE International Conference on Multimedia and Expo}, pages 1--6, 2021.

\bibitem[Lin et~al.(2017)Lin, Goyal, Girshick, He, and Doll{\'a}r]{lin2017focal}
Tsung-Yi Lin, Priya Goyal, Ross Girshick, Kaiming He, and Piotr Doll{\'a}r.
\newblock Focal loss for dense object detection.
\newblock In \emph{Proceedings of the IEEE International Conference on Computer Vision}, pages 2980--2988, 2017.

\bibitem[Loshchilov and Hutter(2018)]{loshchilov2018decoupled}
Ilya Loshchilov and Frank Hutter.
\newblock Decoupled weight decay regularization.
\newblock In \emph{International Conference on Learning Representations}, 2018.

\bibitem[Lu et~al.(2023)Lu, Wu, Tian, Wang, Chen, Liu, and Hu]{lu2023hvqtrans}
Ruiying Lu, YuJie Wu, Long Tian, Dongsheng Wang, Bo Chen, Xiyang Liu, and Ruimin Hu.
\newblock Hierarchical vector quantized transformer for multi-class unsupervised anomaly detection.
\newblock \emph{Advances in Neural Information Processing Systems}, 36:\penalty0 8487--8500, 2023.

\bibitem[Nguyen et~al.(2024)Nguyen, Vu, Tran, and Nguyen]{nguyen2024dataset}
Quang Nguyen, Truong Vu, Anh Tran, and Khoi Nguyen.
\newblock Dataset diffusion: Diffusion-based synthetic data generation for pixel-level semantic segmentation.
\newblock \emph{Advances in Neural Information Processing Systems}, 36, 2024.

\bibitem[Niu et~al.(2020)Niu, Li, Wang, and Lin]{niu2020sdgan}
Shuanlong Niu, Bin Li, Xinggang Wang, and Hui Lin.
\newblock Defect image sample generation with gan for improving defect recognition.
\newblock \emph{IEEE Transactions on Automation Science and Engineering}, 17\penalty0 (3):\penalty0 1611--1622, 2020.

\bibitem[Ojha et~al.(2021)Ojha, Li, Lu, Efros, Lee, Shechtman, and Zhang]{Ojha_CDC}
Utkarsh Ojha, Yijun Li, Jingwan Lu, Alexei~A. Efros, Yong~Jae Lee, Eli Shechtman, and Richard Zhang.
\newblock Few-shot image generation via cross-domain correspondence.
\newblock In \emph{Proceedings of the IEEE/CVF Conference on Computer Vision and Pattern Recognition}, pages 10743--10752, 2021.

\bibitem[P{\'e}rez et~al.(2003)P{\'e}rez, Gangnet, and Blake]{perez2003poisson}
Patrick P{\'e}rez, Michel Gangnet, and Andrew Blake.
\newblock Poisson image editing.
\newblock In \emph{ACM SIGGRAPH 2003}, pages 313--318. 2003.

\bibitem[Rombach et~al.(2022)Rombach, Blattmann, Lorenz, Esser, and Ommer]{rombach2022high}
Robin Rombach, Andreas Blattmann, Dominik Lorenz, Patrick Esser, and Bj{\"o}rn Ommer.
\newblock High-resolution image synthesis with latent diffusion models.
\newblock In \emph{Proceedings of the IEEE/CVF Conference on Computer Vision and Pattern Recognition}, pages 10684--10695, 2022.

\bibitem[Roth et~al.(2022)Roth, Pemula, Zepeda, Sch{\"o}lkopf, Brox, and Gehler]{roth2022patchcore}
Karsten Roth, Latha Pemula, Joaquin Zepeda, Bernhard Sch{\"o}lkopf, Thomas Brox, and Peter Gehler.
\newblock Towards total recall in industrial anomaly detection.
\newblock In \emph{Proceedings of the IEEE/CVF Conference on Computer Vision and Pattern Recognition}, pages 14318--14328, 2022.

\bibitem[Ruiz et~al.(2023)Ruiz, Li, Jampani, Pritch, Rubinstein, and Aberman]{ruiz2023dreambooth}
Nataniel Ruiz, Yuanzhen Li, Varun Jampani, Yael Pritch, Michael Rubinstein, and Kfir Aberman.
\newblock Dreambooth: Fine tuning text-to-image diffusion models for subject-driven generation.
\newblock In \emph{Proceedings of the IEEE/CVF Conference on Computer Vision and Pattern Recognition}, pages 22500--22510, 2023.

\bibitem[Schl{\"u}ter et~al.(2022)Schl{\"u}ter, Tan, Hou, and Kainz]{schluter2022nsa}
Hannah~M Schl{\"u}ter, Jeremy Tan, Benjamin Hou, and Bernhard Kainz.
\newblock Natural synthetic anomalies for self-supervised anomaly detection and localization.
\newblock In \emph{European Conference on Computer Vision}, pages 474--489. Springer, 2022.

\bibitem[Wu et~al.(2023{\natexlab{a}})Wu, Zhao, Chen, Gu, Zhao, He, Zhou, Shou, and Shen]{wu2023datasetdm}
Weijia Wu, Yuzhong Zhao, Hao Chen, Yuchao Gu, Rui Zhao, Yefei He, Hong Zhou, Mike~Zheng Shou, and Chunhua Shen.
\newblock Datasetdm: Synthesizing data with perception annotations using diffusion models.
\newblock \emph{Advances in Neural Information Processing Systems}, 36:\penalty0 54683--54695, 2023{\natexlab{a}}.

\bibitem[Wu et~al.(2023{\natexlab{b}})Wu, Zhao, Shou, Zhou, and Shen]{wu2023diffumask}
Weijia Wu, Yuzhong Zhao, Mike~Zheng Shou, Hong Zhou, and Chunhua Shen.
\newblock Diffumask: Synthesizing images with pixel-level annotations for semantic segmentation using diffusion models.
\newblock In \emph{Proceedings of the IEEE/CVF International Conference on Computer Vision}, pages 1206--1217, 2023{\natexlab{b}}.

\bibitem[Xiao et~al.(2023)Xiao, Yin, Freeman, Durand, and Han]{xiao2023fastcomposer}
Guangxuan Xiao, Tianwei Yin, William~T Freeman, Fr{\'e}do Durand, and Song Han.
\newblock Fastcomposer: Tuning-free multi-subject image generation with localized attention.
\newblock \emph{arXiv preprint arXiv:2305.10431}, 2023.

\bibitem[Xiao et~al.(2018)Xiao, Liu, Zhou, Jiang, and Sun]{xiao2018unifiedupernet}
Tete Xiao, Yingcheng Liu, Bolei Zhou, Yuning Jiang, and Jian Sun.
\newblock Unified perceptual parsing for scene understanding.
\newblock In \emph{Proceedings of the European Conference on Computer Vision}, pages 418--434, 2018.

\bibitem[Xie et~al.(2023)Xie, Li, Huang, Liu, Zhang, Zheng, and Shou]{xie2023boxdiff}
Jinheng Xie, Yuexiang Li, Yawen Huang, Haozhe Liu, Wentian Zhang, Yefeng Zheng, and Mike~Zheng Shou.
\newblock Boxdiff: Text-to-image synthesis with training-free box-constrained diffusion.
\newblock In \emph{Proceedings of the IEEE/CVF International Conference on Computer Vision}, pages 7452--7461, 2023.

\bibitem[Yu et~al.(2021)Yu, Gao, Wang, Yu, Shen, and Sang]{yu2021bisenet}
Changqian Yu, Changxin Gao, Jingbo Wang, Gang Yu, Chunhua Shen, and Nong Sang.
\newblock Bisenet v2: Bilateral network with guided aggregation for real-time semantic segmentation.
\newblock \emph{International Journal of Computer Vision}, 129:\penalty0 3051--3068, 2021.

\bibitem[Zavrtanik et~al.(2021)Zavrtanik, Kristan, and Sko{\v{c}}aj]{zavrtanik2021draem}
Vitjan Zavrtanik, Matej Kristan, and Danijel Sko{\v{c}}aj.
\newblock Draem-a discriminatively trained reconstruction embedding for surface anomaly detection.
\newblock In \emph{Proceedings of the IEEE/CVF International Conference on Computer Vision}, pages 8330--8339, 2021.

\bibitem[Zhang et~al.(2021{\natexlab{a}})Zhang, Cui, Hung, and Lu]{zhang2021defectgan}
Gongjie Zhang, Kaiwen Cui, Tzu-Yi Hung, and Shijian Lu.
\newblock Defect-gan: High-fidelity defect synthesis for automated defect inspection.
\newblock In \emph{Proceedings of the IEEE/CVF Winter Conference on Applications of Computer Vision}, pages 2524--2534, 2021{\natexlab{a}}.

\bibitem[Zhang et~al.(2023)Zhang, Rao, and Agrawala]{zhang2023adding}
Lvmin Zhang, Anyi Rao, and Maneesh Agrawala.
\newblock Adding conditional control to text-to-image diffusion models.
\newblock In \emph{Proceedings of the IEEE/CVF International Conference on Computer Vision}, pages 3836--3847, 2023.

\bibitem[Zhang et~al.(2021{\natexlab{b}})Zhang, Ling, Gao, Yin, Lafleche, Barriuso, Torralba, and Fidler]{zhang2021datasetgan}
Yuxuan Zhang, Huan Ling, Jun Gao, Kangxue Yin, Jean-Francois Lafleche, Adela Barriuso, Antonio Torralba, and Sanja Fidler.
\newblock Datasetgan: Efficient labeled data factory with minimal human effort.
\newblock In \emph{Proceedings of the IEEE/CVF Conference on Computer Vision and Pattern Recognition}, pages 10145--10155, 2021{\natexlab{b}}.

\bibitem[Zhou et~al.(2024{\natexlab{a}})Zhou, Xue, Li, Gong, Li, and Zhou]{zhou2024lfdroadseg}
Huan Zhou, Feng Xue, Yucong Li, Shi Gong, Yiqun Li, and Yu Zhou.
\newblock Exploiting low-level representations for ultra-fast road segmentation.
\newblock \emph{IEEE Transactions on Intelligent Transportation Systems}, 2024{\natexlab{a}}.

\bibitem[Zhou et~al.(2024{\natexlab{b}})Zhou, Pang, Tian, He, and Chen]{iclr2024anomalyclip}
Qihang Zhou, Guansong Pang, Yu Tian, Shibo He, and Jiming Chen.
\newblock Anomalyclip: Object-agnostic prompt learning for zero-shot anomaly detection.
\newblock In \emph{International Conference on Learning Representations}, 2024{\natexlab{b}}.

\bibitem[Zou et~al.(2022)Zou, Jeong, Pemula, Zhang, and Dabeer]{zou2022visa}
Yang Zou, Jongheon Jeong, Latha Pemula, Dongqing Zhang, and Onkar Dabeer.
\newblock Spot-the-difference self-supervised pre-training for anomaly detection and segmentation.
\newblock In \emph{European Conference on Computer Vision}, pages 392--408. Springer, 2022.

\end{thebibliography}
}
\appendix
\newpage
\section{Appendix}
\subsection{Overview}
This supplementary material consists of:
\begin{itemize}
\item Analysis on decoupled anomaly alignment loss and multiple tokens (Sec. \ref{appendix:DAloss}).
\item More implementation details (Sec. \ref{appendix:implementation}).

\item More details of downstream supervised segmentation model implementation and usage (Sec. \ref{appendix_seg}).
\item More ablation studies (Sec. \ref{appendix:ablation}), including ablation studies on the Unbalanced Abnormal Text Prompt design, the Separation and Sharing Fine-tuning loss, the minimum size requirement for training images, the training strategy of SeaS, the cross-attention maps for Decoupled Anomaly Alignment, the features for Coarse Feature Extraction, the features of VAE for Refined Mask Prediction, the normal image supervision for Refined Mask Prediction, the Mask Refinement Module, and the threshold for mask binarization.
\item More qualitative and quantitative results of anomaly image generation (Sec. \ref{appendix:generation_results}).
\item Qualitative comparison results of supervised segmentation models trained on image-mask pairs generated by different anomaly generation methods (Sec. \ref{appendix:seg_result1}).
\item Qualitative comparison results of different supervised segmentation models trained on image-mask pairs generated by SeaS (Sec. \ref{appendix:seg_result2}).
\item Comparison with the Textual Inversion (Sec. \ref{appendix:comp_ti}).
\item More experiments on lighting conditions (Sec. \ref{appendix:lighting}).
\item More results on generation of small defects (Sec. \ref{appendix:small}).
\item More analysis on generation of unseen anomaly types (Sec. \ref{appendix:unseen}).
\item More experiments on comparison with DRAEM (Sec. \ref{appendix:comp_draem}.)

\end{itemize}

\subsection{Analysis on decoupled Anomaly alignment loss and multiple tokens}
\label{appendix:DAloss}
Here we give a more detailed analysis of the learning process of the DA loss.
According to Eq. \ref{eq:decoupleloss}, intuitively, the DA loss may pull the anomaly tokens similar to each other. 
However, the U-Net in Stable Diffusion uses multi-head attention, which ensures that different anomaly tokens cover different attributes of the anomalies.
In Eq. \ref{eq:decoupleloss}, the cross-attention map is the product of the feature map of U-Net and the anomaly tokens. In the implementation of multi-head attention, both the learnable embedding of the anomaly token and the U-Net feature are decomposed into several groups along the channel dimension. E.g., the conditioning vector $ {e}_{a}\in \mathbb{R}^{1 \times C_1} $, which is corresponding to anomaly token, is divided into $ {\{{e}_{a,i} \in \mathbb{R}^{1 \times \frac{C_1}{q}}  | i\in[1,q] \}} $, and the image feature $ {v} \in \mathbb{R}^{r \times r \times C_2} $ is divided into $ {\{{v}_{i} \in \mathbb{R}^{1 \times \frac{C_2}{q}}  | i\in[1,q] \}} $, where $q$ is the number of heads in the multi-head attention. Then the corresponding groups are multiplied, and the outputs of all the heads are averaged. The attention map $ A $ of $ {e}_{a}$ is calculated by:
\begin{equation}
    A = \frac{1}{q}\sum_{i=1}^q  \text{softmax}(\frac{Q_{i}K_{a,i}^{\top}}{\sqrt{d}}), Q_i=\phi_q({v}_{i}),  K_{a,i}=\phi_k({e}_{a,i}). 
\end{equation}
Therefore, in the defect region, the DA loss only ensures the average of each head tends to $1$, but does not require the anomaly tokens to be the same as each other. In addition, each $ {e}_{a} $ is different from each other, and is combined by $ {e}_{a,i} $. 
\textbf{The update direction of each $ {e}_{a,i} $ is related to $ {v}_{i} $ and covers some features of the defect, it encompasses the attributes of anomalies from various perspectives, thereby providing diversified information.
}

\begin{figure}[htb]
\centering
\includegraphics[width=1.0\linewidth]{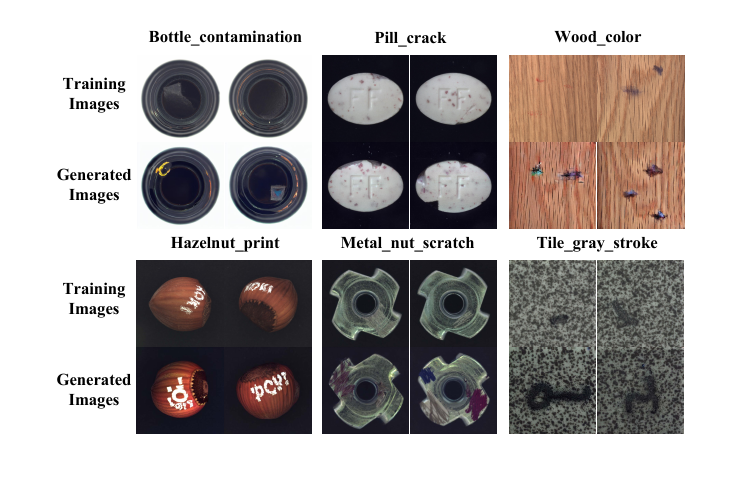}
\caption{Visualization of the generation results for unseen anomalies on MVTec AD.}
\label{fig:appendix_recombined}
\end{figure}

We provide more examples in Fig. \ref{fig:appendix_recombined}, where new anomalies are generated that significantly differ from the training samples in terms of color and shape. For example, we showcase \emph{bottle\_contamination}, \emph{hazelnut\_print}, and \emph{tile\_gray\_stroke} with a novel shape, \emph{wood\_color} and \emph{metal\_nut\_scratch} with a novel color, and \emph{pill\_crack} with a new shape, featuring multiple cracks where the training samples only exhibit a single crack. These examples demonstrate the model's ability to create unseen anomalies based on recombining the decoupled attributes.

\subsection{More implementation details}
\label{appendix:implementation}

\noindent\textbf{More training details.} 
For the Unbalanced Abnormal Text Prompt,
we set the number $N$ of multiple \texttt{<}$\texttt{df}_{n}$\texttt{>} to $4$ and the number $N^\prime$ of \texttt{<}$\texttt{ob}$\texttt{>} to 1, these parameters are fixed across all product classes. 
For a particular type of anomaly, we use the Unbalanced Abnormal (UA) Text Prompt $\mathcal{P}_n$ with different sets of anomaly tokens as the condition to generate the specified type of anomaly.

\begin{center}
    $\mathcal{P}_n$ = \texttt{a <ob> with} \texttt{<}$\texttt{df}_{4 \times n - 3}$\texttt{>,}
    \\
\centerline{\texttt{<}$\texttt{df}_{4 \times n - 2}$\texttt{>,}\texttt{<}$\texttt{df}_{4 \times n -1}$\texttt{>,} \texttt{<}$\texttt{df}_{4 \times n}$\texttt{>}}
\end{center}
\noindent where $n$ represents the index of the anomaly types in the product. 
To generate normal images, we use the embedding the embedding $\textbf{e}_{\text{ob}}$ corresponding to the normal tokens of $\mathcal{P}$, i.e., ``a \texttt{<ob>}", to guide the U-Net in predicting noise.
For example, for the normal token \texttt{<}$\texttt{ob}$\texttt{>}, given the lookup $ \mathcal{U} \in \mathbb{R}^{b \times 768} $, where $ b $ is the number of text embeddings stored by the pre-trained text encoder, we use a placeholder string \texttt{"ob1"} as the input. Firstly, \texttt{"ob1"} is converted to a token ID $ s_{\texttt{ob1}}\in \mathbb{R}^{1 \times 1} $in the tokenizer. Secondly, $ s_{\texttt{ob1}}\in \mathbb{R}^{1 \times 1} $is converted to a one-hot vector $ \mathcal{S}_{\texttt{ob1}} \in \mathbb{R}^{1 \times (b+1)} $. Thirdly, one learnable new embedding $ g \in \mathbb{R}^{1 \times 768} $ corresponding to $ s_{\texttt{ob1}}$ is inserted to the lookup $ \mathcal{U} $, resulting in $ \mathcal{U}^{'} \in \mathbb{R}^{(b+1) \times 768} $. Here, $ g \in \mathbb{R}^{1 \times 768} $ is the learnable embedding of \texttt{<}$\texttt{ob}$\texttt{>}.  \textbf{These embeddings and U-Net are learnable during the fine-tuning process.}

\noindent\textbf{Training image generation model.}
For each product, we perform $800 \times G$ steps for fine-tuning, where $G$ represents the number of anomaly categories of the product. The batch size of the training image generation model is set to 4. During each step of our fine-tuning process, we sample 2 images from the abnormal training set $X_\text{df}$, and 2 images from the normal training set $X_\text{ob}$. We utilize the AdamW \citep{loshchilov2018decoupled} optimizer with a learning rate of U-Net is $4 \times 10^{-6}$. The learning rate of the text embedding is $4 \times 10^{-5}$. 

\noindent\textbf{Training Refined Mask Prediction branch.}
We design a cascaded Refined Mask Prediction (RMP) branch, which is grafted onto the U-Net trained according to SeaS. For each product, we perform $800 \times G$ steps for the RMP model, where $G$ represents the number of anomaly types for the product. The batch size of training the RMP branch is set to 4. During each step of our fine-tuning process, we sample 2 images with their corresponding masks from the abnormal training set $X_\text{df}$, and 2 images from the normal training set $X_\text{ob}$. The masks used to suppress noise in normal images have each pixel value set to 0. The learning rate of the RMP model is $5 \times 10^{-4}$. 

\noindent\textbf{More inference details.}
For all experiments, we use $t$ = 1500 to perform diffusion forward on normal images to get the initial noise. We employ $T$ = 25 steps for sampling.

\noindent\textbf{Metrics.}
For anomaly image generation, we report 4 metrics: the Inception Score (IS) and Intra-cluster pairwise LPIPS Distance (IC-LPIPS) to evaluate the anomaly images, KID \citep{binkowski2018demystifyingkid} to assess the authenticity of normal images, and IC-LPIPS calculated only on anomaly regions (short for IC-LPIPS(a)), to evaluate diversity. The Inception Score (IS), proposed in \citep{barratt2018note}, serves as an independent metric to evaluate the fidelity and diversity of generated images, by measuring the mutual information between input samples and their predicted classes. The IC-LPIPS \citep{Ojha_CDC} is used to evaluate the diversity of generated images, which quantifies the perceptual similarity between image patches in the same cluster.
For pixel-level anomaly segmentation and image-level anomaly detection, we report 3 metrics: Area Under Receiver Operator Characteristic curve (AUROC), Average Precision (AP), and $F_1$-score at the optimal threshold ($F_1$-max). \textbf{All of these metric are calculated using the $\textit{scikit-learn}$ library.}
In addition, we calculate the Intersection over Union (IoU) to more accurately evaluate the anomaly segmentation result.

\noindent\textbf{More training details on anomaly detection methods.}
In this section, we provide more training details of the comparative anomaly detection methods in Tab .\ref{table:combined_replace} and Tab .\ref{table:combined_unad} in the main text. 
For DRAEM \citep{zavrtanik2021draem}, GLASS \citep{chen2024unifiedglass}, and HVQ-Trans \citep{lu2023hvqtrans}, we use the official checkpoints on the MVTec AD dataset, while the others are self-trained due to the lack of official checkpoints. For GLASS, the official foreground masks for the VisA and MVTec 3D AD datasets are not available, so this operation was not used. For PatchCore \citep{roth2022patchcore}, we use the image size of 256 without center cropping, as some anomalies appear at the edges. For MambaAD \citep{he2025mambaad}, we use the provided official checkpoints.

\noindent\textbf{Resource requirement and time consumption.}
We conduct our training on a NVIDIA Tesla A100 40G GPU sequentially for each product category, which may use about 20G memory.
The comparison on time consumption is shown in Tab. \ref{tab:exp_restime_compare}. 
For the MVTec AD datasets, our training takes 73 hours, which is shorter than the 249 hours required by AnomalyDiffusion and the 414 hours required by DFMGAN. In terms of inference time, SeaS costs 720 ms per image, which is shorter than the 3830 ms per image required by the Diffusion-based method AnomalyDiffusion. The inference time of the GAN-based method DFMGAN is 48ms per image.

\begin{table}[t]
\caption{Comparison on resource requirement and time consumption.}
\label{tab:exp_restime_compare}
\begin{center}
    \setlength{\tabcolsep}{2pt}{
    \resizebox{0.8\linewidth}{!}{
\begin{tabular}{c|c|c}
\toprule
\multirow{2}{*}{\textbf{Methods}} & {\textbf{Training}} & \textbf{Inference}          \\
                        & \textbf{Overall Time} & \textbf{Time (per image)} \\
\midrule
\midrule
DFMGAN\citep{Duan2023DFMGAN} & 414 hours                 & \textbf{48 ms}                 \\
AnomalyDiffusion\citep{hu2023anomalydiffusion}        & 249 hours                  & 3830 ms              \\
\midrule
\textbf{SeaS}                   & \textbf{73 hours}                  & 720 ms               \\
\bottomrule
\end{tabular}}
}
\end{center}
\end{table}

\subsection{More details of the supervised segmentation models}
\label{appendix_seg}
As mentioned in the experiment part, we choose three supervised segmentation models (BiSeNet V2 \citep{yu2021bisenet}, UPerNet \citep{xiao2018unifiedupernet}, LFD \citep{zhou2024lfdroadseg}) to verify the validity of the generated image-mask pairs on the downstream supervised anomaly segmentation as well as detection tasks. 
\textbf{For BiSeNet V2 and UPerNet, we generally follow the implementation provided by MMsegmentation. For LFD, we also use the official implementation.}

Specifically, for BiSeNet V2, we choose a backbone structure of a detail branch of three stages with 64, 64 and 128 channels and a semantic branch of four stages with 16, 32, 64 and 128 channels respectively, with a decode head and four auxiliary heads (corresponding to the number of stages in the semantic branch).
As for UPerNet, we choose ResNet-50 as the backbone, with a decode head and an auxiliary head. 

\textbf{In training supervised segmentation models for downstream tasks, we adopt a training strategy of training a unified supervised segmentation model for all classes of products, rather than training separate supervised segmentation models for each class.} Experimental results are shown in Tab. \ref{table:nuscenes}, which indicate that the performance of the unified supervised segmentation model surpasses that of multiple individual supervised segmentation models.

\begin{table}[htb]
\caption{Ablation on the training strategy of supervised segmentation models.}
\label{table:nuscenes}
\begin{center}
\setlength{\tabcolsep}{2pt}
\resizebox{0.48\textwidth}{!}
{

\begin{tabular}{c|cccc|cccc}
    \specialrule{0.6pt}{0.1pt}{0pt}
    
    \multirow{2}{*}{Models} & \multicolumn{4}{c|}{Multiple Models} & \multicolumn{4}{c}{Unified Model}   \\
    & AUROC & AP & $F_1$-max & IoU  &AUROC & AP & $F_1$-max& IoU \\
    \hline
    BiSeNet V2 &96.00 &67.68&65.87 &54.11 & 97.21 & 69.21 & 66.37 & 55.28 \\
    UPerNet  &96.77&73.88&70.49&60.37& 97.87 & 74.42 & 70.70 & 61.24\\
    LFD &93.02&72.97&71.56&55.88& 98.09 & 77.15 & 72.52 & 56.47  \\
    \hline
    Average  &95.26&71.51&69.31&56.79&\textbf{97.72} & \textbf{73.59} & \textbf{69.86} & \textbf{57.66}  \\
    \specialrule{0.8pt}{0.1pt}{0pt}
    \end{tabular}
}
\vspace{-0.3cm}
\end{center}
\end{table}

\subsection{More ablation studies}
\label{appendix:ablation}

\noindent\textbf{Ablation on the Unbalanced Abnormal Text Prompt design}

In the design of the prompt for industrial anomaly image generation, we conduct experiments to validate the effectiveness of our Unbalanced Abnormal (UA) Text Prompt for each anomaly type of each product. We set the number of learnable \texttt{<}$\texttt{df}_{n}$\texttt{>} to $N$, and the number of learnable \texttt{<}$\texttt{ob}_{j}$\texttt{>} to $N^\prime$. As shown in Tab. \ref{tab:appendix_uatp}, by utilizing the UA Text Prompt, i.e.,

\centerline{
    $\mathcal{P}$ = \texttt{a <ob> with} \texttt{<}$\texttt{df}_{1}$\texttt{>,}\texttt{<}$\texttt{df}_{2}$\texttt{>,} \texttt{<}$\texttt{df}_{3}$\texttt{>,} \texttt{<}$\texttt{df}_{4}$\texttt{>}
}
\vspace{-0.1cm}
we are able to provide high-fidelity and diverse images for downstream supervised anomaly segmentation tasks, resulting in the best performance in segmentation metrics.

\noindent\textbf{Ablation on the Separation and Sharing Fine-tuning loss} 

In the design of the DA loss and NA loss for the Separation and Sharing Fine-tuning, we conduct two sets of experiments: (a) We remove the second term in the DA loss (short for w/o ST in Tab. \ref{tab:appendix_loss}); (b) We replace the second term in DA loss with another term in the NA loss (short for with AT in Tab. \ref{tab:appendix_loss}), which aligns the background area with the token \texttt{<}$\texttt{ob}$\texttt{>} according to the mask:
    \begin{equation}
      \mathcal{L}_{\textbf{ob}}= \sum_{l=1}^{L}(||A_{\text{ob}}^{l} - (1 -  M^{l})||^2) + ||\epsilon_{\text{ob}} - \epsilon_{\theta}(\hat{z}_{\text{ob}},t_{\text{ob}},\textbf{e}_{\text{ob}})||_2^2
      \label{eq:decoupleloss_appendix}
    \end{equation}
where $A^{l}_{\text{ob} }\in \mathbb{R}^{r \times r \times 1}$ is the cross-attention map corresponding to the normal token \texttt{<ob>}.
As shown in Tab. \ref{tab:appendix_loss}, the experimental results demonstrate that, our adopted loss design achieves the best performance in downstream supervised segmentation tasks.

\begin{table}[h!]
  \centering
  \caption{Ablation on the Unbalanced Abnormal Text Prompt design.}
  \vspace{0.2cm}
  \label{tab:appendix_uatp}
  \setlength{\tabcolsep}{2pt}{
    \resizebox{0.4\textwidth}{!}{
    \begin{tabular}{c|cccc}
        \toprule
        Prompt & AUROC & AP & $F_{1}$-max & IoU \\ 
        \midrule
        \multicolumn{1}{l|}{$N^\prime$ = 1, $N$ = 1} &96.48&63.69&62.50&52.02 \\ 
        \multicolumn{1}{l|}{$N^\prime$ = 1, $N$ = 4 \textbf{(Ours)}} &\textbf{97.21}&\textbf{69.21}&\textbf{66.37}&\textbf{55.28} \\ 
        \multicolumn{1}{l|}{$N^\prime$ = 4, $N$ = 4} &96.55&66.28&63.95&54.07 \\
        \bottomrule
    \end{tabular}
    }
  }
\end{table}

\begin{table}[h!]
  \centering
  \caption{Ablation on the Separation and Sharing Fine-tuning loss.}
  \vspace{0.2cm}
  \label{tab:appendix_loss}
  \setlength{\tabcolsep}{2pt}{
    \resizebox{0.3\textwidth}{!}{
    \begin{tabular}{c|cccc}
        \toprule
        Loss & AUROC & AP & $F_{1}$-max & IoU \\ 
        \midrule
        w/o ST &96.44&67.73&65.23&54.99 \\ 
        with AT &96.42&63.99&62.43&53.36\\ 
        \textbf{ Ours}&\textbf{97.21}&\textbf{69.21}&\textbf{66.37}&\textbf{55.28}  \\ 
        \bottomrule
    \end{tabular}
    }
  }
\end{table}

\noindent\textbf{Ablation on the minimum size requirement for training images}

\begin{figure}[h!]
\centering
\includegraphics[width=1.0\linewidth]{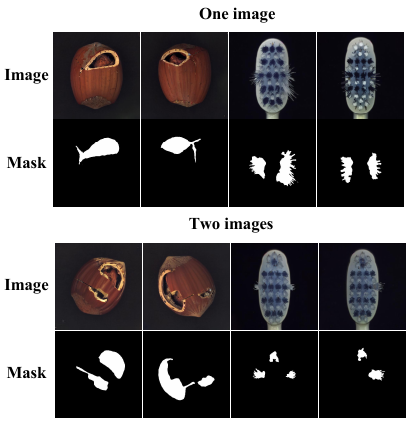}
\caption{Visualization of the ablation study on the minimum size requirement for training images. In the figure, the first row is for generated images, and the second row is for generated masks.}
\label{fig:appendix_minsize}
\end{figure}

In the few-shot setting, for a fair comparison, we follow the common setting in DFMGAN \citep{Duan2023DFMGAN} and AnomalyDiffusion \citep{hu2023anomalydiffusion}, i.e., using one-third abnormal image-mask pairs for each anomaly type in training. In this setting, the minimum number of abnormal training images is 2. Once we adopt a 3-shot setting, we need to reorganize the test set. To ensure that the test set is not reorganized for fair comparison, we take 1-shot and 2-shot settings for all anomaly types during training, i.e., $H=1$ and $H=2$, where $H$ is the image number. The results are shown in Tab. \ref{tab:appendix_min} and Fig. \ref{fig:appendix_minsize}. Observably, the models trained by 1-shot and 2-shot settings still generate anomaly images with decent diversity and authenticity. 

\begin{table}[h!]

    \caption{Ablation on the minimum size requirement for training images.}
    \label{tab:appendix_min}
    \begin{center}

    \setlength{\tabcolsep}{2pt}{
    \resizebox{0.22\textwidth}{!}{
    \begin{tabular}{c|cccc}
        \toprule
        Size  & IS & IC-L  \\ 
        \midrule
        \multicolumn{1}{l|}{$H = 1$} &1.790&0.311 \\
        \multicolumn{1}{l|}{$H = 2$}  &1.794& 0.314 \\
        \multicolumn{1}{l|}{$H = \frac{1}{3} \times H_0$} &\textbf{1.876}&\textbf{0.339} \\ 
        \bottomrule
    \end{tabular}
    }
    }
\end{center}
\end{table}

\noindent\textbf{Ablation on the training strategy of SeaS} 

During each step of the fine-tuning process, we sample the same number of images from the abnormal training set $X_\text{df}$ and the normal training set $X_\text{ob}$. To investigate the efficacy of this strategy, we conduct three distinct sets of experiments: (a) prioritizing training with abnormal images followed by normal images (short for Abnormal-Normal in Tab. \ref{tab:appendix_mix}); (b) prioritizing training with abnormal images followed by anomaly images (short for Normal-Abnormal in Tab. \ref{tab:appendix_mix}); (c) training with a mix of both normal and abnormal images in each batch (short for Abnormal\&Normal in Tab. \ref{tab:appendix_mix}).  As shown in Tab. \ref{tab:appendix_mix}, SeaS yields superior performance in anomaly image generation, characterized by both high fidelity and diversity in the generated images.

\begin{table}[htb]

    \caption{Ablation on training strategy of SeaS.}
    \label{tab:appendix_mix}
    \begin{center}

    \setlength{\tabcolsep}{2pt}{
    \resizebox{0.32\textwidth}{!}{
    \begin{tabular}{c|cccc}
        \toprule
        Strategy  & IS & IC-L  \\ 
        \midrule
        \multicolumn{1}{l|}{Abnormal-Normal} &1.53&0.28 \\
        \multicolumn{1}{l|}{Normal-Abnormal}  &1.70& 0.32 \\
        \multicolumn{1}{l|}{Abnormal\&Normal \textbf{(Ours)}} &\textbf{1.88}&\textbf{0.34} \\ 
        \bottomrule
    \end{tabular}
    }
    
    }
\end{center}
\end{table}

\noindent\textbf{Ablation on the cross-attention maps for Decoupled Anomaly Alignment}

In Decoupled Anomaly Alignment (DA) loss, we leverage cross-attention maps from various layers of the U-Net encoder. 
Specifically, we investigate the impact of integrating different cross-attention maps, denoted as $A^1 \in \mathbb{R}^{64 \times 64}$, $A^2 \in \mathbb{R}^{32 \times 32}$, $A^3 \in \mathbb{R}^{16 \times 16}$ and $A^4 \in \mathbb{R}^{8 \times 8}$. These correspond to the cross-attention maps of the ``\texttt{down-1}'', ``\texttt{down-2}'', ``\texttt{down-3}'', and ``\texttt{down-4}'' layers of the encoder in U-Net respectively. As shown in Tab. \ref{tab:appendix_attn}, the experimental results demonstrate that, employing a combination of $\{A^2, A^3\}$ for DA loss, achieves the best performance in downstream supervised segmentation tasks.

\begin{table}[htb!]
  \centering
    \caption{Ablation on the cross-attention maps for Decoupled Anomaly Alignment.}
    \vspace{0.2cm}
  \label{tab:appendix_attn}
  \setlength{\tabcolsep}{2pt}{
  \resizebox{0.4\textwidth}{!}{
    \begin{tabular}{c|cccc}
        \toprule
        $A^l$ & AUROC & AP & $F_{1}$-max & IoU \\ 
        \midrule
        \multicolumn{1}{l|}{$l$ = 1, 2, 3} &96.42&68.92&66.24 &54.52 \\ 
        \multicolumn{1}{l|}{$l$ = 2, 3, 4} &95.71&64.51&62.33&52.46 \\ 
        \multicolumn{1}{l|}{$l$ = 2, 3 \textbf{(Ours)}}&\textbf{97.21}&\textbf{69.21}&\textbf{66.37}&\textbf{55.28} \\
        \bottomrule
    \end{tabular}
    }
    }
\end{table}

\noindent\textbf{Ablation on the features for Coarse Feature Extraction}

In the coarse feature extraction process,  we extract coarse but highly-discriminative features for anomalies from U-Net decoder. Specifically, we investigate the impact of integrating different features, denoted as $F_1 \in \mathbb{R}^{16 \times 16 \times 1280}$, $F_2 \in \mathbb{R}^{32 \times 32 \times 1280}$, $F_3 \in \mathbb{R}^{64 \times 64 \times 640}$ and $F_4 \in \mathbb{R}^{64 \times 64 \times 320}$. These correspond to the output feature ``\texttt{up-1}'', ``\texttt{up-2}'', ``\texttt{up-3}'', and ``\texttt{up-4}'' layers of the encoder in U-Net respectively. 

\begin{figure}[!h]
\centering
\includegraphics[width=0.95\linewidth]{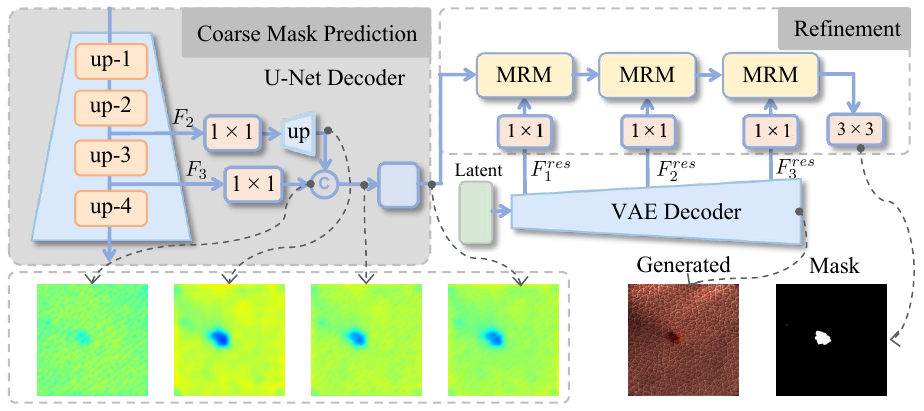}
\caption{Visualization of the U-Net decoder features in mask prediction process.}
\label{fig:appendix_unetdecoder}
\vspace{-0.5cm}
\end{figure}

As shown in Fig. \ref{fig:appendix_unetdecoder}, we use the output features of the “up-2” and “up-3” layers of the decoder in U-Net, and apply convolution blocks and concatenation operations, then we can obtain the unified coarse feature $ \hat{F} \in \mathbb{R}^{ 64 \times 64 \times 192 } $, which can be used to predict masks corresponding to anomaly images.
As shown in Tab. \ref{tab:appendix_unetdecoder}, the experimental results demonstrate that, employing a combination of $\{F_2, F_3\}$ for coarse feature extraction, achieves the best performance in the downstream supervised segmentation task.

\begin{table}[!h]
  \centering
    \caption{Ablation on the features for Coarse Feature Extraction.}
    \vspace{0.2cm}
    \label{tab:appendix_unetdecoder}
    \setlength{\tabcolsep}{2pt}{
    \resizebox{0.4\textwidth}{!}{
    \begin{tabular}{c|cccc}
        \toprule
        $F_y$ & AUROC & AP & $F_{1}$-max & IoU \\
        
        \midrule
        \multicolumn{1}{l|}{$y$ = 1, 2, 3} &94.35&63.58&60.54&52.36 \\ 
        \multicolumn{1}{l|}{$y$ = 2, 3, 4} &96.93&67.42&64.26&\textbf{55.31}\\ 
        \multicolumn{1}{l|}{$y$ = 2, 3 \textbf{(Ours)}}&\textbf{97.21}&\textbf{69.21}&\textbf{66.37}&55.28  \\ 
        \bottomrule
    \end{tabular}
    }
    }
\end{table}

\noindent\textbf{Ablation on the features of VAE for Refined Mask Prediction}

In the Refined Mask Prediction, we combine the high-resolution features of VAE decoder with discriminative features from U-Net, to generate accurately aligned anomaly image-mask pairs. 
In addition, we can also use the VAE encoder features as high-resolution features.
As shown in Tab. \ref{tab:appendix_vaefeatures}, the experimental results show that, using VAE decoder features achieves better performance in downstream supervised segmentation tasks.

\begin{table}[htb]

    \caption{Ablation on the features of VAE for Refined Mask Prediction.}
    \label{tab:appendix_vaefeatures}
    \begin{center}

    \setlength{\tabcolsep}{2pt}{
    \resizebox{0.4\textwidth}{!}{
    \begin{tabular}{c|cccc}
        \toprule
        $F^{res}$  & AUROC & AP & $F_{1}$-max & IoU \\ 
        \midrule
        VAE encoder &96.14&66.26&63.48&54.99 \\
        \textbf{VAE decoder} &\textbf{97.21}&\textbf{69.21}&\textbf{66.37}&\textbf{55.28} \\ 

        \bottomrule
    \end{tabular}
    }
    
    }
    \end{center}
\end{table}

\begin{figure*}[!tb]
\centering
\includegraphics[width=0.7\linewidth]{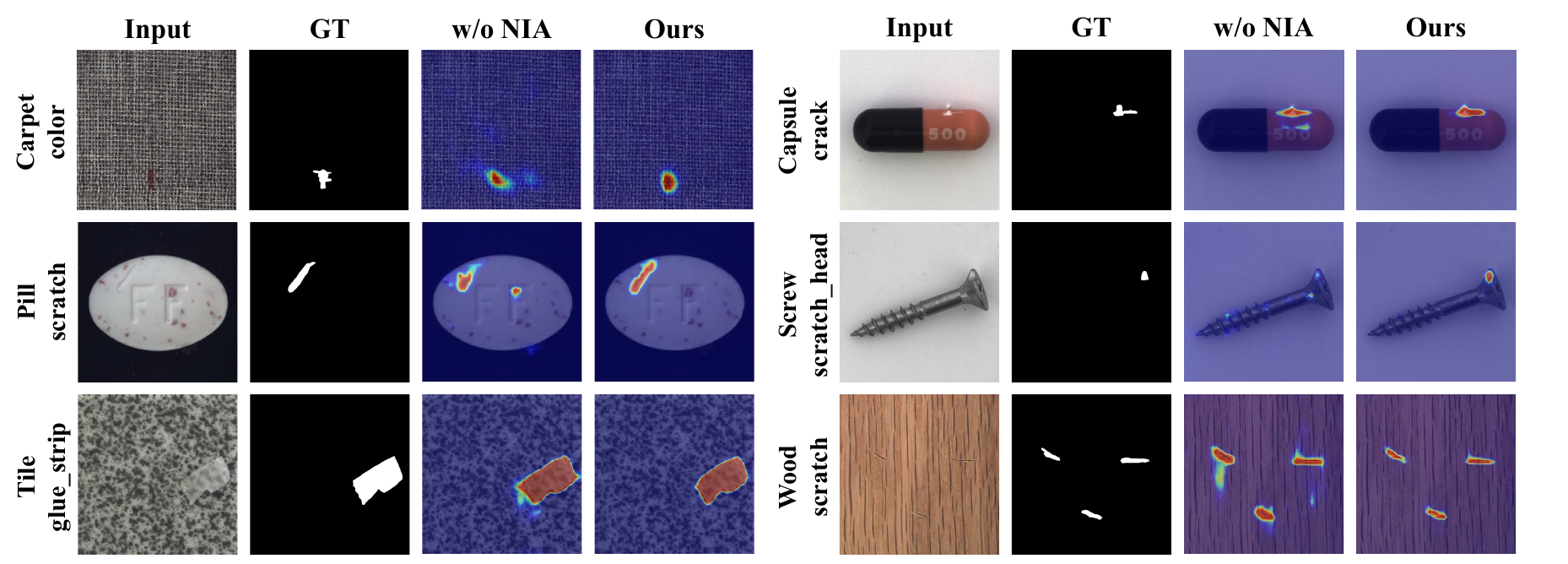}
\caption{Qualitative results of the effect of normal image supervision on MVTec AD.}
\label{fig:appendix_normal}
\end{figure*}

\noindent\textbf{Ablation on the normal image supervision for Refined Mask Prediction}

In the Refined Mask Prediction branch, we predict masks for normal images as the supervision for the mask prediction. We conduct two sets of experiments: (a) We remove the second and the fourth term in the loss for RMP, i.e., the normal image supervision (short for NIA in Tab. \ref{tab:appendix_normal}); (b) We use the complete form in RMP branch loss, i.e., we use the normal image for supervision, as in Eq. 
\eqref{eq:maskloss2}:

\vspace{-0.1cm}
\begin{equation}
\begin{aligned}
\mathcal{L}_\mathcal{M} &=  \mathcal{F} (\hat{M}_\text{df} , \mathbf{M}_\text{df}) +
                        \mathcal{F} (\hat{M}_\text{ob} , \mathbf{M}_\text{ob})  \\
                        &\quad +\mathcal{F} (\hat{M}_\text{df}^\prime , \mathbf{M}_\text{df}^\prime) + 
                        \mathcal{F} (\hat{M}_\text{ob}^\prime , \mathbf{M}_\text{ob}^\prime)
\end{aligned}
\label{eq:maskloss2}
\end{equation}

As shown in Tab. \ref{tab:appendix_normal}, the experimental results show that, using normal images for supervision achieves better performance in downstream supervised segmentation tasks. We also provide further qualitative results of the effect of normal image supervision (short for NIA in Fig. \ref{fig:appendix_normal}) on MVTec AD.

\begin{table}[!htb]
    \caption{Ablation on the normal image supervision for Refined Mask Prediction.}
    \label{tab:appendix_normal}
    \begin{center}

    \setlength{\tabcolsep}{2pt}{
    \resizebox{0.4\textwidth}{!}{
    \begin{tabular}{c|cccc}
        \toprule
        $F^{res}$  & AUROC & AP & $F_{1}$-max & IoU \\ 
        \midrule
        \multicolumn{1}{l|}{w/o NIA} &96.20&66.03&64.09&53.97 \\
        \multicolumn{1}{l|}{\textbf{with NIA \textbf{(Ours)}}} &\textbf{97.21}&\textbf{69.21}&\textbf{66.37}&\textbf{55.28} \\ 
        \bottomrule
    \end{tabular}
    }
    
    }
\end{center}
\end{table}

\noindent\textbf{Ablation on the Mask Refinement Module}

\begin{figure*}[h]
\centering
\includegraphics[width=0.7\linewidth]{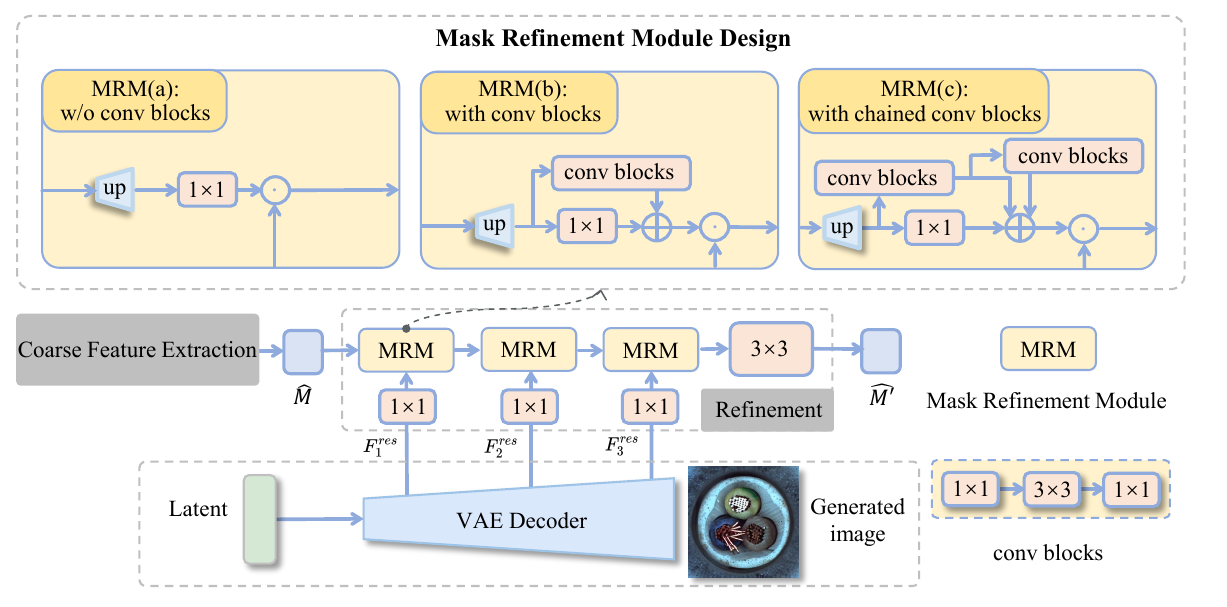}
\vspace{-0.5cm}
\caption{Different structure designs for the mask refinement module in the mask prediction branch.}
\label{fig:appendix_mrm}
\end{figure*}

\begin{figure}[!]
\centering
\includegraphics[width=1.0\linewidth]{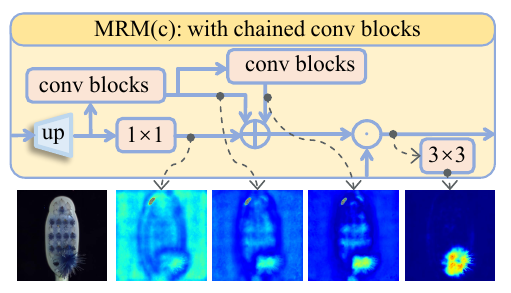}
\vspace{-0.2cm}
\caption{Visualization of the MRM module intermediate results. The top is for the MRM structure diagram, and the bottom is sequentially for the input image, feature maps of the MRM intermediate process and the predicted mask.}
\label{fig:appendix_mrmfeat}
\end{figure}

In the Refined Mask Prediction branch, the Mask Refinement Module (MRM) is utilized to generate refined masks. We devise different structures for MRM, as shown in Fig. \ref{fig:appendix_mrm}, including Case a): those without conv blocks, Case b): with one conv block, and Case c): with chained conv blocks. 
As shown in Fig. \ref{fig:appendix_mrmfeat}, we find that using the conv blocks in Case b), which consists of two 1 $\times$ 1 convolutions and one 3 $\times$ 3 convolution, helps the model learn the features of the defect area more accurately, rather than focusing on the background area for using one convolution alone in Case a). Based on this observation, we further designed a chained conv blocks structure in Case c), and the acquired features better reflect the defect area. This one-level-by-one level of residual learning helps the model achieve better residual correction results for the defect area features. 
As shown in Tab. \ref{tab:appendix_mrm} in the Appendix, Case c) improves the performance by + 0.28\% on AUROC, + 2.29\% on AP and + 2.29\% on $F_1$-max, + 0.32\% on IoU compared with Case b).
We substantiate the superiority of the MRM structures that we design, through the results of downstream supervised segmentation experiments.

\begin{table}[!htb]
    \caption{Ablation on the Mask Refinement Module.}
    \vspace{-0.5cm}
    \label{tab:appendix_mrm}
    \begin{center}
    
    \setlength{\tabcolsep}{2pt}{
    \resizebox{0.4\textwidth}{!}{
    \begin{tabular}{c|cccc}
        \toprule
        Model & AUROC & AP & $F_{1}$-max & IoU \\ 
        \midrule
        with MRM  (a) &96.75&68.18&64.96&\textbf{55.51} \\ 
        with MRM  (b) &96.93&66.92&64.08&54.96 \\  
        \textbf{with MRM (c)} &\textbf{97.21}&\textbf{69.21}&\textbf{66.37}&55.28 \\ 

        \bottomrule
    \end{tabular}
    }
    }
\end{center}
\end{table}

\noindent\textbf{Ablation on the threshold for mask binarization}

In the Refined Mask Prediction branch, we take the threshold $\tau$ for the second channel of refined anomaly masks $\hat{M}_\text{df}^\prime$ to segment the final anomaly mask. We train supervised segmentation models using anomaly masks with $\tau$ settings ranging from 0.1 to 0.5. As shown in Tab. \ref{tab:appendix_threshold}, results indicate that setting $\tau$ = 0.2 yields the best model performance.

\begin{table}[t]
    \caption{Ablation on the threshold for mask binarization.}
    \label{tab:appendix_threshold}
    \begin{center}
    \setlength{\tabcolsep}{2pt}{
    \begin{tabular}{c|cccc}
        \toprule
        threshold & AUROC & AP & $F_{1}$-max & IoU \\ 
        \midrule
        \multicolumn{1}{l|}{$\tau$ = 0.1} &\textbf{97.56}&65.33&63.38&52.40 \\ 
        \multicolumn{1}{l|}{$\tau$ = 0.2 \textbf{(Ours)}}&97.21&\textbf{69.21}&\textbf{66.37}&\textbf{55.28} \\ 
        \multicolumn{1}{l|}{$\tau$ = 0.3} &97.20&66.92&64.35&54.68 \\ 
        \multicolumn{1}{l|}{$\tau$ = 0.4} &95.31&63.55&61.97&53.03 \\ 
        \multicolumn{1}{l|}{$\tau$ = 0.5} &94.11&60.85&59.92&50.87\\
        \bottomrule
    \end{tabular}
    }
\end{center}
\end{table}

\subsection{More qualitative and quantitative anomaly image generation results}
\label{appendix:generation_results}

\noindent\textbf{More detailed quantitative results} 
\label{appendix:generation_metrics}
In this section, we report the detailed generation results for each category on the MVTec AD dataset, VisA dataset, and MVTec 3D AD dataset, which are presented in Tab. \ref{table:appendix_generation}, Tab. \ref{table:appendix_visa_gen}, and Tab. \ref{table:appendix_3dgen}.

\begin{table*}[!t]
\caption{Comparison on IS and IC-LPIPS on MVTec AD. Bold indicates the best performance,
while underlined denotes the second-best result.}
\vspace{-0.5cm}
\label{table:appendix_generation}
\begin{center}
\setlength{\tabcolsep}{2pt}
\resizebox{0.65\linewidth}{!}{
\begin{tabular}{c|cc|cc|cc|cc|cc|cc}
        \specialrule{0.6pt}{0.1pt}{0pt}

        \multirow{2}{*}{Category} & \multicolumn{2}{c|}{\makecell{Crop\&\\Paste} \citep{ICME2021crop-and-paste}} & \multicolumn{2}{c|}{SDGAN \citep{niu2020sdgan}} & \multicolumn{2}{c|}{\makecell{Defect-\\GAN}\citep{zhang2021defectgan} } & \multicolumn{2}{c|}{DFMGAN \citep{Duan2023DFMGAN}} & \multicolumn{2}{c|}{\makecell{Anomaly\\Diffusion} \citep{hu2023anomalydiffusion}} & \multicolumn{2}{c}{Ours}  \\

        & \multicolumn{1}{c} {IS $\uparrow$} &  IC-L $\uparrow$  &{IS $\uparrow$} & IC-L $\uparrow$ &{IS $\uparrow$} & IC-L $\uparrow$ &{IS $\uparrow$} & IC-L $\uparrow$ & IS $\uparrow$ & IC-L $\uparrow$ &{IS $\uparrow$} & IC-L $\uparrow$\\
         
        \hline
        \hline
        bottle   &	1.43  &	 0.04   &	1.57  &	0.06  &	1.39  &	0.07  &	\underline{1.62}  &	0.12  &	1.58  &	\underline{0.19}  &	\textbf{1.78}  &	\textbf{0.21}  \\
        cable  &	1.74  &	0.25   &	1.89  &	0.19  &	1.70  &	0.22  &	1.96  &	0.25  &	\textbf{2.13}  &	\underline{0.41}  &	\underline{2.09}  &	\textbf{0.42}  \\
        capsule  &	1.23  &	0.05    &	1.49  &	0.03  &	\textbf{1.59}  &	0.04  &	\textbf{1.59}  &	0.11  &	\textbf{1.59}  &	\underline{0.21}  &	\underline{1.56}  &	\textbf{0.26}   \\
        carpet &	1.17  &	0.11   &	1.18  &	0.11  &	\textbf{1.24}  &	0.12  &	\underline{1.23}  &	0.13  &	1.16  &	\underline{0.24}  &	1.13 &	\textbf{0.25}    \\
        grid &	2.00  &	0.12    & 1.95 &   0.10 &	2.01  &	0.12  &	1.97  &	\underline{0.13}  &	\underline{2.04}   &	\textbf{0.44}  &	\textbf{2.43}  &	  \textbf{0.44}  \\
        hazelnut &	1.74  &	0.21    &	1.85  &	0.16  &	1.87  &	0.19  &	\underline{1.93}  &	\underline{0.24}  &	\textbf{2.13}  &	\textbf{0.31}  &	1.87  &	\textbf{0.31}    \\
        leather &	1.47  &	0.14    &	2.04  &	0.12  &	\textbf{2.12}  &	0.14  &	\underline{2.06}  &	0.17  &	1.94  &	\textbf{0.41}  &	2.03  &	\underline{0.40}   \\
        metal\_nut & 	1.56  &	0.15    & 1.45  &	0.28  &	1.47  &	0.30  &	1.49  &	\textbf{0.32}  &	\textbf{1.96}  &	0.30  &	\underline{1.64}  &	\underline{0.31}    \\
        pill &	1.49  &	0.11    &	1.61  &	0.07  &	1.61  &	0.10  &	\textbf{1.63}  &	0.16  &	1.61  &	\underline{0.26}  &	\underline{1.62}  &	\textbf{0.33}    \\
        screw &	1.12  &	0.16    &	1.17  &	0.10  &	1.19  &	0.12  &	1.12  &	0.14  &	\underline{1.28}  &	\underline{0.30}  &	\textbf{1.52}  &	\textbf{0.31}    \\
        tile &	1.83  &	0.20    &	2.53  &	0.21  &	2.35  &	0.22  &	2.39  & 0.22  &	\underline{2.54}  &	\textbf{0.55}  &	\textbf{2.60}  & \underline{0.50}   \\
        toothbrush &	1.30  &	0.08    &	1.78  &	0.03  &	\underline{1.85}  &	0.03  &	1.82  &	0.18  &	1.68  &	\underline{0.21}  &	\textbf{1.96}  &	\textbf{0.25}    \\
        transistor &		1.39  &	0.15  & \textbf{1.76}  &	0.13  &	1.47  &	0.13    &	\underline{1.64}  &	\underline{0.25}  &	1.57  &	\textbf{0.34}  &	1.51  &	\textbf{0.34}    \\
        wood &	1.95  &	0.23    &	2.12  &	0.25  &	2.19  &	0.29  &	2.12  &	0.35  &	\underline{2.33}  &	\underline{0.37}  &	\textbf{2.77}  &	\textbf{0.46}    \\
        zipper &	1.23  &	0.11    &	1.25  &	0.10  &	1.25  &	0.10  &	1.29  &	\underline{0.27}  &	\underline{1.39}  &	0.25  &	\textbf{1.63}  &	\textbf{0.30}    \\
        \hline
        Average  &		1.51  &	0.14    &	1.71  &	0.13  &	1.69  &	0.15  &	1.72  &	0.20  &	\underline{1.80}  &	\underline{0.32}  &	\textbf{1.88}  &	\textbf{0.34}   \\
        \specialrule{0.8pt}{0.1pt}{0pt}
    \end{tabular}
}
\vspace{-0.5cm}
\end{center}
\end{table*}

\begin{table}[t!]
\caption{Comparison on IS and IC-LPIPS on VisA. Bold indicates the best performance.}
\label{table:appendix_visa_gen}
\begin{center}
\setlength{\tabcolsep}{2pt}
\resizebox{0.8\linewidth}{!}{
\begin{tabular}{c|cc|cc|cc}
        \specialrule{0.6pt}{0.1pt}{0pt}

        \multirow{2}{*}{Category} & \multicolumn{2}{c|}{\makecell{DFMGAN \\ \citep{Duan2023DFMGAN}}} & \multicolumn{2}{c|}{\makecell{AnomalyDiffusion \\ \citep{hu2023anomalydiffusion}}} & \multicolumn{2}{c}{Ours}  \\

        & \multicolumn{1}{c}{IS $\uparrow$} & IC-L $\uparrow$ & {IS $\uparrow$} &  IC-L $\uparrow$  &{IS $\uparrow$} & IC-L $\uparrow$ \\
         
        \hline
        \hline
        candle   & 1.19&  \textbf{0.23}&  \textbf{1.28}&  0.17&    1.20&  0.12 \\
        capsules  &	1.25&	0.22&	1.39&	0.50&	\textbf{1.58}&	\textbf{0.60}   \\
        cashew  &	1.25&	0.24&	\textbf{1.27} &	0.26&	1.21&	\textbf{0.28}     \\
        chewinggum &	\textbf{1.33}&	0.24&	1.15&	0.19&	1.29&	\textbf{0.27}      \\
        fryum &	\textbf{1.28}&	0.20&	1.20&	0.14&	1.14&	\textbf{0.21}   \\
        macaroni1 &	1.14&	\textbf{0.24}&	\textbf{1.15}&	0.14&	\textbf{1.15}&	0.18     \\
        macaroni2 &	1.47&	0.38&	1.56&	0.38&	\textbf{1.57}&	\textbf{0.39}     \\
        pcb1 & 1.12&	0.16&	\textbf{1.18}&	\textbf{0.35}&	\textbf{1.18}&	0.26      \\
        pcb2 &	1.12 &	0.26&	\textbf{1.26}&	0.21&	1.25&	\textbf{0.27}      \\
        pcb3 &	1.19&	0.18&	1.21&	\textbf{0.24}&	\textbf{1.22} &	0.21     \\
        pcb4 &	\textbf{1.21}&	\textbf{0.28}&	1.14&	0.25&	1.15&	0.22    \\
        pipe\_fryum &	\textbf{1.43} &	\textbf{0.32}&	1.29&	0.17&	1.31&	0.16     \\
        \hline
        Average  &	1.25&	0.25&	1.26&	0.25&	\textbf{1.27}&	\textbf{0.26}   \\
        \specialrule{0.8pt}{0.1pt}{0pt}
    \end{tabular}
}
\vspace{-0.5cm}
\end{center}
\end{table}

\begin{table}[t!]
\caption{Comparison on IS and IC-LPIPS on MVTec 3D AD. Bold indicates the best performance.}
\label{table:appendix_3dgen}
\begin{center}
\setlength{\tabcolsep}{2pt}
\resizebox{0.8\linewidth}{!}{
\begin{tabular}{c|cc|cc|cc}
        \specialrule{0.6pt}{0.1pt}{0pt}

        \multirow{2}{*}{Category} & \multicolumn{2}{c|}{\makecell{DFMGAN \\ \citep{Duan2023DFMGAN}}} & \multicolumn{2}{c|}{\makecell{AnomalyDiffusion \\ \citep{hu2023anomalydiffusion}}} & \multicolumn{2}{c}{Ours}  \\

        & \multicolumn{1}{c}{IS $\uparrow$} & IC-L $\uparrow$ & {IS $\uparrow$} &  IC-L $\uparrow$  &{IS $\uparrow$} & IC-L $\uparrow$ \\
         
        \hline
        \hline
        bagel   & 1.07&  0.26&  1.02&  0.22&    \textbf{1.28}&  \textbf{0.29}    \\
        cable\_gland  &	1.59&	\textbf{0.25}&	1.79&	0.19&	\textbf{2.21}&	0.19    \\
        carrot  &	1.94&	\textbf{0.29}&	1.66&	0.17&	\textbf{2.07}&	0.22     \\
        cookie &	1.80&	0.31&	1.77&	0.29&	\textbf{2.07}&	\textbf{0.38}      \\
        dowel &	\textbf{1.96}&	\textbf{0.37}&	1.60&	0.20&	1.95&	0.26   \\
        foam &	1.50&	0.17&	1.77&	0.30&	\textbf{2.20}&	\textbf{0.39}      \\
        peach &	2.11&	\textbf{0.34}&	1.91&	0.23&	\textbf{2.40}&	0.28     \\
        potato & \textbf{3.05}&	\textbf{0.35}&	1.92&	0.17&	1.98&	0.22      \\
        rope &	1.46&	0.29&	1.28&	0.25&	\textbf{1.53}&	\textbf{0.41}      \\
        tire &	1.53&	0.25&	1.35&	0.20&	\textbf{1.81}&	\textbf{0.31}      \\
        \hline
        Average  &	1.80&	0.29&	1.61&	0.22&	\textbf{1.95}&	\textbf{0.30}   \\
        \specialrule{0.8pt}{0.1pt}{0pt}
    \end{tabular}
}
\vspace{-0.5cm}
\end{center}
\end{table}

\noindent\textbf{More qualitative generation results}

\begin{figure*}[t]
\centering
\includegraphics[width=0.7\linewidth]{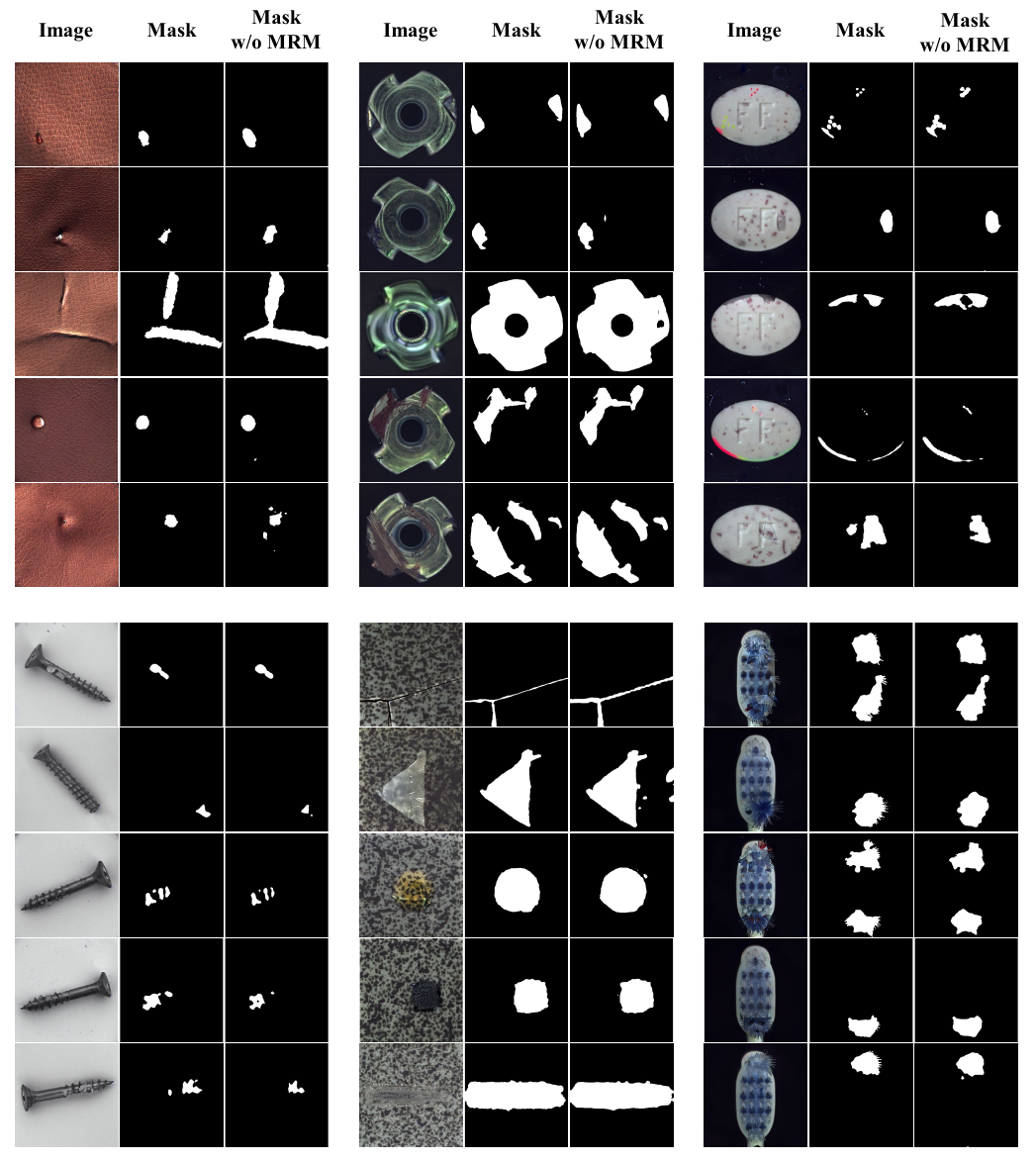}
\vspace{-0.5cm}
\caption{Qualitative results of our anomaly image generation results on MVTec AD.
In the first row, from left to right, are the results for \emph{leather}, \emph{metal\_nut}, and \emph{pill} categories. In the
second row, from left to right, are the results for \emph{screw}, \emph{tile}, and \emph{toothbrush} categories.}
\label{Fig:appendix_gen2}
\end{figure*}

\begin{figure*}[h]
\centering
\includegraphics[width=0.7\linewidth]{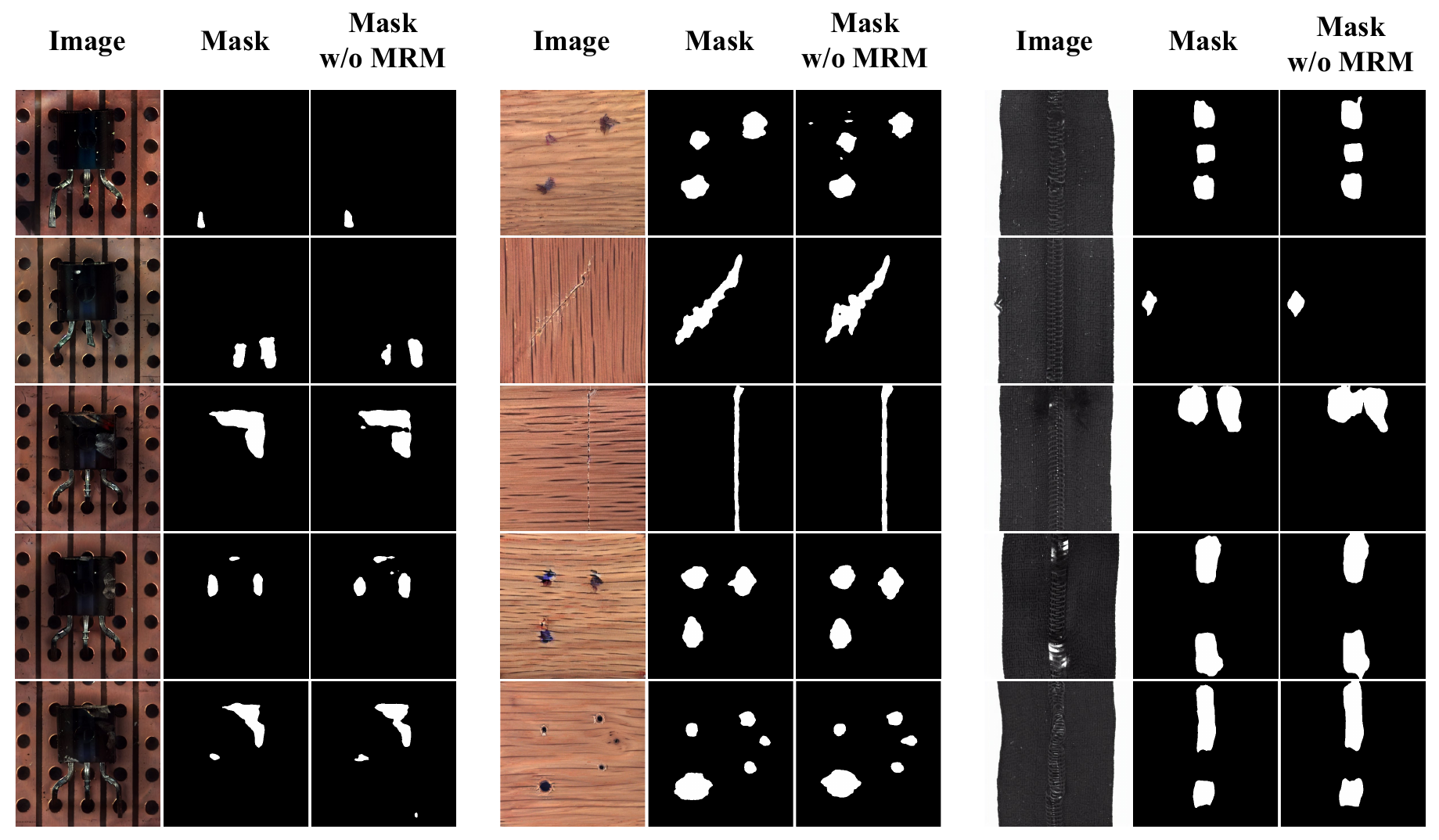}
\caption{Qualitative results of our anomaly image generation results on MVTec AD.
In the first row, from left to right, are the results for \emph{transistor}, \emph{wood}, and \emph{zipper} categories. }
\label{Fig:appendix_gen3}
\vspace{-0.5cm}
\end{figure*}

We provide further qualitative results of every category on the MVTec AD dataset, from Fig. \ref{Fig:appendix_gen2} to Fig. \ref{Fig:appendix_gen3}. We report the anomaly image generation results of SeaS for varying types of anomalies. The first column represents the generated anomaly images, the second column represents the corresponding generated masks, and the third column represents the masks generated without using the Mask Refinement Module.

\begin{figure*}[h]
\centering
\includegraphics[width=0.7\linewidth]{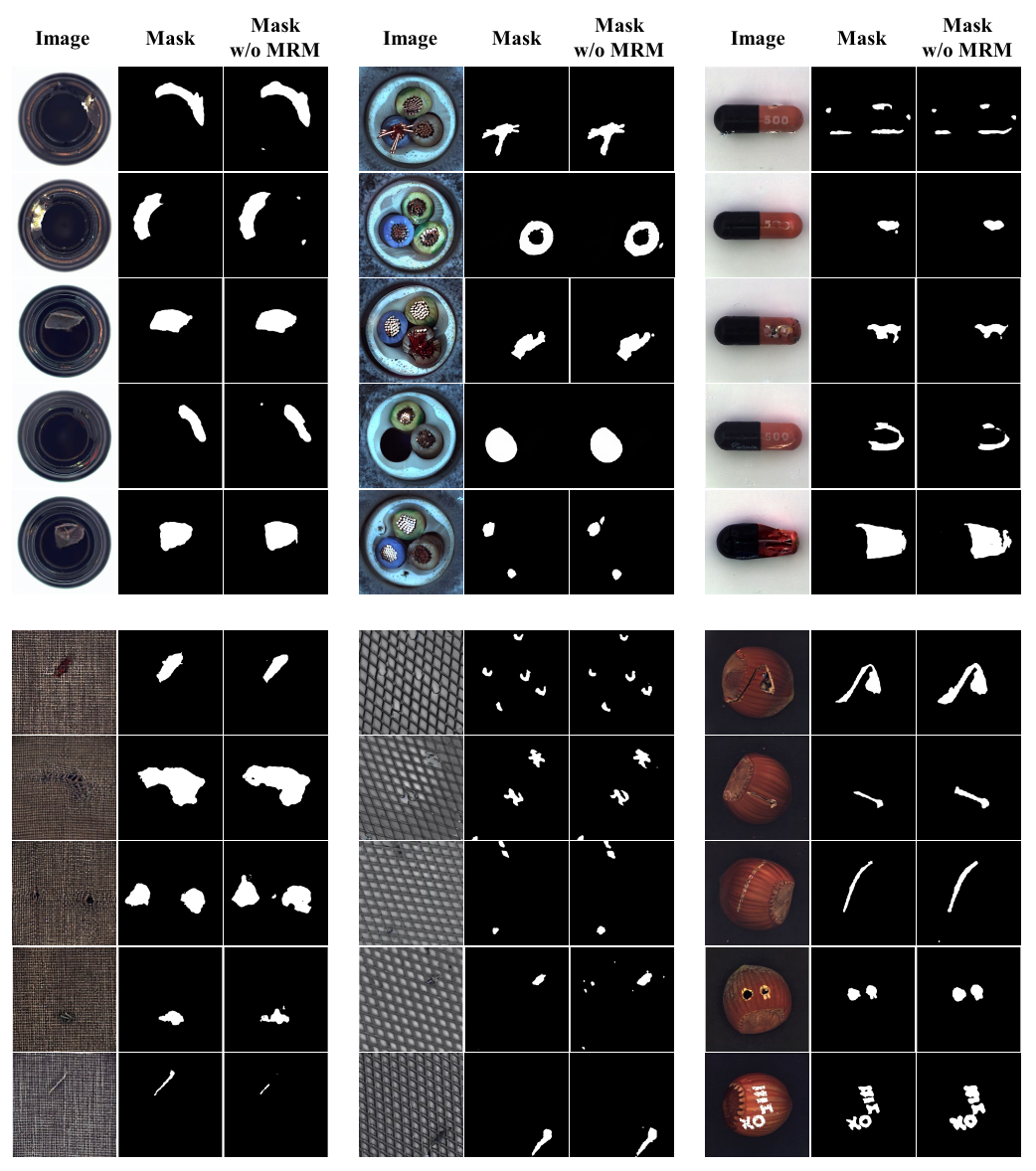}

\caption{Qualitative results of our anomaly image generation results on MVTec AD.
In the first row, from left to right, are the results for \emph{bottle}, \emph{cable}, and \emph{capsule} categories. In the
second row, from left to right, are the results for \emph{carpet}, \emph{grid}, and \emph{hazelnut} categories.}
\label{Fig:appendix_gen1}
\vspace{-0.5cm}
\end{figure*}

\begin{figure*}[h]
\centering
\includegraphics[width=1.0\linewidth]{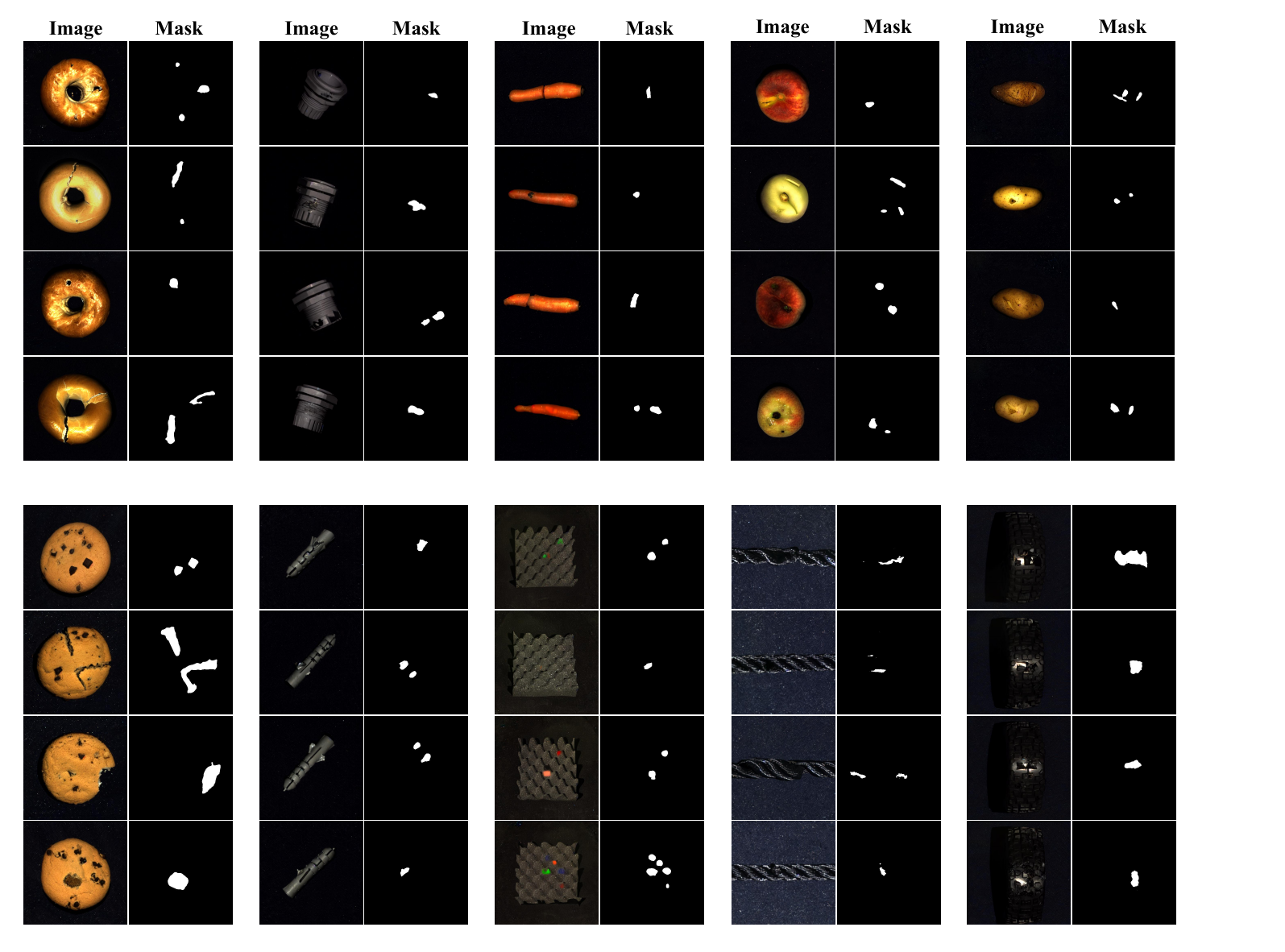}
\vspace{-0.3cm}
\caption{Qualitative results of our anomaly image generation results on MVTec 3D AD.
In the first row, from left to right, are the results for \emph{bagel}, \emph{cable\_gland}, \emph{carrot}, \emph{peach}, and \emph{potato} categories. In the
second row, from left to right, are the results for \emph{cookie}, \emph{dowel}, \emph{foam}, \emph{rope}, and \emph{tire} categories.}
\label{Fig:appendix_3dgen_all}
\vspace{-0.5cm}
\end{figure*}

We provide further qualitative results of every category on the MVTec 3D AD dataset in Fig. \ref{Fig:appendix_3dgen_all}. We report the anomaly image generation results of SeaS for varying types of anomalies. The first column represents the generated anomaly images, and the second column represents the corresponding generated masks.

\subsection{More qualitative and quantitative comparison results of supervised segmentation models trained on image-mask pairs generated by different anomaly generation methods}
\label{appendix:seg_result1}

\begin{figure*}[htb]
\centering
\includegraphics[width=0.67\linewidth]{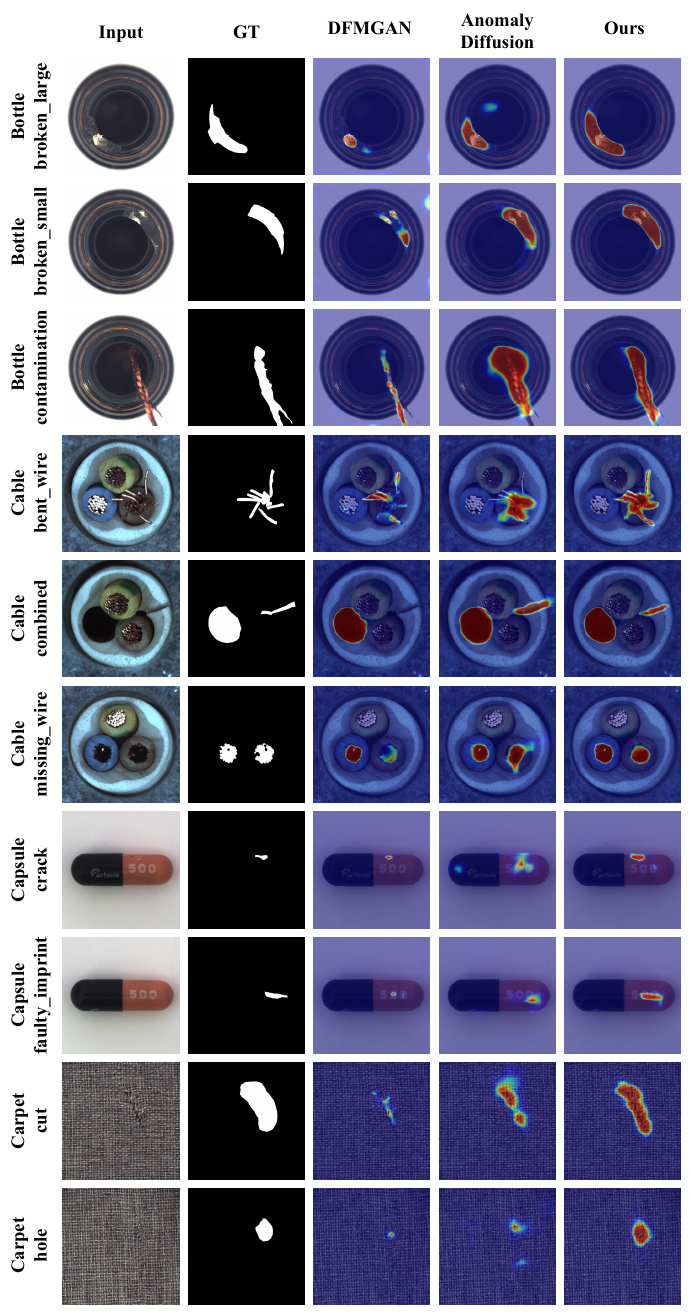}
\vspace{-0.3cm}
\caption{Comparison results with the anomaly supervised segmentation model BiSeNet V2 on MVTec AD. In the figure, from top to bottom are the results for \emph{bottle}, \emph{cable}, \emph{capsule} and \emph{carpet} categories.}
\label{fig:appendix_seg1}
\end{figure*}

\begin{figure*}[htb]
\centering
\includegraphics[width=0.67\linewidth]{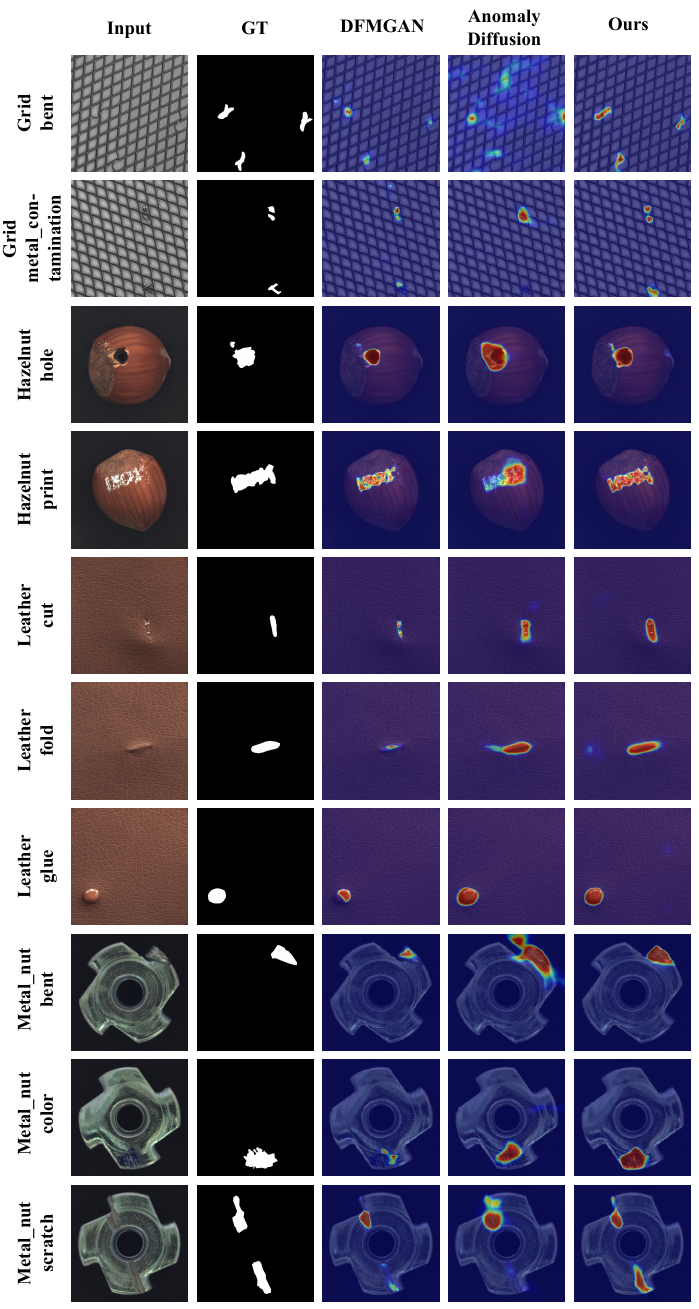}
\caption{Comparison results with the anomaly supervised segmentation model BiSeNet V2 on MVTec AD. In the figure, from top to bottom are the results for \emph{grid}, \emph{hazelnut}, \emph{leather} and \emph{metal\_nut} categories.}
\label{fig:appendix_seg2}
\end{figure*}

\begin{figure*}[htb]
\centering
\includegraphics[width=0.67\linewidth]{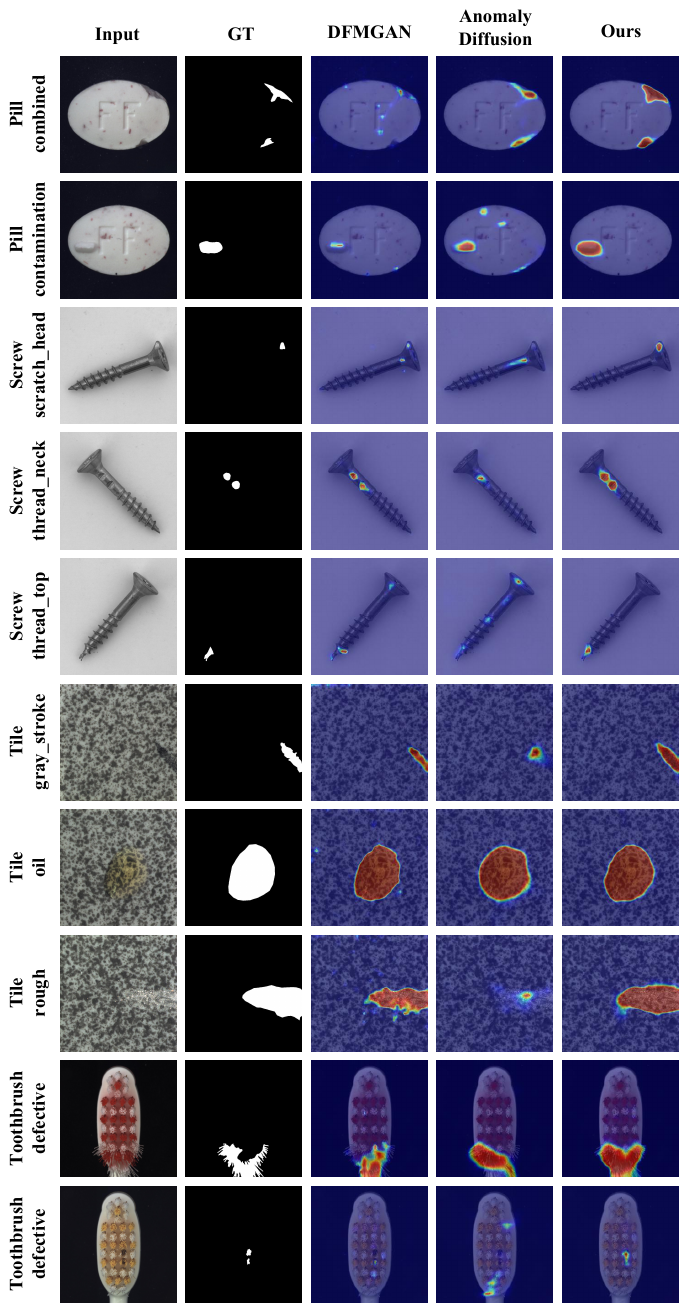}
\vspace{-0.3cm}
\caption{Comparison results with the anomaly supervised segmentation model BiSeNet V2 on MVTec AD. In the figure, from top to bottom are the results for \emph{pill}, \emph{screw}, \emph{tile} and \emph{toothbrush} categories.}
\label{fig:appendix_seg3}
\end{figure*}

\begin{figure*}[htb]
\centering
\includegraphics[width=0.67\linewidth]{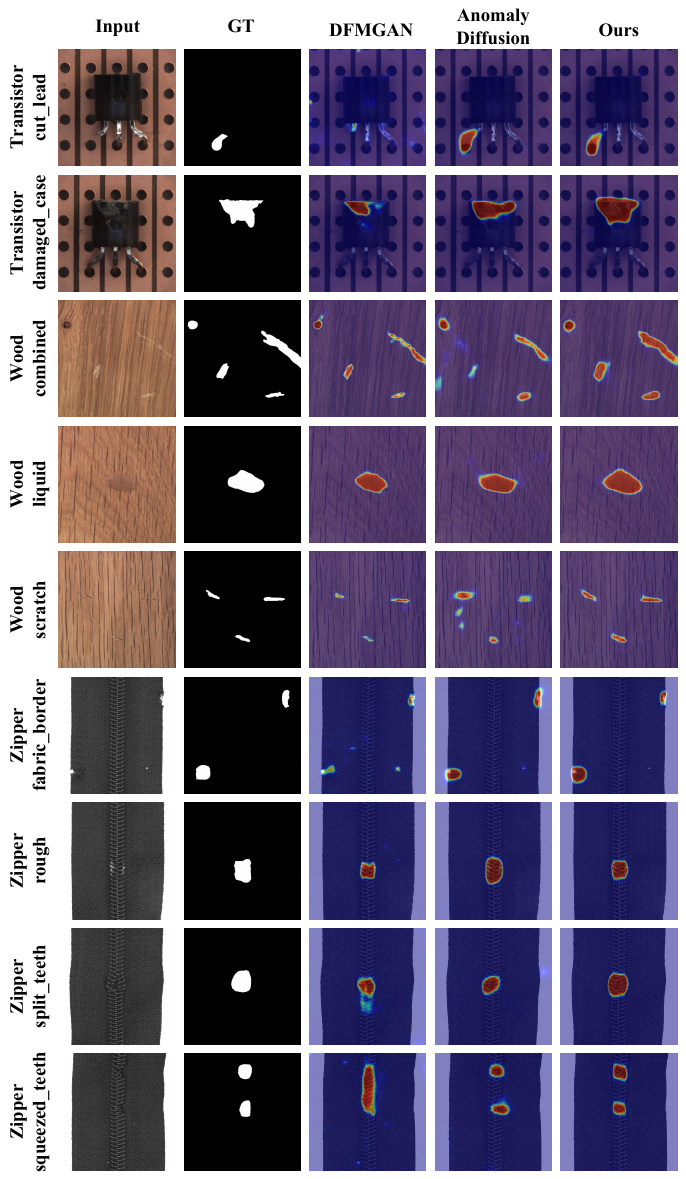}
\caption{Comparison results with the anomaly supervised segmentation model BiSeNet V2 on MVTec AD. In the figure, from top to bottom are the results for \emph{transistor}, \emph{wood} and \emph{zipper} categories.}
\label{fig:appendix_seg4}
\end{figure*}

We provide further qualitative results with different anomaly generation methods on the MVTec AD dataset. We report the generation results of SeaS for varying types of anomalies in each category. Results are from Fig. \ref{fig:appendix_seg1} to Fig. \ref{fig:appendix_seg4}.

\begin{figure*}[htb]
\centering
\includegraphics[width=0.8\linewidth]{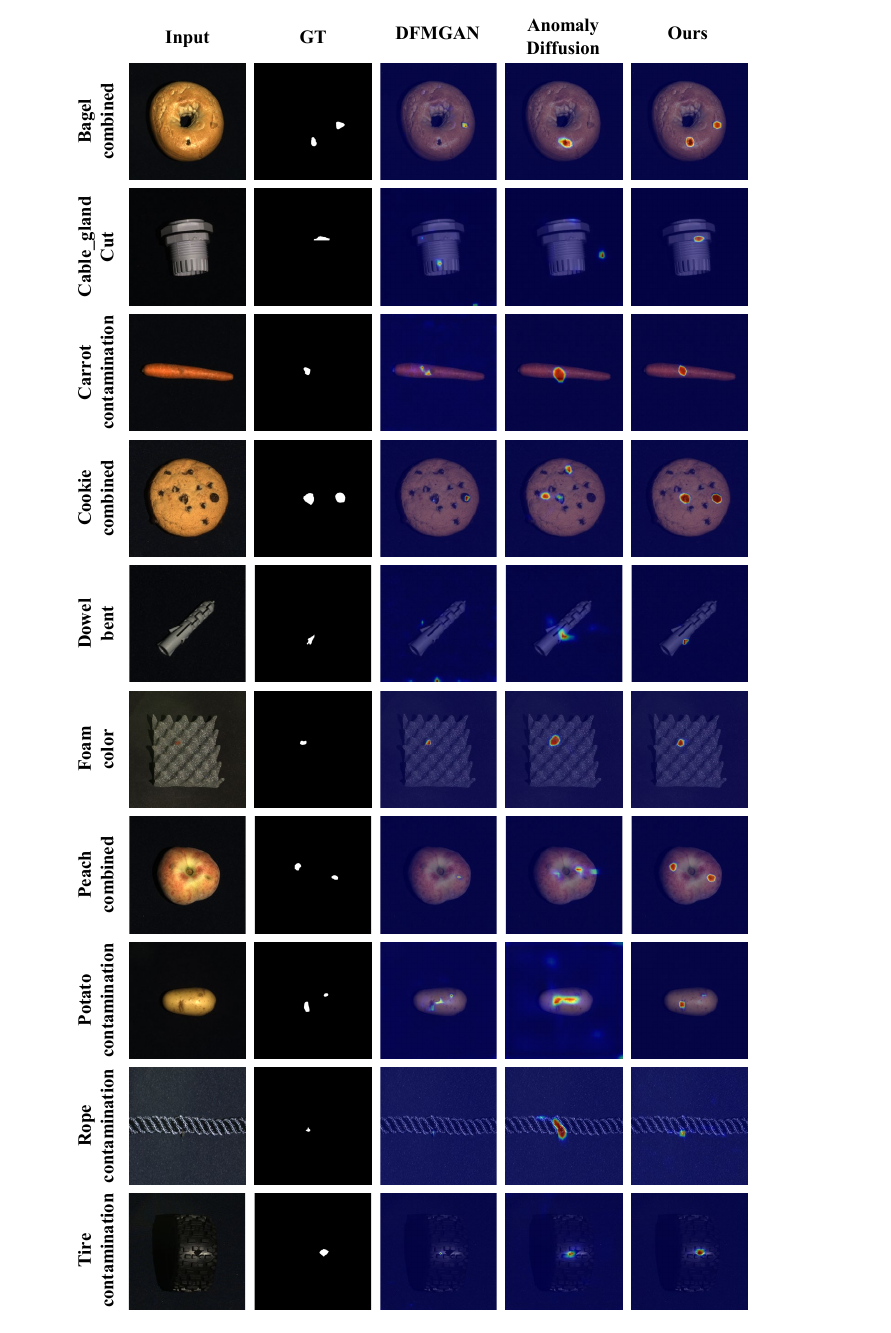}
\caption{Qualitative supervised anomaly segmentation results with BiSeNet V2 on MVTec 3D AD.}
\label{Fig:appendix_3dseg_ap1}
\vspace{-0.5cm}
\end{figure*}

We provide further qualitative comparisons on downstream supervised segmentation trained by the generated images. The segmentation anomaly maps are shown in Fig. \ref{Fig:appendix_3dseg_ap1}. There are fewer false positives (e.g., \emph{potato\_combined}) and fewer false negatives (e.g., \emph{bagel\_contamination}), when the BiSeNet V2 is trained on the image-mask pairs generated by our method.

\clearpage

\clearpage

\clearpage

\clearpage

\clearpage

We report the detailed segmentation results of SeaS for each category on the MVTec AD datasets, compared with DFMGAN \citep{Duan2023DFMGAN} and AnomalyDiffusion \citep{hu2023anomalydiffusion}, which are presented from Tab. \ref{tab:pixel_BiSeNetall} to Tab. \ref{tab:image_ufnetall}

\begin{table*}[htb]
\caption{Comparison on supervised anomaly segmentation on BiSeNet V2.}
\label{tab:pixel_BiSeNetall}
\begin{center}
\setlength{\tabcolsep}{2pt}
\resizebox{0.8\linewidth}{!}{
\begin{tabular}{c|cccc|cccc|cccc}
    \specialrule{0.6pt}{0.1pt}{0pt}
    
    \multirow{2}{*}{Category} & \multicolumn{4}{c|}{DFMGAN} & \multicolumn{4}{c|}{AnomalyDiffusion} & \multicolumn{4}{c}{Ours}  \\
     &AUROC &  AP & $F_1$-max& IoU &AUROC &  AP & $F_1$-max& IoU &AUROC &  AP & $F_1$-max& IoU \\
    \hline
    \hline
    bottle &  89.34&64.67&62.78&44.71&99.00&88.02&80.53&68.25&\textbf{99.46}&\textbf{93.43}&\textbf{85.59}&\textbf{75.86} \\
    cable & \textbf{93.87}&67.98&64.74&44.02&92.84&69.86&66.32&46.49&89.85&\textbf{72.07}&\textbf{71.58}&\textbf{53.24} \\
    capsule& 74.88&16.43&23.01&29.97&\textbf{92.71}&\textbf{38.11}&\textbf{40.67}&19.44&86.33&24.64&30.54&\textbf{39.70} \\
    carpet  &94.53&42.53&47.44&39.88&98.65&73.10&65.83&43.25&\textbf{99.61}&\textbf{82.30}&\textbf{72.94}&\textbf{55.52} \\
    grid  &96.86&24.40&37.40&29.93&80.59&8.08&16.79&14.26&\textbf{99.36}&\textbf{37.91}&\textbf{42.50}&\textbf{39.80} \\
    hazelnut  & \textbf{99.87}&\textbf{96.75}&\textbf{90.07}&\textbf{71.68}&97.71&63.34&59.87&43.12&97.82&78.55&73.09&68.47 \\
    leather  &97.50&51.10&52.26&\textbf{50.67}&\textbf{99.30}&57.49&\textbf{59.62}&43.94&98.91&\textbf{59.84}&58.62&45.82 \\
    metal\_nut  &99.39&97.59&92.52&70.40&99.03&95.67&88.69&58.8&\textbf{99.69}&\textbf{98.29}&\textbf{93.23}&\textbf{74.40} \\
    pill  &97.09&83.98&79.26&36.39&\textbf{99.44}&\textbf{93.16}&\textbf{86.62}&41.18&98.31&76.97&68.00&\textbf{55.43} \\
    screw  &\textbf{97.94}&37.10&41.01&31.63&94.08&17.95&25.90&20.00&97.64&\textbf{40.20}&\textbf{45.35}&\textbf{38.43} \\
    tile  &99.65&97.08&91.16&\textbf{75.94}&97.79&85.58&78.28&60.46&\textbf{99.67}&\textbf{97.29}&\textbf{91.48}&75.75 \\
    toothbrush  &97.70&\textbf{51.32}&54.05&23.38&\textbf{98.43}&49.64&\textbf{54.08}&26.53&97.15&46.09&49.02&\textbf{28.56} \\
    transistor  & 84.31&45.34&46.07&30.00&\textbf{98.85}&\textbf{85.27}&\textbf{77.95}&49.83&96.75&69.52&66.11&\textbf{57.24} \\
    wood  & 98.32&64.82&63.11&\textbf{58.99}&96.78&63.38&60.31&45.73&\textbf{98.38}&\textbf{80.81}&\textbf{74.03}&56.22 \\
    zipper  &97.29&65.18&63.24&49.93&98.81&78.89&72.66&62.03&\textbf{99.23}&\textbf{80.27}&\textbf{73.41}&\textbf{64.80} \\
    \hline
    Average  & 94.57&60.42&60.54&45.83&96.27&64.5&62.27&42.89&\textbf{97.21}&\textbf{69.21}&\textbf{66.37}&\textbf{55.28}
\\
    \specialrule{0.8pt}{0.1pt}{0pt}
    \end{tabular}
}
\vspace{-0.3cm}
\end{center}
\end{table*}

\begin{table*}[htb]
\caption{Comparison on image-level anomaly detection on BiSeNet V2.}
\label{tab:image_BiSeNet}
\begin{center}
\setlength{\tabcolsep}{2pt}
\resizebox{0.72\linewidth}{!}{
\begin{tabular}{c|ccc|ccc|ccc}
    \specialrule{0.6pt}{0.1pt}{0pt}
    
    \multirow{2}{*}{Category}  & \multicolumn{3}{c|}{DFMGAN} & \multicolumn{3}{c|}{AnomalyDiffusion} & \multicolumn{3}{c}{Ours}  \\
    &AUROC &  AP & $F_1$-max &AUROC &  AP & $F_1$-max &AUROC &  AP & $F_1$-max \\
    \hline
    \hline
    bottle &   96.74&98.75&95.35&98.14&99.34&97.67&\textbf{100.00}&\textbf{100.00}&\textbf{100.00}  \\
    cable  &  79.47&85.00&74.13&\textbf{95.37}&\textbf{96.71}&\textbf{92.91}&94.61&96.39&89.83 
 \\
    capsule  &  85.51&95.16&\textbf{89.82}&84.06&95.01&89.74&\textbf{88.81}&\textbf{96.92}&89.21 
\\
    carpet  &  91.42&96.29&88.89&90.55&96.41&90.32&\textbf{98.16}&\textbf{99.31}&\textbf{97.56} 
 \\
    grid  &  \textbf{99.64}&\textbf{99.82}&97.56&81.19&89.92&83.95&99.17&99.63&\textbf{98.73} 
 \\
    hazelnut  & \textbf{100.00}&\textbf{100.00}&\textbf{100.00}&93.39&95.74&90.91&\textbf{100.00}&\textbf{100.00}&\textbf{100.00}
 \\
    leather  &  98.31&99.23&95.24&\textbf{100.00}&\textbf{100.00}&\textbf{100.00}&95.83&98.38&95.93 
 \\
    metal\_nut  & 97.37&99.16&94.66&99.01&99.66&97.71&\textbf{100.00}&\textbf{100.00}&\textbf{100.00} 
 \\
    pill  &  84.86&95.27&91.00&90.38&97.43&91.35&\textbf{96.59}&\textbf{99.12}&\textbf{95.24} 
\\
    screw  &  74.95&85.50&80.72&58.18&75.32&\textbf{81.25}&\textbf{77.24}&\textbf{89.55}&80.60 
\\
    tile  &  99.47&99.74&99.12&98.78&99.44&97.39&\textbf{100.00}&\textbf{100.00}&\textbf{100.00} 
 \\
    toothbrush  &  78.33&87.73&83.72&78.33&89.26&79.17&\textbf{90.42}&\textbf{94.49}&\textbf{89.47} 
 \\
    transistor  &  79.52&75.77&69.57&94.40&94.68&94.34&\textbf{99.23}&\textbf{98.39}&\textbf{94.92} 
 \\
    wood  &  98.87&99.46&97.67&90.48&94.12&93.33&\textbf{100.00}&\textbf{100.00}&\textbf{100.00}
 \\
    zipper  & 98.97&99.64&97.56&98.89&99.62&97.56&\textbf{100.00}&\textbf{100.00}&\textbf{100.00} 
 \\
    \hline
    Average  &  90.90&94.43&90.33&90.08&94.84&91.84&\textbf{96.00}&\textbf{98.14}&\textbf{95.43} 
\\
    \specialrule{0.8pt}{0.1pt}{0pt}
    \end{tabular}
}
\vspace{-0.3cm}
\end{center}
\end{table*}

\begin{table*}[htb]
\caption{Comparison on supervised anomaly segmentation on UPerNet.}
\label{tab:pixel_upernetall}
\begin{center}
\setlength{\tabcolsep}{2pt}
\resizebox{0.8\linewidth}{!}{
\begin{tabular}{c|cccc|cccc|cccc}
    \specialrule{0.6pt}{0.1pt}{0pt}
    
    \multirow{2}{*}{Category} & \multicolumn{4}{c|}{DFMGAN} & \multicolumn{4}{c|}{AnomalyDiffusion} & \multicolumn{4}{c}{Ours}  \\
    & AUROC &  AP & $F_1$-max& IoU &AUROC &  AP & $F_1$-max& IoU &AUROC &  AP & $F_1$-max& IoU \\
    \hline
    \hline
    bottle   & 87.94&56.89&56.56&45.41&\textbf{99.54}&\textbf{93.01}&\textbf{85.94}&75.31&99.28&91.73&84.53&\textbf{78.73} 
\\
    cable  & 87.52&64.30&65.61&41.02&91.00&68.12&67.49&51.84&\textbf{91.08}&\textbf{76.25}&\textbf{74.63}&\textbf{59.00} 
\\
    capsule  & 67.92&12.31&20.32&30.47&\textbf{97.64}&\textbf{51.90}&\textbf{51.66}&37.00&92.09&39.60&43.89&\textbf{50.18} 
\\
    carpet  & 95.85&36.05&34.52&48.10&99.45&\textbf{82.13}&72.55&53.17&\textbf{99.67}&82.01&\textbf{73.53}&\textbf{60.60} 
\\
    grid  &  97.49&29.67&36.15&31.37&94.22&28.97&38.50&32.93&\textbf{99.18}&\textbf{44.94}&\textbf{48.28}&\textbf{44.21} 
\\
    hazelnut  & 99.36&79.76&71.10&72.90&97.77&70.48&67.93&54.47&\textbf{99.54}&\textbf{81.84}&\textbf{75.48}&\textbf{73.30} 
\\
    leather  & 80.97&17.60&26.21&30.17&\textbf{99.48}&63.46&60.54&48.70&99.42&\textbf{68.26}&\textbf{65.52}&\textbf{57.01} 
\\
    metal\_nut  & 98.44&95.64&91.48&64.92&98.62&95.11&88.62&61.31&\textbf{99.70}&\textbf{98.33}&\textbf{92.90}&\textbf{76.07} 
\\
    pill  &  97.58&83.74&80.02&42.33&\textbf{99.33}&\textbf{95.04}&\textbf{88.77}&49.18&98.59&81.16&74.26&\textbf{62.62} 
\\
    screw  & 97.49&\textbf{53.83}&\textbf{53.02}&42.05&93.89&36.60&42.68&34.08&\textbf{98.97}&52.02&51.65&\textbf{46.61} 
\\
    tile  & \textbf{99.79}&\textbf{97.29}&\textbf{91.11}&77.46&94.70&73.34&67.79&58.54&99.67&95.89&90.71&\textbf{77.89} 
\\
    toothbrush  & 97.42&51.09&59.23&28.33&97.52&60.67&59.46&33.98&\textbf{98.50}&\textbf{63.62}&\textbf{63.07}&\textbf{42.09} 
\\
    transistor  & 82.07&36.31&39.48&27.44&\textbf{94.26}&\textbf{73.68}&\textbf{69.50}&53.64&93.88&70.37&68.12&\textbf{56.98} 
\\
    wood  & 97.90&69.02&62.21&63.10&96.09&70.10&64.38&51.44&\textbf{99.28}&\textbf{85.28}&\textbf{76.28}&\textbf{65.09} 
\\
    zipper  & 97.28&71.60&66.64&54.54&\textbf{99.54}&\textbf{86.18}&\textbf{78.50}&66.47&99.17&85.01&77.57&\textbf{68.21} 
\\
    \hline
    Average  & 92.33&57.01&56.91&46.64&96.87&69.92&66.95&50.80&\textbf{97.87}&\textbf{74.42}&\textbf{70.70}&\textbf{61.24} 
\\
    \specialrule{0.8pt}{0.1pt}{0pt}
    \end{tabular}
}
\vspace{-0.3cm}
\end{center}
\end{table*}

\begin{table*}[htb]
\caption{Comparison on image-level anomaly detection on UPerNet.}
\label{tab:image_upernet}
\begin{center}
\setlength{\tabcolsep}{2pt}
\resizebox{0.72\linewidth}{!}{
\begin{tabular}{c|ccc|ccc|ccc}
    \specialrule{0.6pt}{0.1pt}{0pt}
    
    \multirow{2}{*}{Category} & \multicolumn{3}{c|}{DFMGAN} & \multicolumn{3}{c|}{AnomalyDiffusion} & \multicolumn{3}{c}{Ours}  \\
    & AUROC &  AP & $F_1$-max &AUROC &  AP & $F_1$-max &AUROC &  AP & $F_1$-max \\
    \hline
    \hline
    bottle   & 94.19&97.86&93.18&\textbf{100.00}&\textbf{100.00}&\textbf{100.00}&\textbf{100.00}&\textbf{100.00}&\textbf{100.00}
 \\
    cable  & 85.64&90.03&80.33&\textbf{95.58}&\textbf{97.06}&\textbf{92.56}&94.40&96.38&92.44 
\\
    capsule  &  81.04&94.26&87.01&\textbf{96.00}&\textbf{98.77}&\textbf{95.48}&94.43&98.44&92.21 
 \\
    carpet  &  96.72&98.58&93.75&98.68&99.53&98.36&\textbf{99.94}&\textbf{99.97}&\textbf{99.20} 
\\
    grid  &  98.33&99.13&96.30&96.67&98.73&97.44&\textbf{99.76}&\textbf{99.88}&\textbf{98.73} 
 \\
    hazelnut  &  99.84&99.87&97.96&99.17&99.43&97.87&\textbf{100.00}&\textbf{100.00}&\textbf{100.00} 
 \\
    leather  &  79.91&90.70&81.75&\textbf{100.00}&\textbf{100.00}&\textbf{100.00}&\textbf{100.00}&\textbf{100.00}&\textbf{100.00} 
 \\
    metal\_nut  &  98.30&99.38&97.71&98.65&99.62&98.41&\textbf{99.72}&\textbf{99.91}&\textbf{99.21} 
 \\
    pill  &  88.54&96.56&92.39&91.23&97.78&90.91&\textbf{98.28}&\textbf{99.58}&\textbf{97.92} 
 \\
    screw  &  89.01&94.54&88.24&85.06&93.87&85.33&\textbf{93.47}&\textbf{97.07}&\textbf{90.45} 
 \\
    tile  & 99.68&99.81&99.13&99.68&99.81&99.13&\textbf{100.00}&\textbf{100.00}&\textbf{100.00} 
 \\
    toothbrush  &  75.00&86.99&80.00&90.00&95.13&90.00&\textbf{95.00}&\textbf{97.65}&\textbf{94.74} 
 \\
    transistor  &  83.04&73.59&74.19&\textbf{100.00}&\textbf{100.00}&\textbf{100.00}&99.52&99.16&96.43 
 \\
    wood  &  93.36&95.60&95.45&98.62&99.49&97.62&\textbf{99.87}&\textbf{99.94}&\textbf{98.82} 
 \\
    zipper  &  98.48&99.51&98.14&\textbf{100.00}&\textbf{100.00}&\textbf{100.00}&\textbf{100.00}&\textbf{100.00}&\textbf{100.00} 
 \\
    \hline
    Average  &  90.74&94.43&90.37&96.62&98.61&96.21&\textbf{98.29}&\textbf{99.20}&\textbf{97.34} 
\\
    \specialrule{0.8pt}{0.1pt}{0pt}
    \end{tabular}
}
\vspace{-0.3cm}
\end{center}
\end{table*}

\begin{table*}[htb]
\caption{Comparison on supervised anomaly segmentation on LFD.}
\label{tab:pixel_ufnetall}
\begin{center}
\setlength{\tabcolsep}{2pt}
\resizebox{0.8\linewidth}{!}{
\begin{tabular}{c|cccc|cccc|cccc}
    \specialrule{0.6pt}{0.1pt}{0pt}
    
    \multirow{2}{*}{Category} & \multicolumn{4}{c|}{DFMGAN} & \multicolumn{4}{c|}{AnomalyDiffusion} & \multicolumn{4}{c}{Ours}  \\
    & AUROC &  AP & $F_1$-max& IoU &AUROC &  AP & $F_1$-max& IoU &AUROC &  AP & $F_1$-max& IoU \\
    \hline
    \hline
    bottle   & 90.41&61.51&58.49&40.19&98.71&89.64&81.55&67.10&\textbf{99.28}&\textbf{92.65}&\textbf{84.86}&\textbf{73.82} 
\\
    cable  & 96.49&79.40&\textbf{75.25}&53.47&\textbf{97.89}&\textbf{79.85}&72.75&53.69&94.53&75.41&72.70&\textbf{55.98} 
\\
    capsule  & 91.82&\textbf{56.11}&\textbf{58.56}&32.50&\textbf{95.80}&38.17&48.92&32.04&91.80&49.76&53.69&\textbf{41.14} 
\\
    carpet  & 89.10&48.04&49.89&39.46&94.83&53.15&51.79&42.21&\textbf{99.10}&\textbf{82.74}&\textbf{74.51}&\textbf{57.56} 
\\
    grid  & 89.18&34.89&41.21&19.21&85.19&24.32&34.76&18.22&\textbf{98.78}&\textbf{62.24}&\textbf{58.44}&\textbf{41.69} 
\\
    hazelnut  & \textbf{99.36}&\textbf{95.16}&\textbf{89.80}&\textbf{76.43}&98.54&77.39&70.42&45.97&98.97&88.00&81.77&73.39 
\\
    leather  & 97.82&51.86&52.25&48.09&98.99&65.73&62.85&42.65&\textbf{99.11}&\textbf{76.49}&\textbf{69.30}&\textbf{56.51} 
\\
    metal\_nut  &  98.16&95.16&90.99&63.02&\textbf{99.38}&\textbf{97.34}&\textbf{91.63}&64.59&99.23&96.66&91.42&\textbf{75.15} 
\\
    pill  & 95.80&75.90&70.31&31.73&\textbf{98.96}&\textbf{92.51}&\textbf{85.35}&50.04&98.11&79.63&72.54&\textbf{56.73} 
\\
    screw  & 93.96&38.00&41.69&30.88&92.68&44.64&49.17&34.08&\textbf{98.27}&\textbf{52.40}&\textbf{52.32}&\textbf{41.02} 
\\
    tile  & 97.37&88.79&82.05&66.30&92.98&79.59&73.52&55.08&\textbf{99.38}&\textbf{96.24}&\textbf{89.90}&\textbf{75.50} 
\\
    toothbrush  & 95.17&55.21&53.95&28.83&\textbf{98.31}&\textbf{68.60}&\textbf{66.14}&\textbf{29.67}&96.97&54.84&53.19&27.91 
\\
    transistor  & 97.68&\textbf{89.68}&\textbf{84.18}&46.98&98.20&83.97&75.84&44.22&\textbf{98.80}&84.32&77.02&\textbf{55.57} 
\\
    wood  & 97.47&77.72&70.91&58.77&95.68&67.54&63.06&42.78&\textbf{98.60}&\textbf{88.57}&\textbf{81.46}&\textbf{62.94} 
\\
    zipper  & 93.80&58.43&56.82&46.44&98.42&84.05&77.08&64.14&\textbf{99.15}&\textbf{86.67}&\textbf{79.09}&\textbf{69.37} 
\\
    \hline
    Average  & 94.91&67.06&65.09&45.49&96.30&69.77&66.99&45.77&\textbf{98.01}&\textbf{77.77}&\textbf{72.81}&\textbf{57.62}
\\
    \specialrule{0.8pt}{0.1pt}{0pt}
    \end{tabular}
}
\vspace{-0.3cm}
\end{center}
\end{table*}

\begin{table*}[!htb]
\caption{Comparison on image-level anomaly detection on LFD.}
\label{tab:image_ufnetall}
\begin{center}
\setlength{\tabcolsep}{2pt}
\resizebox{0.72\linewidth}{!}{
\begin{tabular}{c|ccc|ccc|ccc}
    \specialrule{0.6pt}{0.1pt}{0pt}
    
    \multirow{2}{*}{Category} & \multicolumn{3}{c|}{DFMGAN} & \multicolumn{3}{c|}{AnomalyDiffusion} & \multicolumn{3}{c}{Ours}  \\
    & AUROC &  AP & $F_1$-max &AUROC &  AP & $F_1$-max &AUROC &  AP & $F_1$-max \\
    \hline
    \hline
    bottle   & 96.98&98.76&95.35&\textbf{100.00}&\textbf{100.00}&\textbf{100.00}&\textbf{100.00}&\textbf{100.00}&\textbf{100.00} 
 \\
    cable  & 90.98&94.21&88.14&\textbf{99.52}&\textbf{99.55}&\textbf{97.71}&92.05&94.95&88.70 
 \\
    capsule  & 86.32&95.99&88.46&83.25&94.62&89.44&\textbf{93.80}&\textbf{98.19}&\textbf{93.42 }
 \\
    carpet  & 88.02&95.33&87.60&86.00&93.42&87.22&\textbf{97.98}&\textbf{99.22}&\textbf{96.67} 
 \\
    grid  & 85.48&92.61&85.71&93.69&97.08&91.14&\textbf{96.79}&\textbf{98.76}&\textbf{96.10} 
 \\
    hazelnut  & 99.90&99.91&98.97&98.28&98.60&95.83&\textbf{100.00}&\textbf{100.00}&\textbf{100.00} 
 \\
    leather  &  95.93&98.15&93.65&99.90&99.95&99.20&\textbf{100.00}&\textbf{100.00}&\textbf{100.00} 
 \\
    metal\_nut  & 96.16&98.57&96.18&\textbf{99.01}&\textbf{99.65}&\textbf{98.46}&98.58&99.54&97.64 
 \\
    pill  &  82.85&94.40&92.00&94.15&98.42&94.47&\textbf{98.16}&\textbf{99.50}&\textbf{96.84} 
 \\
    screw  & 82.60&92.15&82.22&81.54&91.32&82.05&\textbf{87.83}&\textbf{94.39}&\textbf{85.54} 
 \\
    tile  & 98.94&99.43&96.55&98.25&99.13&95.65&\textbf{99.36}&\textbf{99.69}&\textbf{99.12} 
 \\
    toothbrush  & 77.08&87.68&80.95&\textbf{100.00}&\textbf{100.00}&\textbf{100.00}&87.92&94.08&87.80 
 \\
    transistor  &  88.04&85.06&77.78&97.38&96.57&92.86&\textbf{98.10}&\textbf{96.90}&\textbf{94.55} 
 \\
    wood  & 99.87&99.94&98.82&97.24&98.70&96.47&\textbf{100.00}&\textbf{100.00}&\textbf{100.00} 
 \\
    zipper  & 97.07&98.78&96.25&99.01&99.71&99.39&\textbf{100.00}&\textbf{100.00}&\textbf{100.00} 
 \\
    \hline
    Average  & 91.08&95.40&90.58&95.15&97.78&94.66&\textbf{96.70}&\textbf{98.35}&\textbf{95.76} 
\\
    \specialrule{0.8pt}{0.1pt}{0pt}
    \end{tabular}
}
\vspace{-0.3cm}
\end{center}
\end{table*}

\subsection{More qualitative comparison results of different supervised segmentation models trained on image-mask pairs generated by SeaS} 
\label{appendix:seg_result2}

\begin{figure*}[htb]
\centering
\includegraphics[width=0.8\linewidth]{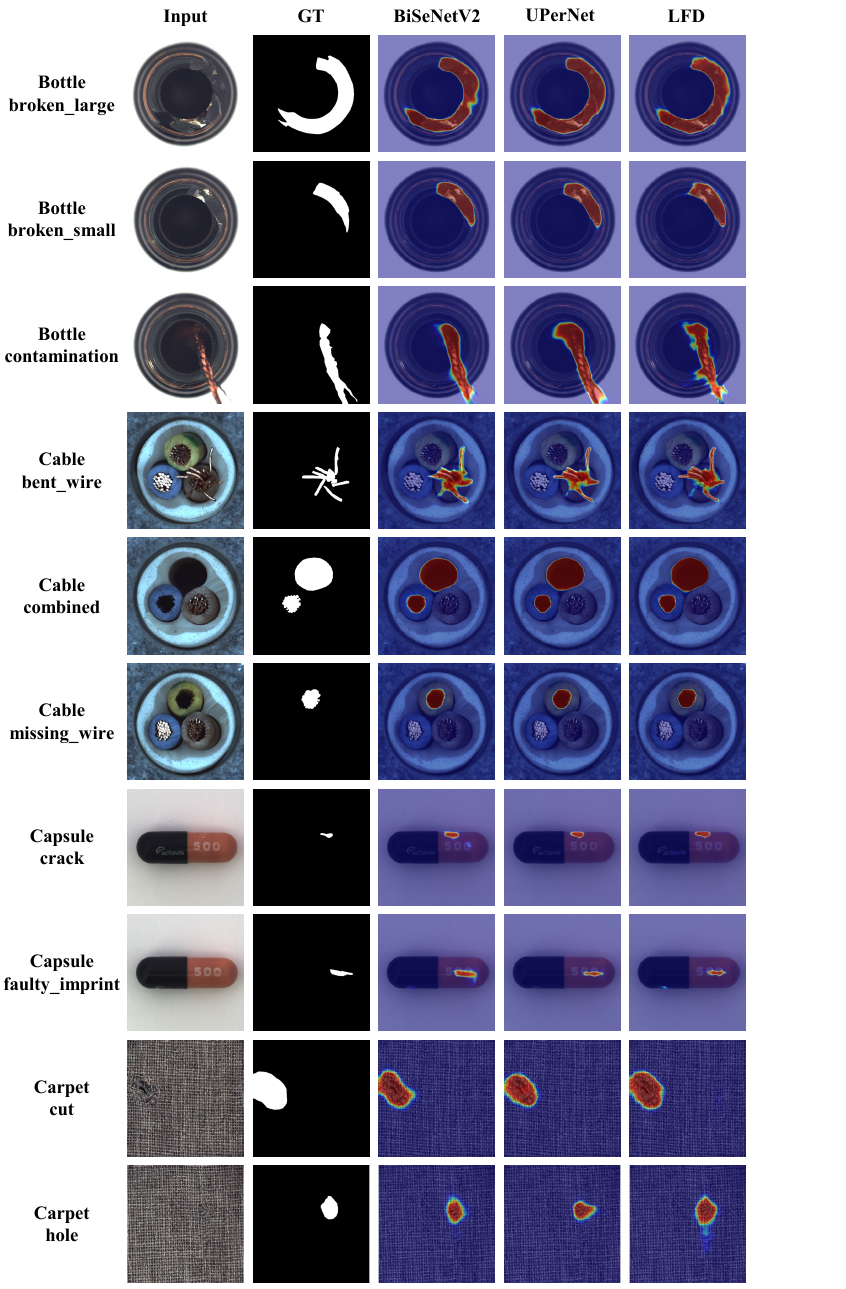}
\caption{Qualitative comparison results with the supervised segmentation models on MVTec AD. In the figure, from top to bottom are the results for \emph{bottle}, \emph{cable}, \emph{capsule} and \emph{carpet} categories.}
\label{fig:appendix_seg1_new}
\end{figure*}

\begin{figure*}[htb]
\centering
\includegraphics[width=0.8\linewidth]{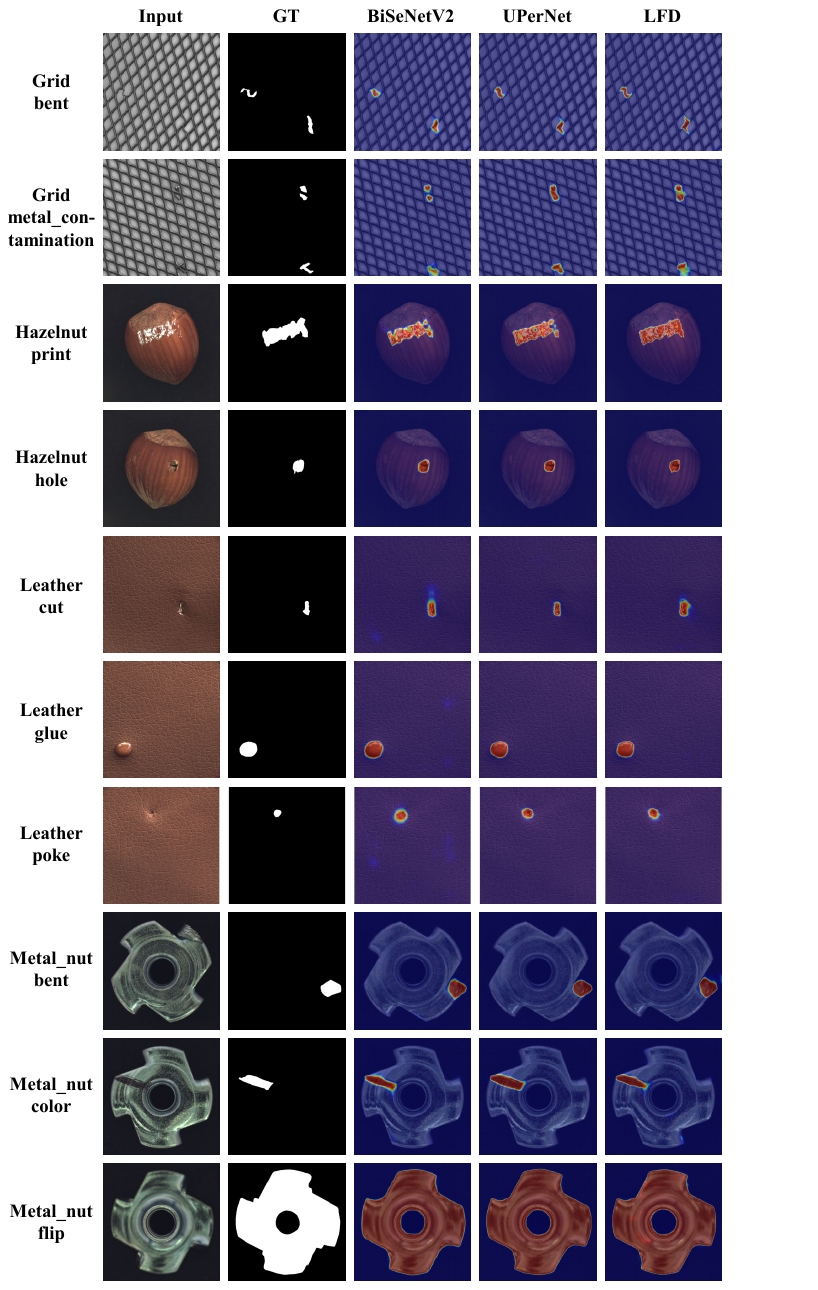}
\caption{Qualitative comparison results with the supervised segmentation models on MVTec AD. In the figure, from top to bottom are the results for \emph{grid}, \emph{hazelnut}, \emph{leather} and \emph{metal\_nut} categories.}
\label{fig:appendix_seg2_new}
\vspace{-0.5cm}
\end{figure*}

\begin{figure*}[htb]
\centering
\includegraphics[width=0.8\linewidth]{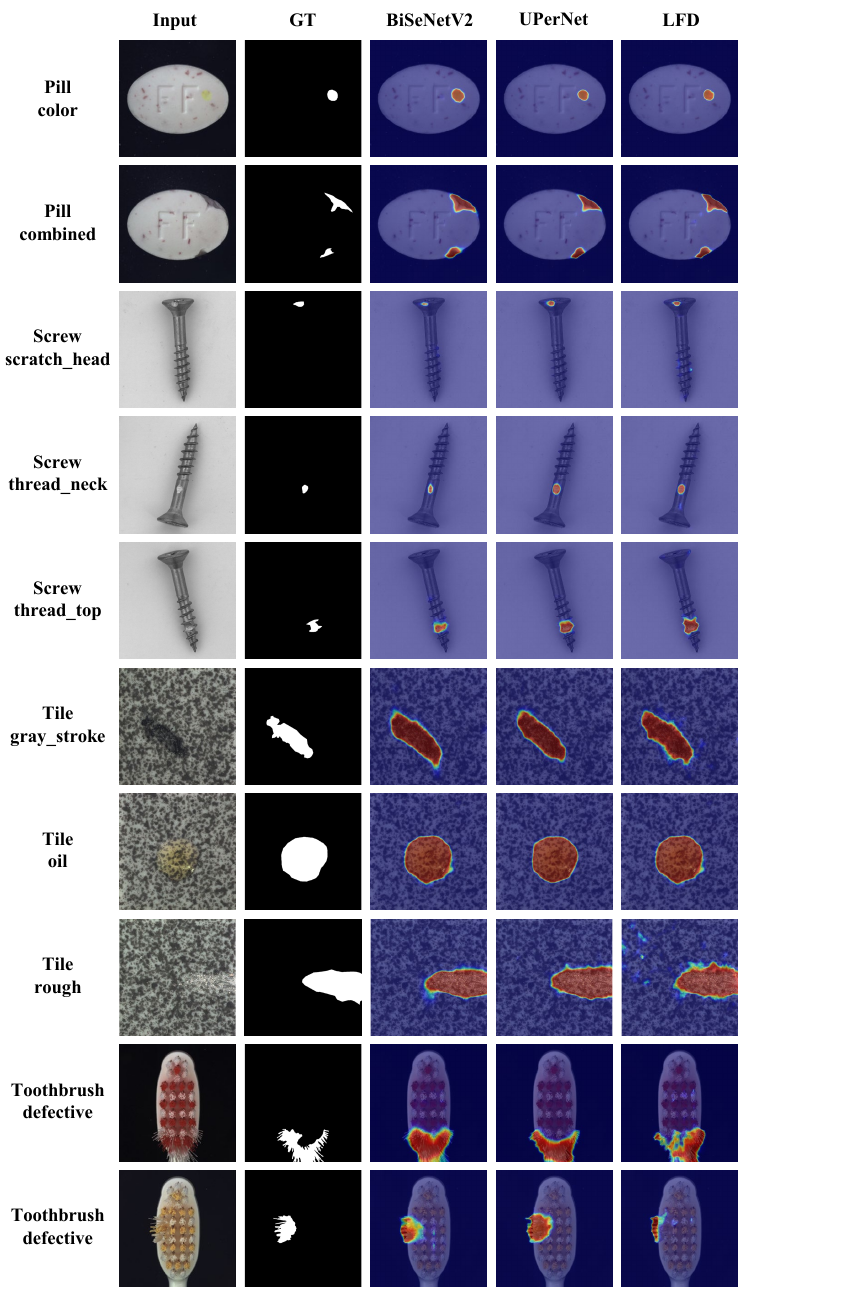}
\caption{Qualitative comparison results with the supervised segmentation models on MVTec AD. In the figure, from top to bottom are the results for \emph{pill}, \emph{screw}, \emph{tile} and \emph{toothbrush} categories.}
\label{fig:appendix_seg3_new}
\vspace{-0.5cm}
\end{figure*}

\begin{figure*}[!htb]
\centering
\includegraphics[width=0.8\linewidth]{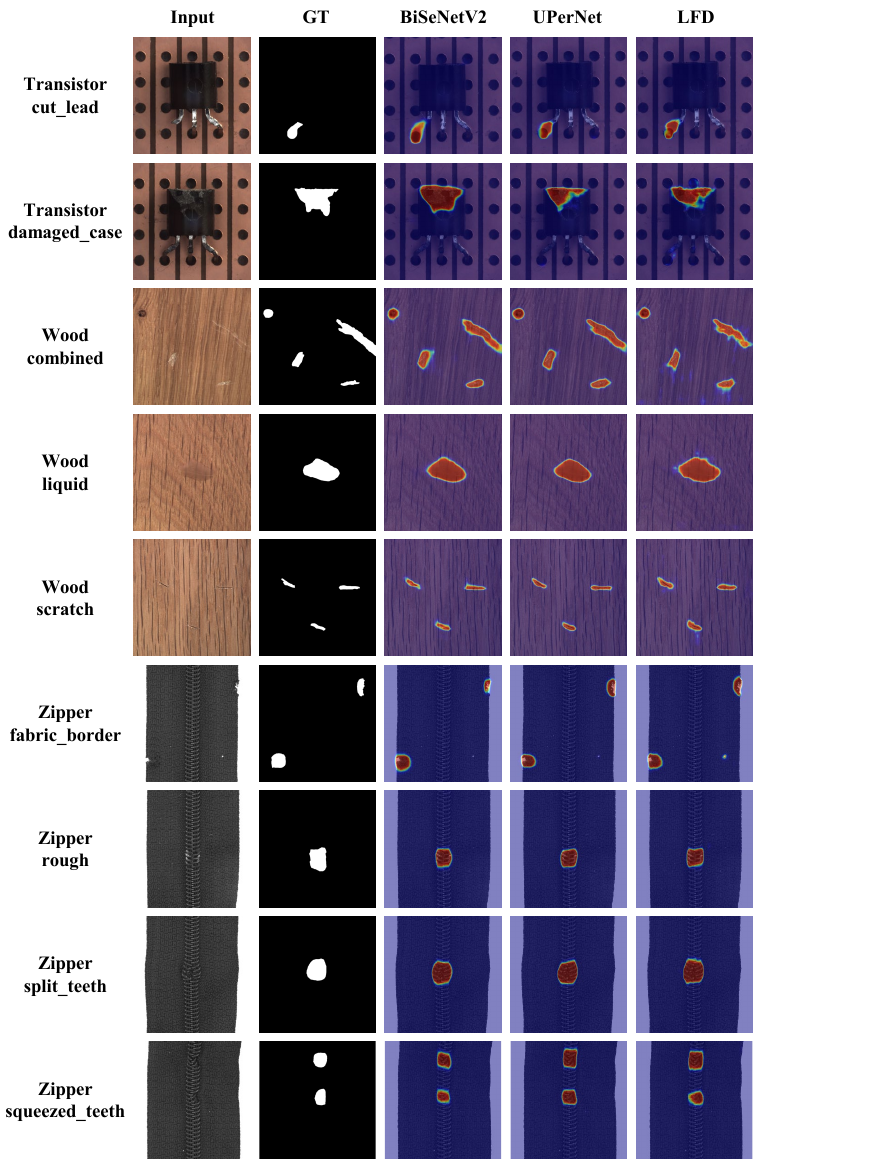}
\caption{Qualitative comparison results with the supervised segmentation models on MVTec AD. In the figure, from top to bottom are the results for \emph{transistor}, \emph{wood}, and \emph{zipper} categories.}
\label{fig:appendix_seg4_new}
\end{figure*}

\begin{figure*}[h]
\centering
\includegraphics[width=0.75\linewidth]{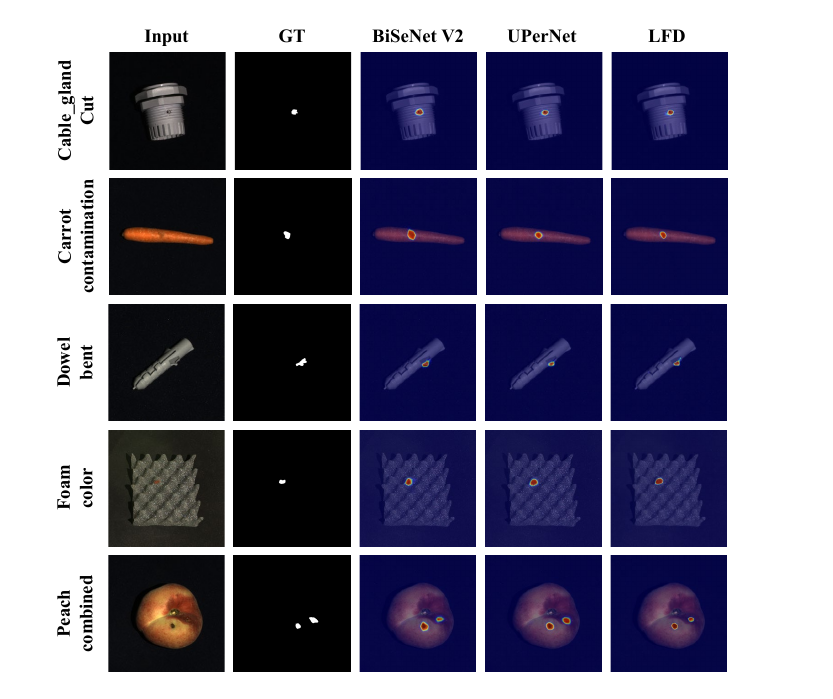}
\vspace{-0.3cm}
\caption{Qualitative comparison results with the supervised anomaly segmentation models on MVTec 3D AD. In the figure, from top to bottom are the results for \emph{cable\_gland}, \emph{carrot}, \emph{dowel}, \emph{foam} and \emph{peach} categories.}
\label{fig:appendix_3dseg2}
\end{figure*}

In this section, we provide further qualitative results with different supervised segmentation models on the MVTec AD and MVTec 3D AD datasets. We choose three models with different parameter quantity scopes (BiSeNet V2 \citep{yu2021bisenet}: 3.341M, UPerNet \citep{xiao2018unifiedupernet}: 64.042M, LFD \citep{zhou2024lfdroadseg}: 0.936M). We report the segmentation results of SeaS for varying types of anomalies in each category. Results are from Fig. \ref{fig:appendix_seg1_new} to Fig. \ref{fig:appendix_3dseg2}.

\clearpage

\clearpage

\clearpage

\clearpage

\vspace{-0.3cm}

\clearpage

\subsection{Comparison with the Textual Inversion}
\label{appendix:comp_ti}
\begin{figure}[htb]
\centering
\includegraphics[width=1.0\linewidth]{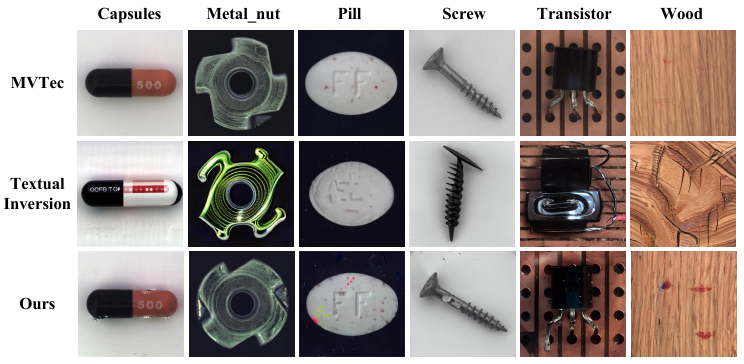}
\caption{Qualitative comparison on the generation results with Textual Inversion.}
\label{fig:appendix_ti}
\vspace{-0.5cm}
\end{figure}

We conduct the experiment of only using the Textual Inversion (TI) \citep{gal2022TI} method to learn the product, and the generated images are shown in Fig. \ref{fig:appendix_ti}. The TI method struggles to generate images similar to the real product due to the limited number of learnable parameters. In contrast, for the AIG method, the products satisfy global consistency with minor variations in local details, while the anomalies hold randomness, so the generated products should be globally consistent with the real products. Therefore, unlike the AG method AnomalyDiffusion \citep{hu2023anomalydiffusion}, where the TI method alone is sufficient to meet the anomaly generation needs, we fine-tune the U-Net to ensure the global consistency of the generated products.

\subsection{More experiments on lighting conditions} 
\label{appendix:lighting}
\begin{figure}[tb]
\centering
\includegraphics[width=1.0\linewidth]{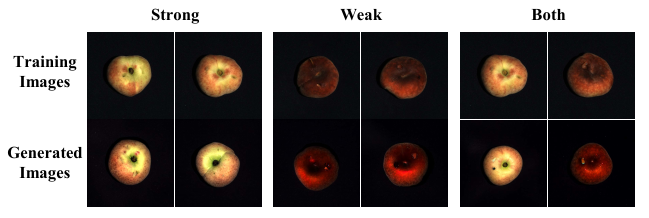}
\caption{Visualization of the generation results on MVTec3D AD on different lighting conditions and backgrounds. In the figure, the first row is for the training images and the second row is for the generated images.}
\label{fig:appendix_lighting}
\vspace{-0.5cm}
\end{figure}

We choose one defect class from peach, a product in the MVTec3D dataset, that has significant variations in lighting conditions and backgrounds, to conduct experiments. Images with strong lighting conditions depict the top side of the peach, whereas those with weak lighting conditions show the bottom side. Consequently, the background in the images, whether the top or bottom of the peach, also differs. We selected three training sets with different lighting conditions for experiments: 1) only images from the top side with strong lighting condition, 2) only images from the bottom side with weak lighting condition,  3) half of the images from the top side with strong lighting condition, and a half from the bottom side with weak lighting condition. The generated images of different settings are shown in Fig. \ref{fig:appendix_lighting}. It can be seen that SeaS is robust against lighting conditions and background variations.

\subsection{More results on generation of small defects.} 
\label{appendix:small}

\begin{figure}[tb]
\centering
\includegraphics[width=0.6\linewidth]{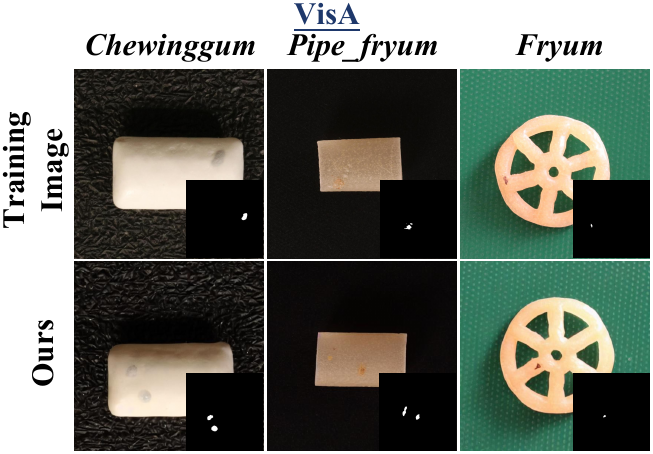}
\caption{Generation results of small-scale anomalies.}
\label{fig:appendix_small}
\vspace{-0.5cm}
\end{figure}

SeaS is capable of preserving fine-grained details in small-scale anomalies, as shown in Fig. \ref{fig:appendix_small}. However, generating extremely subtle anomalies may be challenging due to the limited resolution of the latent space. We will explore this point in our future work.

\subsection{More analysis on generation of unseen anomaly types.}
\label{appendix:unseen}

\begin{figure}[tb]
\centering
\includegraphics[width=0.8\linewidth]{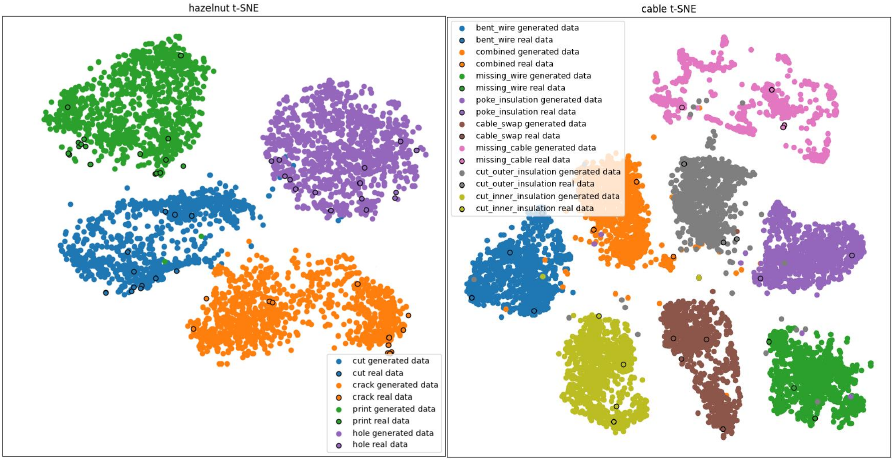}
\caption{T-SNE visualization of different anomaly types of the same product in real and generated data.}
\label{fig:appendix_tsne}
\vspace{-0.5cm}
\end{figure}

SeaS can generate diverse unseen anomalies within known anomaly types as analyzed in Appendix \ref{appendix:DAloss}. However, generating truly unseen anomaly types remains challenging.
The t-SNE visualizations in Fig. \ref{fig:appendix_tsne} show that different types of anomalies of the same product form compact clusters. Intra-cluster variation is achievable, but cross-cluster generalization is limited by the lack of prior knowledge. We believe that generalizing to unseen anomaly types is important and will explore this in future work.

\subsection{More experiments on comparison with DRAEM.}
\label{appendix:comp_draem}
As shown in Tab. \ref{table:combined_replace_add}, training DRAEM\citep{zavrtanik2021draem} with the same anomaly images used in SeaS leads to better results than using only anomaly-free images. However, DRAEM + SeaS achieves further improvements, demonstrating that the gain is not only from real anomalies but also from the diverse and realistic anomalies generated by SeaS.

\begin{table*}[!h]
\vspace{-0.5cm}
\caption{Comparison on combining generated anomalies with synthesis-based anomaly detection method across multiple datasets.}
\vspace{-0.6cm}
\label{table:combined_replace_add}
\renewcommand{\arraystretch}{1}
\begin{center}
\setlength{\tabcolsep}{2pt}
\resizebox{1\linewidth}{!}{
\begin{tabular}{c|ccc|cccc|ccc|cccc|ccc|cccc}  
    \toprule
    \multirow{3}{*}{\makecell{\textbf{Segmentation}\\\textbf{Models}}} & \multicolumn{7}{c|}{\textbf{MVTec AD}} & \multicolumn{7}{c|}{\textbf{VisA}} & \multicolumn{7}{c}{\textbf{MVTec 3D AD}} \\
    \cmidrule{2-22}  
    & \multicolumn{3}{c|}{\textbf{Image-level}} & \multicolumn{4}{c|}{\textbf{Pixel-level}} & \multicolumn{3}{c|}{\textbf{Image-level}} & \multicolumn{4}{c|}{\textbf{Pixel-level}} & \multicolumn{3}{c|}{\textbf{Image-level}} & \multicolumn{4}{c}{\textbf{Pixel-level}} \\
    & \textbf{AUROC} & \textbf{AP} & \textbf{$F_1$-max} & \textbf{AUROC} & \textbf{AP} & \textbf{$F_1$-max} & \textbf{IoU} & \textbf{AUROC} & \textbf{AP} & \textbf{$F_1$-max} & \textbf{AUROC} & \textbf{AP} & \textbf{$F_1$-max} & \textbf{IoU} & \textbf{AUROC} & \textbf{AP} & \textbf{$F_1$-max} & \textbf{AUROC} & \textbf{AP} & \textbf{$F_1$-max} & \textbf{IoU} \\
    \midrule
    \midrule
    DRAEM & 98.00 & 98.45 & 96.34 & 97.90 & 67.89 & 66.04 & \textbf{60.30} & 86.28&85.30&  81.66  &  92.92   &17.15&22.95& 13.57 &79.16&90.90 & 89.78 & 86.73&  14.02& 17.00& 12.42  \\
    DRAEM + training data & 97.43 & 98.84 & 97.84 & 96.41 & 74.42 & 71.86 & 59.84 
    & 83.74 & 86.00 & 82.75 & 94.63 & 39.22 & 43.06 & 29.02
    &73.86 & 88.46 & 86.69 & 82.43 & 19.36 & 25.05 &
17.01  \\
    DRAEM + SeaS & \textbf{98.64} & \textbf{99.40} & \textbf{97.89} & \textbf{98.11} & \textbf{76.55} & \textbf{72.70} &  58.87 &  \textbf{88.12}   &  \textbf{87.04}   &  \textbf{83.04}   &  \textbf{98.45}   & \textbf{49.05}   & \textbf{48.62}   &  \textbf{35.00} &  \textbf{85.45}   &  \textbf{93.58}  &  \textbf{90.85}  &   \textbf{95.43}  &  \textbf{20.09}   &  \textbf{26.10}  &   \textbf{17.07}   \\
    \bottomrule
\end{tabular}
}
\vspace{-0.55cm}
\end{center}
\end{table*}

\end{document}